\documentclass[preprint,12pt]{elsarticle}

\usepackage[T1]{fontenc}
\usepackage[utf8]{inputenc} 

\usepackage{amssymb,url}

\biboptions{numbers,sort&compress}
\usepackage{amsmath,amssymb, mathtools}   
\usepackage{amsthm}            
\usepackage{booktabs}          
\def\sep{, }

\usepackage{graphicx}          
\usepackage{xcolor}            
\colorlet{red}{black}
\colorlet{blue}{black}
\usepackage{float}             
\usepackage{placeins}          
\usepackage[font=small,labelfont=bf]{caption}
\usepackage[hidelinks]{hyperref}
\graphicspath{{figures/}}

\newcommand{\beginsupplement}{%
  \clearpage
  \renewcommand{\theHsection}{supp.\arabic{section}}%
  \renewcommand{\theHsubsection}{\theHsection.\arabic{subsection}}%
  \renewcommand{\theHsubsubsection}{\theHsubsection.\arabic{subsubsection}}%
  \renewcommand{\theHequation}{supp.\arabic{equation}}%
  \setcounter{section}{0}%
  \renewcommand{\thesection}{S\arabic{section}}%
  \setcounter{subsection}{0}%
  \renewcommand{\thesubsection}{\thesection.\arabic{subsection}}%
  \setcounter{figure}{0}%
  \renewcommand{\thefigure}{S\arabic{figure}}%
  \setcounter{table}{0}%
  \renewcommand{\thetable}{S\arabic{table}}%
  \setcounter{equation}{0}%
  \renewcommand{\theequation}{S\arabic{equation}}%
}

\newtheorem{theorem}{Theorem}[section]

\newtheorem{remark}[theorem]{Remark}

\begin{document}
\begin{frontmatter}

\title{Uncertainty calibration in\\ Bayesian M/EEG source imaging}
\author[inst1,inst2]{Ismail Huseynov\corref{cor}}
\author[inst1,inst2]{Mohammad Orabe}
\author[inst3]{Srikantan Nagarajan}
\author[inst1,inst2,inst4]{Stefan Haufe\corref{cor}}

\affiliation[inst1]{organization={Physikalisch‑Technische Bundesanstalt Braunschweig und Berlin}, city={Berlin}, country={Germany}}
\affiliation[inst2]{organization={Technische Universität Berlin}, city={Berlin}, country={Germany}}
\affiliation[inst3]{organization={University of California San Francisco}, addressline={Biomagnetic Imaging Laboratory, Department of Radiology and Biomedical Imaging}, city={San Francisco}, state={CA}, country={United States}}
\affiliation[inst4]{organization={Charité-Universitätsmedizin Berlin}, addressline={Charitéplatz 1}, city={Berlin}, country={Germany}}

\cortext[cor]{Corresponding authors: haufe@tu-berlin.de, ismail.huseynov@ptb.de}

\begin{abstract}

Many applications of M/EEG source imaging require not only accurate source localization but also reliable uncertainty estimates. However, most inverse solutions are still evaluated predominantly through point estimates, and even when Bayesian methods provide full posterior distributions, it is often unclear whether the resulting credible regions are empirically well calibrated. This motivates a systematic framework for both posterior uncertainty quantification and calibration in M/EEG source reconstruction.

We focus on empirical Bayesian inverse methods with closed-form Gaussian posterior summaries, including Bayesian minimum norm (BMN) and sparse Bayesian learning (SBL) methods. We hypothesize that these methods can provide near-calibrated uncertainties when combined with adaptive noise learning. Since voxel-wise SBL produces degenerate posterior variances at sources estimated to be zero, we combine SBL with sparse basis field expansions (S-FLEX), allowing retained basis coefficients to contribute nonzero marginal posterior variance at those locations. For BMN with sLORETA-based depth weighting, we derive joint source- and noise-variance updates through evidence maximization. Both methods showed residual miscalibration in our experiments. Therefore, we consider isotonic regression to adjust credible-region levels based on empirical coverage estimates as an additional post-hoc recalibration step.

In simulations, we benchmark BMN+sLORETA and SBL+S-FLEX for both fixed- and free-orientation source models, with and without post-hoc recalibration, and for different noise estimators. We evaluate calibration, as well as localization accuracy and posterior uncertainty, across different signal-to-noise ratios (SNR), source sparsity levels, and subject-specific head models. 

SBL+S-FLEX exhibits lower localization error, measured by the Wasserstein distance metric, smaller posterior uncertainty, and empirical coverage closer to nominal credibility than BMN+sLORETA before recalibration. Isotonic recalibration further improves the calibration of both methods, with the largest gains for the conservative BMN+sLORETA posterior, whereas transfer across mismatched SNR, sparsity, and head model conditions remains condition dependent. The three noise specifications (oracle, baseline-estimated, and adaptively learned) yielded broadly consistent reconstruction and uncertainty trends within each estimator.

Finally, application to real auditory MEG data demonstrates not only more focal localization but also tighter uncertainties for SBL+S-FLEX as compared to dense BMN+sLORETA, suggesting higher statistical power in identifying the expected sources in bilateral auditory cortices.
\end{abstract}


\end{frontmatter}

\section{Introduction}
Magneto- and electroencephalography (M/EEG) are indispensable non-invasive tools for monitoring brain activity and provide insight into both normal function and pathology. In clinical contexts such as surgical planning for drug-resistant epilepsy, precise localization of abnormal activity to specific brain structures is imperative~\citep{Pittau2014}. Achieving such precision requires solving the inverse problem that maps sensor-level M/EEG measurements to the underlying neuronal current sources. Because the M/EEG inverse problem is highly underdetermined and measurements are contaminated by noise, prior assumptions and noise modeling critically affect source localization accuracy. The problem is therefore commonly characterized as \emph{ill-posed} and \emph{severely ill-conditioned}, meaning that even small perturbations in the sensor data may lead to large differences in the reconstructed sources. Although a broad range of Bayesian and empirical-Bayesian source imaging methods has been developed for this setting~\citep{TrujilloBarreto2004,Sato2004,Phillips2005,Mattout2006,Friston2008,Henson2010,Jun2008,Lucka2012,Wipf2009,Wipf2010,Cai2019,Cai2021,Hashemi2021a,Hashemi2021b,Hashemi2024,Ghosh2025}, most studies focus on point estimates, such as posterior means, as the main output quantity while neglecting their associated uncertainty, for example as encoded in posterior variances or local covariance blocks. In high-stakes biomedical applications, however, calibrated uncertainty can be as important as localization accuracy itself. For example, in presurgical epilepsy evaluation, an overconfident source estimate may give the misleading impression that abnormal activity is precisely localized within or outside a candidate resection zone, even when the inverse problem admits substantial spatial uncertainty~\citep{Pittau2014}. This creates a strong need not only to quantify posterior uncertainty, but also to assess whether that uncertainty is well \emph{calibrated}, meaning that nominal credible regions are accurately mirrored by empirical coverages.

The study of uncertainty quantification (UQ) and uncertainty calibration (UC) in M/EEG source imaging is still comparatively limited. Bayesian modeling has enabled state-of-the-art source reconstruction algorithms through application-specific priors and hierarchical variance models~\citep{Cai2021,Wipf2009,Wipf2010,Cai2019,Hashemi2021a,Hashemi2021b,Hashemi2024,Ghosh2025}. A key advantage of these methods is that, beyond posterior means, they also yield posterior covariance information that can in principle be used for uncertainty quantification. In applications such as presurgical epilepsy evaluation, this means that the inverse solution should ideally provide not only a map of estimated source strengths, but also information about the reliability of these estimates, for example through probabilities of source presence, posterior variances, or credible intervals and regions for local neural current activity~\citep{Pittau2014}.

Minimum-norm estimators were among the first methods that used Bayesian 
maximum-a-posteriori (MAP) estimation for ill-posed electromagnetic inverse problems~\citep{Hamalainen1994}. Dale and colleagues introduced dynamic statistical parametric mapping, combining anatomical and functional constraints with MEG to obtain spatiotemporal cortical activity maps~\citep{Dale2000}. Pascual-Marqui and colleagues developed the LORETA/sLORETA/eLORETA family, which uses standardized current-density estimates to reduce localization bias in distributed source imaging~\citep{PascualMarqui1994,PascualMarqui2002}. Friston and colleagues developed empirical-Bayesian and multiple-sparse-prior formulations, in which source covariance components and hyperparameters are estimated from the data~\citep{Phillips2005,Mattout2006,Friston2008,Henson2010}. In parallel, He and colleagues advanced electrophysiological source imaging for high-density EEG and clinical applications, including epilepsy-related source localization~\citep{He2018,Michel2019}. All of these algorithms can be viewed as Type-I penalized likelihood estimation methods~\citep{Wipf2009,Hashemi2021a}). These contributions have strongly shaped modern Bayesian source reconstruction, but their regularizers and source priors were developed primarily to improve localization accuracy of the MAP estimate rather than to achieve empirical calibration of posterior credible intervals or regions.

Early work on uncertainty quantification in M/EEG inverse modeling used Bayesian analysis to estimate error bounds for EEG source reconstructions and examined their behavior in simulations~\citep{Russell1998}. Conductivity-related uncertainty in EEG source imaging has also been addressed using Bayesian
approximation error modelling and neural stochastic differential equation networks~\citep{Rimpilainen2019,Wabina2023}. SESAMEEG explicitly quantifies posterior uncertainty in the number and locations of M/EEG sources using approximate Bayesian inference
via sequential Monte Carlo~\citep{Luria2024}.

A central unresolved question is whether nominal credibility translates into empirical coverage across different source, noise, and head-model conditions, and how systematic discrepancies
can be corrected. We address this question for Type-II Bayesian source-imaging methods with closed-form conditional posterior summaries through a unified simulation benchmark of localization accuracy, posterior uncertainty, and calibration, together with post-hoc recalibration evaluated on held-out data.

Uncertainty calibration asks whether given uncertainty information has the
intended empirical coverage behavior when evaluated against simulated or
validation cases with known ground truth. In measured-data inverse problems,
exact ground truth is generally unavailable; however, one can ask whether
source parameters drawn from a plausible data-generating process fall inside
nominal (e.g., 95\%) credible regions in the expected fraction (95\%) of test
cases. If this agreement holds (approximately), the inverse procedure is
empirically well-calibrated. If empirical coverage is systematically larger
than the nominal level, the method is \textit{underconfident}; if empirical
coverage is systematically smaller, the method is \textit{overconfident}.
This distinction is important in clinical interpretation. An underconfident
method may make a correct localization appear too uncertain to support a
decision, whereas an overconfident method may give unjustified confidence in
an incorrect localization. In the epilepsy example, both failure regimes are
problematic: the former may obscure a true seizure-onset zone, while the latter
may overstate evidence for an incorrect candidate region.

Bayesian posteriors represent uncertainty about inverse parameters
conditional on the observed data, the prior, and the assumed model. A
Bayesian credible region is a set assigned a prescribed posterior probability
mass, for example 0.95. Thus, conditional on the model and observed data,
the posterior probability assigned to the parameter lying in a 95\% credible
region is 0.95. This posterior statement is distinct from the empirical
coverage property considered below.

However, the quantification of Bayesian credible regions is not, by itself, a guarantee of empirical coverage in practical applications. It is conditional on correct model specification and a prior that accurately encodes domain knowledge. In current M/EEG source imaging practice, priors are mainly used to encode low-level structural assumptions on the posterior mean such as sparsity or smoothness in certain domains. Besides zero-mean assumptions, quantitative knowledge is rarely available to be encoded in M/EEG priors. In particular, prior and noise co/variances critically shape posterior uncertainties but are rarely fully specified. These parameters can be modeled in a fully-Bayesian (hierarchical) way~\citep{Jun2008,Lucka2012} or be estimated directly through Bayesian evidence maximization also known as Type-II maximum-likelihood estimation~\citep{Wipf2009,Friston2008}. Both approaches, however, typically provide only approximations to the posterior. Together with shrinkage effects and posterior degeneracies of sparsity-inducing mechanisms, all these factors may lead to systematic discrepancies between nominal posterior credibility and empirical coverage. To the best of the authors' knowledge, such discrepancies have not been systematically quantified yet.

The present study benchmarks uncertainty calibration in Bayesian source imaging across a broad range of simulated settings and introduces methodological extensions to address four questions:

\begin{enumerate}
\item How can uncertainty and calibration be assessed for fixed- and free-orientation sources?

\item How can voxel-wise marginal posterior degeneracy in sparse Bayesian
learning (SBL) be mitigated to support meaningful UQ and UC?

\item How do sparse and dense empirical Bayesian methods with Gaussian posteriors compare with respect to uncertainty and calibration? How does adaptive learning of noise variances compare to using oracle or baseline-derived estimates?

\item Can post-hoc recalibration improve calibration on unseen data
with different source configurations, noise levels, and anatomical
head models?
\end{enumerate}
We address these questions through two complementary contributions: UQ- and UC-ready empirical-Bayesian inverse methods and a benchmarking framework for assessing their calibration performance. The methodological idea is to reduce ad hoc variance tuning by learning variance-related parameters directly from the observed data, so that the inverse methods can be applied in a zero-shot empirical-Bayesian manner. On the methodological side, our study focuses on two Type-II Bayesian learning approaches with distinct structural behavior. The first considered method is Bayesian minimum norm (BMN), which employs a dense Gaussian prior and learns a \emph{single common scalar} source variance hyperparameter~\citep{SekiharaNagarajan2015}. We extend this dense-prior model to a fully empirical-Bayesian minimum-norm variant in which the source variance and the sensor-noise variance are learned jointly from the observed data, and we combine it with sLORETA-based depth weighting to reduce depth bias while preserving a dense posterior representation. The second method is \(\gamma\)-MAP, also known as Champagne, which learns \emph{source-wise} variance hyperparameters and thereby promotes sparse reconstructions through automatic relevance determination (ARD) \citep{Wipf2009,Wipf2010,Cai2019,Hashemi2021a,Cai2021,Ghosh2025}. While sparsity can be highly beneficial for localization, it introduces a major difficulty for UQ: when sources are pruned, the corresponding posterior variances may collapse toward zero, so credible regions become degenerate or ill-defined. To mitigate this voxel-wise marginal degeneracy, we incorporate sparse basis field expansions, which replace strictly local sparse support by a spatially
extended basis representation. Retained basis functions can thereby contribute
nonzero posterior variance to cortical locations whose corresponding sparse coefficients are inactive, yielding informative local covariance summaries for uncertainty analysis~\citep{Haufe2008,Haufe2009}. The resulting joint source covariance may nevertheless remain globally rank deficient. Thus, BMN+sLORETA with adaptive noise learning represents the dense-prior case, whereas $\gamma$-MAP with S-FLEX represents the sparse-prior case in which the uncertainty retained in the sparse coefficient representation is spatially distributed over cortical source space. Both methods are further combined with post-hoc isotonic recalibration as an additional layer for improving empirical calibration.

On the benchmarking side, we establish a unified simulation-based framework for UQ and UC in M/EEG source imaging. This framework addresses the questions of vector-valued source uncertainty and fixed-versus-free orientation by covering both fixed-orientation and free-orientation source models within a single dimension-matched perspective. Starting from the full multivariate Gaussian posterior, we use Gaussian marginalization to derive voxel-wise uncertainty summaries suitable for evaluation and visualization. This leads naturally to \emph{credible intervals} in the fixed-orientation setting, credible ellipses for reduced free-orientation MEG in the locally SVD-defined two-dimensional MEG-sensitive subspace, and \emph{credible ellipsoids} for free-orientation EEG. These dimension-matched credible regions form the basis for empirical coverage analysis and calibration assessment across modalities and source models.

We then use this framework to conduct an extensive simulation benchmark of calibration performance. BMN+sLORETA and \(\gamma\)-MAP+S-FLEX are evaluated under three noise specifications: oracle noise, baseline-estimated noise, and adaptively learned noise. For each method and noise setting, we assess calibration before and after post-hoc recalibration, including matched, pooled-mismatch, and fixed recalibration variants. The experiments are carried out in fixed-orientation and reduced free-orientation MEG settings, and they vary the signal-to-noise level, the number of true active sources, and the subject-specific head model. Performance is quantified using localization accuracy, absolute posterior uncertainty, and calibration metrics that capture overall miscalibration, signed bias, maximal underconfidence, and maximal overconfidence. This design allows us to compare dense and sparse priors, fixed- and free-orientation source models, different noise-estimation regimes, and different recalibration regimes on unseen data. Altogether, the proposed methods and benchmarking framework provide a common basis for studying reconstruction accuracy, posterior uncertainty, and UC for Bayesian M/EEG inverse solvers. Finally, we illustrate the proposed uncertainty summaries using the standard auditory condition of the Brainstorm CTF MEG dataset.

\section{Methods}

\subsection{Source Reconstruction}

\subsubsection{Orientation-aware problem formulation}
\label{subsec:uq_problem_formulation}

We consider the linear forward model for M/EEG measurements:
\begin{equation}
  \mathbf{y}(t) \;=\; \mathbf{L}\,\mathbf{x}(t) \;+\; \mathbf{e}(t),
  \qquad t=1,\dots,T,
  \label{eq:forward_model_general}
\end{equation}
where \(\mathbf{y}(t)\in\mathbb{R}^{M}\) denotes the sensor measurement at time \(t\), \(\mathbf{x}(t)\in\mathbb{R}^{dN}\) is the unknown source current vector at time \(t\), \(\mathbf{L}\in\mathbb{R}^{M\times dN}\) is the lead-field matrix, and \(\mathbf{e}(t)\in\mathbb{R}^{M}\) is additive sensor noise at time \(t\), where \(M\) is the number of sensors, \(N\) is the number of source locations, and \(d\in\{1,2,3\}\) is the local source dimension. Concatenating all time points yields
\[
  \mathbf{Y}
  =
  [\mathbf{y}(1),\dots,\mathbf{y}(T)]
  \in\mathbb{R}^{M\times T},
  \qquad
  \mathbf{X}
  =
  [\mathbf{x}(1),\dots,\mathbf{x}(T)]
  \in\mathbb{R}^{dN\times T} \;.
\]
We assume temporally independent zero-mean Gaussian sensor noise with isotropic covariance,
\begin{equation}
  \mathbf{e}(t)
  \sim
  \mathcal{N}(\mathbf{0},\lambda \mathbf{I}_M),
  \qquad t=1,\dots,T,
  \label{eq:noise_model_general}
\end{equation}
where \(\lambda>0\) denotes the sensor-noise variance. In the synthetic-data setting described here, this variance is induced by the \(\alpha_{\mathrm{SNR}}\) scaling described in Section~\ref{subsec:synthetic_data_generation}.

To treat fixed- and free-orientation source models within a unified formulation, we represent the source vector at each time point in block form:
\[
  \mathbf{x}(t)
  =
  \begin{bmatrix}
    \mathbf{x}_1(t)\\
    \vdots\\
    \mathbf{x}_N(t)
  \end{bmatrix}
  \in \mathbb{R}^{dN},
  \qquad
  \mathbf{x}_i(t)\in\mathbb{R}^{d},
\]
where \(\mathbf{x}_i(t)\) is the local source block at cortical location \(i\). The parameter \(d\) specifies how many source coefficients are assigned to a single location:
\[
  d=
  \begin{cases}
    1, & \text{fixed-orientation},\\[1mm]
    2, & \text{free-orientation MEG},\\[1mm]
    3, & \text{free-orientation EEG}.
  \end{cases}
\]
In the fixed-orientation setting, each source location contributes a single scalar coefficient along the prescribed normal vector to the cortical surface. In the free-orientation EEG setting, each source location is represented by a three-dimensional current vector. In the free-orientation MEG setting, we retain the two locally most MEG-sensitive source directions obtained from the singular-value
decomposition of the three-component lead-field block. This yields an
effective two-dimensional representation whose local subspace need not
coincide with the cortical tangent plane.

Accordingly, the lead-field matrix takes the form
\[
  \mathbf{L}\in\mathbb{R}^{M\times dN}.
\]
The orientation-specific constructions of \(\mathbf{L}\) are given in Supplementary Section~\ref{sec:source_orientation_leadfields}.

With this notation, the stacked forward model becomes
\begin{equation}
  \mathbf{Y}
  =
  \mathbf{L}\mathbf{X}
  +
  \mathbf{E},
  \qquad
  \mathbf{E}
  =
  [\mathbf{e}(1),\dots,\mathbf{e}(T)]
  \in\mathbb{R}^{M\times T}.
  \label{eq:stacked_forward_model}
\end{equation}

The primary goal of M/EEG source imaging is to recover the unknown source matrix \(\mathbf{X}\) from the sensor data \(\mathbf{Y}\), given the lead-field matrix \(\mathbf{L}\). Since typically \(M\ll dN\), the inverse problem is severely ill-posed and requires regularization. In this work, we focus on Type-II Bayesian learning methods with Gaussian likelihoods and Gaussian source priors, in which source and, possibly, noise variance hyperparameters are learned from data by marginal-likelihood optimization~\citep{Wipf2009,Wipf2010,SekiharaNagarajan2015,Cai2019,Hashemi2021a,Cai2021,Ghosh2025}. Conditional on learned hyperparameters, these models yield closed-form Gaussian posterior distributions characterized by posterior means together with posterior covariance matrices, which form the basis of the subsequent uncertainty analysis.

\subsubsection{Full posterior distribution}
\label{subsec:full_posterior}

We consider a Gaussian likelihood together with a zero-mean Gaussian source prior. Let \(\boldsymbol{\theta}\) denote the hyperparameters governing the source-prior covariance and let \(\boldsymbol{\vartheta}\) denote the hyperparameters governing the sensor-noise covariance. For fixed values of these hyperparameters, Bayes' theorem yields
\begin{equation}
  p(\mathbf{X}\mid\mathbf{Y},
  \boldsymbol{\theta},
  \boldsymbol{\vartheta})
  =
  \frac{
  p(\mathbf{Y}\mid\mathbf{X},\boldsymbol{\vartheta})\,
  p(\mathbf{X}\mid\boldsymbol{\theta})
  }{
  p(\mathbf{Y}\mid\boldsymbol{\theta},\boldsymbol{\vartheta})
  }.
  \label{eq:bayes_general}
\end{equation}
The additive Gaussian noise model introduced above defines the Gaussian likelihood. The source prior is also assumed to be Gaussian,
\[p\!\bigl(\mathbf{x}(t)\mid\boldsymbol{\theta}\bigr)
=
\mathcal{N}\!\bigl(
\mathbf{x}(t)
\mid
\mathbf{0},
\boldsymbol{\Sigma}_{\mathrm{prior}}(\boldsymbol{\theta})
\bigr).\]
It follows that, conditional on fixed or learned hyperparameters, the posterior is Gaussian in all inverse models considered below. The specific covariance parameterizations and hyperparameter-learning rules are introduced separately for each inverse solver in the next subsection. After the hyperparameters have been fixed or estimated, we suppress them from the notation and use the shorthand
\[p(\mathbf{X}\mid\mathbf{Y})
\equiv
p\!\left(
\mathbf{X}\mid
\mathbf{Y},
\widehat{\boldsymbol{\theta}},
\widehat{\boldsymbol{\vartheta}}
\right),\]
with the understanding that, for settings in which a hyperparameter is prescribed rather than learned, the corresponding fixed value is used in place of its estimate.

Under the standard assumptions of temporal independence of both the source prior and the sensor-noise model, and conditional on fixed or learned hyperparameters, the posterior distribution factorizes over time:
\begin{equation}
  p(\mathbf{X}\mid\mathbf{Y})
  \;=\;
  \prod_{t=1}^{T} p\!\bigl(\mathbf{x}(t)\mid\mathbf{y}(t)\bigr)
  \;=\;
  \prod_{t=1}^{T}
  \mathcal{N}\!\bigl(
  \mathbf{x}(t)
  \mid
  \boldsymbol{\mu}_{\mathbf{x}}(t),
  \boldsymbol{\Sigma}_{\mathbf{x}}
  \bigr),
  \label{eq:factorized_posterior}
\end{equation}
where
\[
  \boldsymbol{\mu}_{\mathbf{x}}(t)\in\mathbb{R}^{dN},
  \qquad
  \boldsymbol{\Sigma}_{\mathbf{x}}\in\mathbb{R}^{dN\times dN}.
\]
Here \(\boldsymbol{\mu}_{\mathbf{x}}(t)\) denotes the posterior mean at time \(t\), whereas \(\boldsymbol{\Sigma}_{\mathbf{x}}\) is the posterior covariance at a single time point. In the models considered here, the source and sensor-noise models are temporally i.i.d. Therefore, once the hyperparameters have been fixed or learned from the data, the posterior covariance is identical for all \(t\). Thus, the posterior mean varies across time, whereas the posterior covariance remains static.

Equivalently, the posterior over the vectorized source matrix is Gaussian:
\begin{equation}
  p(\mathbf{X}\mid\mathbf{Y})
  =
  \mathcal{N}\!\bigl(
    \mathrm{vec}(\mathbf{X})
    \mid
    \mathrm{vec}(\boldsymbol{\mu}_\mathbf{X}),
    \boldsymbol{\Sigma}_\mathbf{X}
  \bigr),
  \label{eq:full_posterior_vec}
\end{equation}
with
\[
  \boldsymbol{\mu}_\mathbf{X}
  =
  \bigl[
  \boldsymbol{\mu}_{\mathbf{x}}(1)
  \;\cdots\;
  \boldsymbol{\mu}_{\mathbf{x}}(T)
  \bigr]
  \in\mathbb{R}^{dN\times T},
\]
and
\[
  \boldsymbol{\Sigma}_{\mathbf{X}}
  =
  \bigoplus_{t=1}^{T}\boldsymbol{\Sigma}_{\mathbf{x}}
  \in\mathbb{R}^{(dNT)\times(dNT)}.
\]
Here \(\bigoplus\) denotes the direct sum, i.e.\ a block-diagonal matrix with \(T\) identical diagonal blocks \(\boldsymbol{\Sigma}_{\mathbf{x}}\) and zeros elsewhere:
\[
  \boldsymbol{\Sigma}_{\mathbf{X}}
  =
  \begin{bmatrix}
    \boldsymbol{\Sigma}_{\mathbf{x}} & 0 & \cdots & 0 \\
    0 & \boldsymbol{\Sigma}_{\mathbf{x}} & \cdots & 0 \\
    \vdots & \vdots & \ddots & \vdots \\
    0 & 0 & \cdots & \boldsymbol{\Sigma}_{\mathbf{x}}
  \end{bmatrix}.
\]
This block-diagonal structure reflects posterior independence across time points under the model assumptions.

Although the full posterior is defined over the entire source matrix \(\mathbf{X}\), the factorization in \eqref{eq:factorized_posterior} shows that source reconstruction at a given instant \(t\) is fully characterized by the single-time posterior
\[
  p\!\bigl(\mathbf{x}(t)\mid\mathbf{y}(t)\bigr)
  =
  \mathcal{N}\!\bigl(
  \mathbf{x}(t)
  \mid
  \boldsymbol{\mu}_{\mathbf{x}}(t),
  \boldsymbol{\Sigma}_{\mathbf{x}}
  \bigr).
\]
Therefore, in the following subsection, the inverse solvers are formulated at each time point, while the subsequent uncertainty quantification and calibration analysis is based on temporally aggregated posterior summaries derived from the full posterior over \(\mathbf{X}\), as detailed in Supplementary Section~\ref{app:temporal_average_gaussian}.

\subsubsection{Inverse solvers}
\label{subsec:inverse_solvers}

We consider two Type-II Bayesian inverse methods:
$\gamma$-MAP (Champagne) and Bayesian minimum norm (BMN)
\citep{SekiharaNagarajan2015}. Both methods estimate source variance hyperparameters through Type-II marginal-likelihood maximization, with the sensor-noise variance either prescribed or jointly estimated. Under the
homoscedastic noise model, joint estimation proceeds through alternating convex-bounding updates, following \citet{Cai2021} for $\gamma$-MAP with S-FLEX. The corresponding update rules and posterior mean and covariance expressions are presented separately for each method.

\paragraph{\texorpdfstring{\(\gamma\)-MAP}{gamma-MAP} with S-FLEX}
\label{subsec:gammamap_S-FLEX}

In voxel-space $\gamma$-MAP, prior-level source pruning sets the corresponding posterior means and variances to zero, implying certainty about
inactivity conditional on the fitted model. To mitigate this
degeneracy, we incorporate sparse basis field expansions
\citep{Haufe2008,Haufe2009}, representing the source vector as
\begin{equation*}
  \mathbf x(t)=\mathbf B\mathbf c(t),
\end{equation*}
where $\mathbf B\in\mathbb R^{dN\times dN}$ is the spatial basis operator and $\mathbf c(t)\in\mathbb R^{dN}$ contains the basis
coefficients, grouped into $N$ blocks of dimension $d$.

In analogy to the S-FLEX basis fields of \citet{Haufe2009}, we construct the scalar spatial basis from isotropic, three-dimensional Gaussian bell-shaped profiles centered at source locations in anatomical space and sampled on the cortical source grid. Their width is parameterized relative to the source-grid spacing. The Gaussian kernel and its normalization, together with the orientation-specific constructions of \(\mathbf B\), are detailed in Supplementary
Sections~\ref{sec:sflex_grid_width} and~\ref{sec:sflex_orientation}.

With $\mathbf G=\mathbf L\mathbf B$, the coefficient-space model is
\begin{equation*}
  \mathbf y(t)=\mathbf G\mathbf c(t)+\mathbf e(t),
  \qquad
  \mathbf e(t)\sim
  \mathcal N(\mathbf 0,\boldsymbol{\Sigma}_{\mathbf e}),
\end{equation*}
with prior
\begin{equation*}
  \mathbf c(t)\sim
  \mathcal N(\mathbf 0,\boldsymbol{\Gamma}_{\mathbf c}),
  \qquad
  \boldsymbol{\Gamma}_{\mathbf c}
  =
  \operatorname{blkdiag}
  (\gamma_1\mathbf I_d,\ldots,\gamma_N\mathbf I_d),
\end{equation*}
where $\gamma_k\geq0$ is the variance of coefficient group $k$. In the fixed-noise formulation, the sensor-noise covariance
$\boldsymbol{\Sigma}_{\mathbf e}$ is treated as known.
The marginal data covariance in coefficient space is
\begin{equation*}
  \boldsymbol{\Sigma}_{\mathbf y,\mathbf c}
  =
  \mathbf G\boldsymbol{\Gamma}_{\mathbf c}\mathbf G^\top
  +
  \boldsymbol{\Sigma}_{\mathbf e}.
\end{equation*}

The variance hyperparameters are estimated by minimizing the
Type-II negative log-marginal likelihood, up to constants and
a positive scaling factor:
\begin{equation}
  \mathcal L(\boldsymbol{\Gamma}_{\mathbf c},\boldsymbol{\Sigma}_{\mathbf e})
  =
  \log|\boldsymbol{\Sigma}_{\mathbf y,\mathbf c}|
  +
  \frac{1}{T}\sum_{t=1}^{T}
  \mathbf y(t)^\top
  \boldsymbol{\Sigma}_{\mathbf y,\mathbf c}^{-1}
  \mathbf y(t). \label{eq:typeII_objective}
\end{equation}
In the fixed-noise setting, this objective is minimized over
$\boldsymbol{\Gamma}_{\mathbf c}$ while
$\boldsymbol{\Sigma}_{\mathbf e}$ is held fixed. Conditional on the current hyperparameters, the coefficient posterior is Gaussian, with
\begin{align*}
  \boldsymbol{\mu}_{\mathbf c}(t)
  =
  \boldsymbol{\Gamma}_{\mathbf c}\mathbf G^\top
  \boldsymbol{\Sigma}_{\mathbf y,\mathbf c}^{-1}\mathbf y(t),
  \qquad
  \boldsymbol{\Sigma}_{\mathbf c}
  =
  \boldsymbol{\Gamma}_{\mathbf c}
  -
  \boldsymbol{\Gamma}_{\mathbf c}\mathbf G^\top
  \boldsymbol{\Sigma}_{\mathbf y,\mathbf c}^{-1}
  \mathbf G\boldsymbol{\Gamma}_{\mathbf c}.
\end{align*}

We seek to minimize the Type-II objective using convex-bounding (CB) updates \citep{SekiharaNagarajan2015,Cai2021}. Because the objective is nonconvex in the variance hyperparameters, global optimality is not guaranteed, and the iterations may
converge to a local stationary point \citep{Hashemi2021a}.
Let $\boldsymbol{\mu}_{\mathbf c,k}^{(q)}(t)\in\mathbb R^d$
denote the $k$-th posterior-mean block at iteration $q$, and let
$\mathbf G_k\in\mathbb R^{M\times d}$ denote the corresponding block of $\mathbf G$. The grouped variance update is
\begin{equation}
  \gamma_k^{(q+1)}
  =
  \sqrt{
    \frac{
      \frac{1}{T}\sum_{t=1}^{T}
      \left\|
        \boldsymbol{\mu}_{\mathbf c,k}^{(q)}(t)
      \right\|_2^2
    }{
      \operatorname{tr}\!\left[
        \mathbf G_k^\top
        \left(
          \boldsymbol{\Sigma}_{\mathbf y,\mathbf c}^{(q)}
        \right)^{-1}
        \mathbf G_k
      \right]
    }
  },
  \qquad k=1,\ldots,N.
  \label{eq:gammamap_gamma_update_main_}
\end{equation}
The numerator is the time-averaged posterior-mean energy of coefficient group $k$, and the denominator measures its sensitivity through the effective forward operator.

After convergence, the posterior is evaluated at the estimated hyperparameters and mapped to source space:
\begin{equation}
  \boldsymbol{\mu}_{\mathbf x}(t)
  =
  \mathbf B\boldsymbol{\mu}_{\mathbf c}(t),
  \qquad
  \boldsymbol{\Sigma}_{\mathbf x}
  =
  \mathbf B\boldsymbol{\Sigma}_{\mathbf c}\mathbf B^\top.
  \label{eq:gammamap_x_mean_cov_main_}
\end{equation}
With S-FLEX, pruning acts on the basis coefficients.
The posterior covariance at source location $i$ is
\begin{equation*}
  \boldsymbol{\Sigma}_{\mathbf x,ii}
  =
  \mathbf B_i
  \boldsymbol{\Sigma}_{\mathbf c}
  \mathbf B_i^\top,
\end{equation*}
where $\mathbf B_i\in\mathbb R^{d\times dN}$ contains the rows of $\mathbf B$ associated with that location.
Consequently, setting $\gamma_i=0$ removes the contribution of the $i$-th coefficient group, but contributions from other
retained basis functions may still yield nonzero posterior
variances at source $i$.
The rank properties of the resulting covariance are discussed in Supplementary Section~\ref{sec:sflex_covariance_rank}.

\paragraph{Adaptive noise learning for
\texorpdfstring{\(\gamma\)-MAP}{gamma-MAP} with S-FLEX}
\label{subsec:gammamap_adaptive_noise_learning}

Following the joint Type-II framework of \citet{Cai2021},
we extend the fixed-noise formulation to jointly estimate
the coefficient variances and sensor-noise variance.
Under the homoscedastic white-noise assumption,
\begin{equation*}
  \boldsymbol{\Sigma}_{\mathbf e}
  =
  \lambda\mathbf I_M,
  \qquad
  \boldsymbol{\Sigma}_{\mathbf y,\mathbf c}
  =
  \mathbf G
  \boldsymbol{\Gamma}_{\mathbf c}
  \mathbf G^\top
  +
  \lambda\mathbf I_M.
\end{equation*}
The Type-II objective in Eq.~\eqref{eq:typeII_objective}
is minimized jointly over $\boldsymbol{\Gamma}_{\mathbf c}$
and $\lambda$.

The source and noise variances are updated alternately.
At iteration $q$, the source variances
$\boldsymbol{\gamma}^{(q+1)}$ are obtained from
Eq.~\eqref{eq:gammamap_gamma_update_main_}, using the posterior
mean and marginal data covariance evaluated at
$\boldsymbol{\gamma}^{(q)}$ and $\lambda^{(q)}$.
The posterior is then recomputed with
$\boldsymbol{\gamma}^{(q+1)}$ and $\lambda^{(q)}$.
The resulting intermediate covariance and posterior-mean
residual are
\begin{align*}
  \boldsymbol{\Sigma}_{\mathbf y,\mathbf c}^{\mathrm{int}}
  =
  \mathbf G
  \boldsymbol{\Gamma}_{\mathbf c}^{(q+1)}
  \mathbf G^\top
  +
  \lambda^{(q)}\mathbf I_M,
  \qquad
  \mathbf r^{\mathrm{int}}(t)
  =
  \mathbf y(t)
  -
  \mathbf G\boldsymbol{\mu}_{\mathbf c}^{\mathrm{int}}(t).
\end{align*}
The common noise variance is subsequently updated by the CB rule
\begin{equation}
  \lambda^{(q+1)}
  =
  \sqrt{
    \frac{
      \frac{1}{T}\sum_{t=1}^{T}
      \left\|\mathbf r^{\mathrm{int}}(t)\right\|_2^2
    }{
      \operatorname{tr}\!\left[
        \left(
          \boldsymbol{\Sigma}_{\mathbf y,\mathbf c}^{\mathrm{int}}
        \right)^{-1}
      \right]
    }
  }.
  \label{eq:gammamap_noise_update}
\end{equation}
Iterations terminate when the relative changes in the source
and noise variance hyperparameters fall below a prescribed
tolerance. The posterior is evaluated at the converged
hyperparameters and transformed to source space using
Eq.~\eqref{eq:gammamap_x_mean_cov_main_}.

\paragraph{Bayesian minimum norm (BMN) with sLORETA normalization}
\label{subsec:bmn_sloreta}

BMN uses a single shared source-variance hyperparameter
$\gamma>0$~\citep{SekiharaNagarajan2015}. We incorporate sLORETA-based depth normalization through
$\widetilde{\mathbf L}=\mathbf L\mathbf W$, where
$\mathbf W\in\mathbb R^{dN\times dN}$ is the normalization matrix.

The ridge parameter is scaled relative to the positive eigenvalues
of $\mathbf L\mathbf L^\top$:
\begin{equation*}
  \beta_{\mathrm{abs}}
  =
  \beta_{\mathrm{rel}}\,
  \operatorname{mean}
  \left\{
    \nu_j:
    \nu_j\in\operatorname{eig}(\mathbf L\mathbf L^\top),
    \ \nu_j>0
  \right\}.
\end{equation*}
For fixed-orientation sources, $\mathbf W$ is diagonal with
\begin{equation*}
  W_{ii}
  =
  \left[
    \mathbf l_i^\top
    \left(
      \mathbf L\mathbf L^\top
      +
      \beta_{\mathrm{abs}}\mathbf I_M
    \right)^{-1}
    \mathbf l_i
  \right]^{-1/2}.
\end{equation*}
For multicomponent sources, $\mathbf W$ is block diagonal with
\begin{equation*}
  \mathbf W_n=\mathbf S_n^{-1/2},
  \qquad
  \mathbf S_n
  =
  \mathbf L_n^\top
  \left(
    \mathbf L\mathbf L^\top
    +
    \beta_{\mathrm{abs}}\mathbf I_M
  \right)^{-1}
  \mathbf L_n.
\end{equation*}
The weights are subsequently normalized by a common geometric factor, whose global scaling is absorbed by $\gamma$.

With $\mathbf c(t)=\mathbf W^{-1}\mathbf x(t)$, the model becomes
\begin{equation*}
  \mathbf y(t)
  =
  \widetilde{\mathbf L}\mathbf c(t)+\mathbf e(t),
  \qquad
  \mathbf c(t)\sim
  \mathcal N(\mathbf 0,\gamma\mathbf I_{dN}).
\end{equation*}
For fixed $\boldsymbol{\Sigma}_{\mathbf e}$, the shared variance
$\gamma$ is estimated by minimizing
Eq.~\eqref{eq:typeII_objective} under the parameterization
$\mathbf G=\widetilde{\mathbf L}$ and
$\boldsymbol{\Gamma}_{\mathbf c}=\gamma\mathbf I_{dN}$, with
$\boldsymbol{\Sigma}_{\mathbf y}
=\gamma\widetilde{\mathbf L}\widetilde{\mathbf L}^{\top}
+\boldsymbol{\Sigma}_{\mathbf e}$. 

Conditional on $\gamma$, the posterior in normalized source coordinates is Gaussian, with
\begin{align*}
  \boldsymbol{\mu}_{\mathbf c}(t)
  =
  \gamma\widetilde{\mathbf L}^\top
  \boldsymbol{\Sigma}_{\mathbf y}^{-1}\mathbf y(t),
  \qquad
  \boldsymbol{\Sigma}_{\mathbf c}
  =
  \gamma\mathbf I_{dN}
  -
  \gamma^2\widetilde{\mathbf L}^\top
  \boldsymbol{\Sigma}_{\mathbf y}^{-1}
  \widetilde{\mathbf L}.
\end{align*}
The CB update for the shared variance is
\begin{equation}
  \gamma^{(q+1)}
  =
  \sqrt{
    \frac{
      \frac{1}{T}\sum_{t=1}^{T}
      \left\|
        \boldsymbol{\mu}_{\mathbf c}^{(q)}(t)
      \right\|_2^2
    }{
      \operatorname{tr}\!\left[
        \widetilde{\mathbf L}^\top
        \left(
          \boldsymbol{\Sigma}_{\mathbf y}^{(q)}
        \right)^{-1}
        \widetilde{\mathbf L}
      \right]
    }
  }.
  \label{eq:bmn_gamma_update}
\end{equation}
After convergence, the posterior is evaluated at the estimated variance and transformed to physical source coordinates:
\begin{equation*}
  \boldsymbol{\mu}_{\mathbf x}(t)
  =
  \mathbf W\boldsymbol{\mu}_{\mathbf c}(t),
  \qquad
  \boldsymbol{\Sigma}_{\mathbf x}
  =
  \mathbf W\boldsymbol{\Sigma}_{\mathbf c}\mathbf W^\top.
\end{equation*}

\paragraph{Adaptive noise learning for BMN with sLORETA normalization}
\label{subsec:bmn_adaptive_noise_learning}

The BMN formulation of \citet{SekiharaNagarajan2015} estimates the shared source variance by expectation maximization. Here, we adopt alternating convex-bounding updates for joint source and noise variance estimation, with the aim of improving
convergence speed in practice.

Under $\boldsymbol{\Sigma}_{\mathbf e}=\lambda\mathbf I_M$,
the Type-II objective in Eq.~\eqref{eq:typeII_objective}
is minimized jointly over $\gamma$ and $\lambda$.
The alternating procedure follows that of adaptive
$\gamma$-MAP with S-FLEX, with
$\mathbf G=\widetilde{\mathbf L}$ and
$\boldsymbol{\Gamma}_{\mathbf c}=\gamma\mathbf I_{dN}$.
At iteration $q$, the shared source variance is updated using Eq.~\eqref{eq:bmn_gamma_update}, evaluated at
$\gamma^{(q)}$ and $\lambda^{(q)}$.

The posterior is then recomputed at $\gamma^{(q+1)}$ and $\lambda^{(q)}$, and the noise variance is updated using
Eq.~\eqref{eq:gammamap_noise_update}, with
\begin{align*}
  \boldsymbol{\Sigma}_{\mathbf y,\mathbf c}^{\mathrm{int}}
  \;\longrightarrow\;
  \gamma^{(q+1)}
  \widetilde{\mathbf L}\widetilde{\mathbf L}^{\top}
  +\lambda^{(q)}\mathbf I_M,
  \qquad
  \mathbf r^{\mathrm{int}}(t)
  \;\longrightarrow\;
  \mathbf y(t)
  -\widetilde{\mathbf L}
  \boldsymbol{\mu}_{\mathbf c}^{\mathrm{int}}(t).
\end{align*}
The same convergence criterion is used, and the final posterior is transformed to physical source coordinates through $\mathbf x(t)=\mathbf W\mathbf c(t)$.

\subsection{Uncertainty Quantification}
\label{subsec:uncertainty_quantification}

Given the time-invariant posterior covariance structure established above, uncertainty
quantification can be formulated naturally for the temporally aggregated source vector
\[
  \bar{\mathbf{x}}
  \;=\;
  \frac{1}{T}\sum_{t=1}^{T}\mathbf{x}(t)
  \in\mathbb{R}^{dN}.
\]
This aggregation provides a compact summary of source activity over the analysis window while retaining a closed-form Gaussian posterior, which forms the basis for the source-wise credible regions and subsequent calibration analysis.

Because \(\bar{\mathbf{x}}\) is a linear transformation of \(\mathrm{vec}(\mathbf{X})\), its posterior is again Gaussian:
\begin{equation}
  \bar{\mathbf{x}} \mid \mathbf{Y}
  \sim
  \mathcal{N}\!\bigl(\bar{\boldsymbol{\mu}},\bar{\boldsymbol{\Sigma}}\bigr),
  \label{eq:avg_posterior}
\end{equation}
with
\[
  \bar{\boldsymbol{\mu}}
  =
  \frac{1}{T}\sum_{t=1}^{T}\boldsymbol{\mu}_{\mathbf{x}}(t),
  \qquad
  \bar{\boldsymbol{\Sigma}}
  =
  \mathrm{Cov}\!\left(\frac{1}{T}\sum_{t=1}^{T}\mathbf{x}(t)\right)
  =
  \frac{1}{T^2}\sum_{t=1}^{T}\boldsymbol{\Sigma}_{\mathbf{x}}
  =
  \frac{1}{T}\boldsymbol{\Sigma}_{\mathbf{x}}.
\]
Hence, temporal averaging preserves Gaussianity and scales the posterior covariance by the factor \(1/T\). Equivalently, posterior standard deviations scale by \(1/\sqrt{T}\). In practice, the analysis window used for temporal averaging should be chosen based on domain knowledge. In event-related potential or field (ERP/ERF) experiments, the averaging window should capture individual positive or negative source-level components to avoid cancellation between components of opposite sign. Alternatively, for non-time-locked activity, posterior uncertainty may be evaluated at individual time samples or for coefficients obtained through suitable linear spatial or temporal transformations, such as Fourier transforms. 

\subsubsection{Source-wise marginal posterior summaries}
\label{subsec:marginal_uncertainty}
The full posterior contains dependencies across source locations and source components. For uncertainty quantification, we obtain local posterior summaries by marginalizing the multivariate Gaussian posterior to each source block (see Supplementary
Section~\ref{app:gauss_marginals}).

We partition the posterior mean according to the \(N\) source locations:
\[
  \boldsymbol{\mu}_{\mathbf{x}}(t)
  =
  \begin{bmatrix}
    \boldsymbol{\mu}_{1}(t)\\
    \vdots\\
    \boldsymbol{\mu}_{N}(t)
  \end{bmatrix},
  \qquad
  \boldsymbol{\mu}_{i}(t)\in\mathbb{R}^{d},
\]
and write the posterior covariance in block form as
\[
  \boldsymbol{\Sigma}_{\mathbf{x}}
  =
  \begin{bmatrix}
    \boldsymbol{\Sigma}_{11} & \cdots & \boldsymbol{\Sigma}_{1N}\\
    \vdots & \ddots & \vdots\\
    \boldsymbol{\Sigma}_{N1} & \cdots & \boldsymbol{\Sigma}_{NN}
  \end{bmatrix},
  \qquad
  \boldsymbol{\Sigma}_{ij}\in\mathbb{R}^{d\times d}.
\]
The diagonal block \(\boldsymbol{\Sigma}_{ii}\) describes local posterior uncertainty at source \(i\), while off-diagonal blocks
encode dependence between source locations.

For the temporally aggregated posterior in
Eq.~\eqref{eq:avg_posterior}, the mean retains this block structure:
\[
  \bar{\boldsymbol{\mu}}
  =
  \begin{bmatrix}
    \bar{\boldsymbol{\mu}}_1\\
    \vdots\\
    \bar{\boldsymbol{\mu}}_N
  \end{bmatrix},
  \qquad
  \bar{\boldsymbol{\mu}}_i
  =
  \frac{1}{T}\sum_{t=1}^{T}\boldsymbol{\mu}_i(t).
\]
Its local covariance block is
\[
  \bar{\boldsymbol{\Sigma}}_{ii}
  =
  \frac{1}{T}\boldsymbol{\Sigma}_{ii}.
\]
The resulting aggregated posterior marginal at source \(i\) is
\begin{equation*}
  \bar{\mathbf{x}}_i\mid\mathbf{Y}
  \sim
  \mathcal{N}\!\bigl(
    \bar{\boldsymbol{\mu}}_i,
    \bar{\boldsymbol{\Sigma}}_{ii}
  \bigr).
\end{equation*}
These local marginals form the basis for the dimension-matched
credible regions and calibration analysis below.

\subsubsection{Dimension-matched credible regions}
\label{subsec:credible_regions}

For a Gaussian source-wise marginal posterior with
positive-definite covariance $\bar{\boldsymbol\Sigma}_{ii}$,
the squared Mahalanobis distance from the posterior mean
follows a chi-square distribution with $d$ degrees of
freedom \citep{Anderson2003}. The credible region at
nominal credibility $c\in(0,1)$ is therefore
\begin{equation}
  \mathcal C_i^{(c)}
  =
  \left\{
    \mathbf z\in\mathbb R^d:
    (\mathbf z-\bar{\boldsymbol\mu}_i)^\top
    \bar{\boldsymbol\Sigma}_{ii}^{-1}
    (\mathbf z-\bar{\boldsymbol\mu}_i)
    \le \chi^2_{d,c}
  \right\},
  \label{eq:generic_credible_region}
\end{equation}
where $\chi^2_{d,c}$ denotes the $c$-quantile of the
chi-square distribution with $d$ degrees of freedom.
The local source dimension determines both the geometry
of the region and its probability threshold.

\paragraph{Free-orientation EEG ($d=3$)}
For full Cartesian free-orientation EEG,
$\bar{\boldsymbol\mu}_i\in\mathbb R^3$ and
$\bar{\boldsymbol\Sigma}_{ii}\in\mathbb R^{3\times3}$.
Evaluating Eq.~\eqref{eq:generic_credible_region} with
$d=3$ and threshold $\chi^2_{3,c}$ gives a credible
ellipsoid in Cartesian source coordinates.

\paragraph{Reduced free-orientation MEG ($d=2$)}
For reduced free-orientation MEG, the local coordinates
span the MEG-sensitive subspace defined in Supplementary
Section~\ref{sec:source_orientation_leadfields},
Eq.~\eqref{eq:meg_tangent_basis}, with
$\bar{\boldsymbol\mu}_i\in\mathbb R^2$ and
$\bar{\boldsymbol\Sigma}_{ii}\in\mathbb R^{2\times2}$.
Evaluating Eq.~\eqref{eq:generic_credible_region} with
$d=2$ and threshold $\chi^2_{2,c}$ gives a credible
ellipse in these coordinates.

\paragraph{Fixed orientation ($d=1$)}
For the fixed-orientation representation defined in
Supplementary Section~\ref{sec:source_orientation_leadfields},
Eq.~\eqref{eq:fixed_orientation_leadfield}, uncertainty
concerns the signed current amplitude along the prescribed
surface normal. Writing
$\bar{\sigma}_i^2=\bar{\Sigma}_{ii}>0$ and using
$\chi^2_{1,c}=z_{(1+c)/2}^{\,2}$,
Eq.~\eqref{eq:generic_credible_region} with $d=1$
reduces to the credible interval
\begin{equation*}
  \mathcal C_i^{(c)}
  =
  \left[
    \bar{\mu}_i-z_{(1+c)/2}\bar{\sigma}_i,\;
    \bar{\mu}_i+z_{(1+c)/2}\bar{\sigma}_i
  \right],
\end{equation*}
where $z_{(1+c)/2}$ denotes the $(1+c)/2$-quantile
of the standard normal distribution.

\paragraph{SVD-defined directional marginal for
free-orientation sources}

To visualize free-orientation source dynamics, we project the local posterior onto a source-specific direction $\mathbf u_i$, defined by the leading left singular vector of the uncentered posterior-mean trajectory over a prespecified reference interval and held fixed throughout the epoch. In the auditory MEG analysis, this yields signed source time courses and pointwise credible intervals at selected cortical locations, complementing the joint credible regions used for calibration assessment.

Conditional on the observed data, the directional marginal is
\[
  \mathbf u_i^\top\mathbf x_i(t)\mid\mathbf Y
  \sim
  \mathcal N
  \left(
    \mathbf u_i^\top\boldsymbol\mu_i(t),
    \mathbf u_i^\top\boldsymbol\Sigma_{ii}\mathbf u_i
  \right).
\]
The construction and the corresponding credible intervals are detailed in Supplementary Section~\ref{subsec:supp_svd_directional_marginal}.

\subsection{Uncertainty Calibration}
\label{sec:uncertainty_calibration}

The previous subsection introduced aggregated source-wise credible regions \(\mathcal{C}_i^{(c)}\) at nominal credibility level \(c\in(0,1)\). Ideally, such regions should be calibrated in the sense that their nominal credibility agrees with their actual coverage probability. More precisely, a source-wise credible-region procedure is said to be \emph{calibrated} if, for every \(c\in(0,1)\),
\begin{equation}
  \mathbb{P}\!\left(
    \bar{\mathbf{x}}_i^{\mathrm{true}}
    \in
    \mathcal{C}_i^{(c)}(\mathbf{Y})
  \right)
  =
  c,
  \label{eq:calibration_definition}
\end{equation}
where the probability is taken with respect to the data-generating mechanism and the induced posterior procedure. In practice, however, model mismatch, shrinkage effects, and hyperparameter learning may induce systematic discrepancies between nominal credibility and
actual coverage. As a result, posterior credible regions can be miscalibrated even when they are derived from a closed-form Gaussian posterior.

We first assess calibration empirically by comparing nominal credibility levels with the observed coverage of the corresponding credible regions. We then consider post-hoc recalibration as a separate procedure for correcting systematic deviations from
the nominal coverage levels.

\subsection{Empirical calibration assessment}
\label{subsec:empirical_calibration_assessment}

\subsubsection{Empirical coverage and calibration curve}
\label{subsec:calibration_curve}

For each source location \(i=1,\dots,N\), let
\(\bar{\mathbf{x}}_i^{\mathrm{true}}\) denote the corresponding aggregated ground-truth source quantity expressed in the same local coordinate system as the credible region \(\mathcal{C}_i^{(c)}\).

The empirical coverage at nominal level \(c\) is defined by
\begin{equation}
  \hat{c}(c)
  \;=\;
  \frac{1}{N}\sum_{i=1}^{N}
  \mathbb{I}\!\left[
    \bar{\mathbf{x}}_i^{\mathrm{true}}
    \in
    \mathcal{C}_i^{(c)}
  \right],
  \label{eq:empirical_coverage_general}
\end{equation}
where \(\mathbb{I}[\cdot]\) denotes the indicator function.

Evaluating \(\hat{c}(c)\) over a grid of nominal credibility levels
\[
  0 < c_1 < c_2 < \cdots < c_K < 1
\]
yields the empirical calibration curve
\begin{equation*}
  c \;\longmapsto\; \hat{c}(c).
  \label{eq:calibration_curve_function}
\end{equation*}
Perfect calibration corresponds to the diagonal relation
\[
  \hat{c}(c)=c.
\]
If the empirical calibration curve lies above the diagonal, the posterior credible regions are \textit{underconfident}, meaning that they cover the ground truth more often than intended. If the curve lies below the diagonal, they are \textit{overconfident}, meaning that they cover the ground truth less often than intended.

This definition applies uniformly across all settings considered in this work. The resulting calibration curve provides the primary coverage-based description of uncertainty calibration, while we derive scalar summary calibration metrics from this curve later in
the evaluation section.

\subsubsection{Run-wise calibration curves}
\label{subsec:runwise_calibration}

Let
\[
  r = 1,\dots,R
\]
index completed simulation runs generated under a fixed experimental setting. For each run \(r\), source \(i=1,\dots,N\), and nominal credibility level \(c\), we obtain an aggregated credible region \(\mathcal{C}_{i,r}^{(c)}\) together with the corresponding
aggregated ground-truth source block \(\bar{\mathbf{x}}_{i,r}^{\mathrm{true}}\). The run-wise empirical coverage is defined by
\begin{equation}
  \hat{c}_{r}(c)
  \;=\;
  \frac{1}{N}\sum_{i=1}^{N}
  \mathbb{I}\!\left[
    \bar{\mathbf{x}}_{i,r}^{\mathrm{true}}
    \in
    \mathcal{C}_{i,r}^{(c)}
  \right].
  \label{eq:runwise_empirical_coverage}
\end{equation}
Thus, each run yields its own empirical calibration curve
\[
  c \;\longmapsto\; \hat{c}_{r}(c).
\]
These run-wise curves are used both for evaluating the variability of
calibration across independent simulations and, in the post-hoc procedure described below, for learning and evaluating recalibration maps on disjoint sets of runs.

\subsection{Post-hoc recalibration}
\label{subsec:posthoc_recalibration}

After evaluating the original posterior credible regions, we examine whether systematic miscalibration can be corrected by a post-hoc recalibration step. Using calibration runs, we learn a monotone mapping between nominal credibility levels and empirical coverage. This mapping is then applied to unseen test runs to obtain calibrated credibility levels. All post-calibration results reported in this work are obtained by this external recalibration of the empirical calibration curve. 

\subsubsection{Calibration and test split}
\label{subsec:calibration_test_split}

The available runs are partitioned into two disjoint sets,
\[ \mathcal{R}_{\mathrm{train}}
  \qquad\text{and}\qquad
  \mathcal{R}_{\mathrm{test}},
\]
where the training runs are used to estimate the recalibration map and the test runs are reserved for final evaluation. The concrete numbers of training and test runs are specified later in the simulation setup.

This separation ensures that the recalibration map is learned independently of the runs used for final calibration evaluation.

\subsubsection{Isotonic recalibration from training runs}
\label{subsec:isotonic_regression}

Empirical coverage is evaluated on a discrete grid of nominal
credibility levels
\[
  0 < c_1 < c_2 < \cdots < c_K < 1.
\]
At each level $c_k$, the mean training coverage is obtained
by averaging the run-wise coverages:
\begin{equation}
  \hat{c}_{\mathrm{train}}(c_k)
  =
  \frac{1}{|\mathcal{R}_{\mathrm{train}}|}
  \sum_{r\in\mathcal{R}_{\mathrm{train}}}
  \hat{c}_{r}(c_k),
  \qquad k=1,\dots,K.
  \label{eq:mean_train_curve}
\end{equation}
This curve serves as an empirical estimate of the calibration behavior of the estimator under the chosen setting.

To obtain a monotone recalibration map, we fit an isotonic regression function
\begin{equation}
  \hat g
  \;\in\;
  \arg\min_{f\in\mathcal{F}_{\uparrow}}
  \sum_{k=1}^{K}
  \Bigl(
    \hat{c}_{\mathrm{train}}(c_k)-f(c_k)
  \Bigr)^2,
  \label{eq:isotonic_fit}
\end{equation}
where
\[
  \mathcal{F}_{\uparrow}
  =
  \bigl\{
    f:[0,1]\to[0,1]
    \;|\;
    f \text{ is non-decreasing}
  \bigr\}.
\]
The monotonicity constraint is consistent with the nestedness
of credible regions, which implies that empirical coverage is
non-decreasing in the nominal credibility level. Thus, the
isotonic fit provides a monotone representation of the empirical
calibration relationship, which is subsequently inverted
numerically to obtain recalibrated credibility levels.

The fitted function $\hat g$ estimates the empirical
relationship between an input credibility level and its
resulting coverage. To determine which input level is
required to achieve a desired target coverage $c$, we
consider the generalized inverse
\begin{equation}
  \hat g^{-1}(c)
  :=
  \inf\bigl\{
    u\in[0,1] : \hat g(u)\ge c
  \bigr\},
  \qquad c\in[0,1],
  \label{eq:isotonic_inverse}
\end{equation}
whenever the defining set is nonempty.

For each target nominal level $c_k$, we therefore obtain
the recalibrated level
\begin{equation}
  c_k^{\mathrm{cal}}
  :=
  \hat g^{-1}(c_k),
  \qquad k=1,\ldots,K,
  \label{eq:recalibrated_levels}
\end{equation}
for attainable target levels. These adjusted levels are
subsequently used in the construction of post-hoc
recalibrated credible regions.

\begin{remark}
For a target credibility level $c_k$, recalibration replaces the
original threshold $\chi^2_{d,c_k}$ by
$\chi^2_{d,c_k^{\mathrm{cal}}}$, where
$c_k^{\mathrm{cal}}=\widehat g^{-1}(c_k)$. Thus,
\[
  C_i^{(c_k^{\mathrm{cal}})}
  =
  \left\{
    x\in\mathbb R^d:
    (x-\bar{\boldsymbol\mu}_i)^\top
    \bar{\boldsymbol\Sigma}_{ii}^{-1}
    (x-\bar{\boldsymbol\mu}_i)
    \le \chi^2_{d,c_k^{\mathrm{cal}}}
  \right\}.
\]
When $c_k,c_k^{\mathrm{cal}}\in(0,1)$, define
\begin{equation}
  b_d(c_k)
  :=\frac{\chi^2_{d,c_k^{\mathrm{cal}}}}
           {\chi^2_{d,c_k}}.
  \label{eq:recalibration_region_scale}
\end{equation}
The same region is the level-$c_k$ credible region of
$\mathcal N_d\!\left(
\bar{\boldsymbol\mu}_i,
b_d(c_k)\bar{\boldsymbol\Sigma}_{ii}
\right)$. Its center, principle directions, and axis ratios are
unchanged, while its axis lengths are multiplied by
$\sqrt{b_d(c_k)}$ relative to $C_i^{(c_k)}$.
This equivalence holds at each credibility level separately and may equivalently be regarded as a credibility-dependent rescaling of the posterior covariance.
\end{remark}

\subsubsection{Application to held-out runs}
\label{subsec:evaluation_on_test_runs}

The recalibration map estimated from the training runs is applied to the held-out runs \(r\in\mathcal{R}_{\mathrm{test}}\). For each test run and each nominal level \(c_k\), we evaluate both the original empirical coverage
\begin{equation}
\hat{c}^{\mathrm{pre}}_{r}(c_k)
  =
  \frac{1}{N}\sum_{i=1}^{N}
  \mathbb{I}\!\left[
    \bar{\mathbf{x}}_{i,r}^{\mathrm{true}}
    \in
    \mathcal{C}_{i,r}^{(c_k)}
  \right],
  \label{eq:test_pre_curve}
\end{equation}
and the post-calibration empirical coverage
\begin{equation}
  \hat{c}^{\mathrm{post}}_{r}(c_k)
  =
  \frac{1}{N}\sum_{i=1}^{N}
  \mathbb{I}\!\left[
    \bar{\mathbf{x}}_{i,r}^{\mathrm{true}}
    \in
    \mathcal{C}_{i,r}^{(c_k^{\mathrm{cal}})}
  \right].
  \label{eq:test_post_curve}
\end{equation}
Averaging over the held-out runs yields the mean pre- and post-calibration curves
\begin{align*}
  \hat{c}^{\mathrm{pre}}_{\mathrm{test}}(c_k)
  =
  \frac{1}{|\mathcal{R}_{\mathrm{test}}|}
  \sum_{r\in\mathcal{R}_{\mathrm{test}}}
  \hat{c}^{\mathrm{pre}}_{r}(c_k),
  \qquad
  \hat{c}^{\mathrm{post}}_{\mathrm{test}}(c_k)
  =
  \frac{1}{|\mathcal{R}_{\mathrm{test}}|}
  \sum_{r\in\mathcal{R}_{\mathrm{test}}}
  \hat{c}^{\mathrm{post}}_{r}(c_k),
\end{align*}
which summarize calibration behavior on unseen runs before and after isotonic recalibration.

The scalar summary metrics used to quantify these calibration curves are introduced later in the evaluation section, together with the remaining source localization and uncertainty metrics.

\section{Experimental Setup}
\label{sec:numerical_simulations}

\begin{figure}[!htbp]
    \centering
    \includegraphics[width=\textwidth]{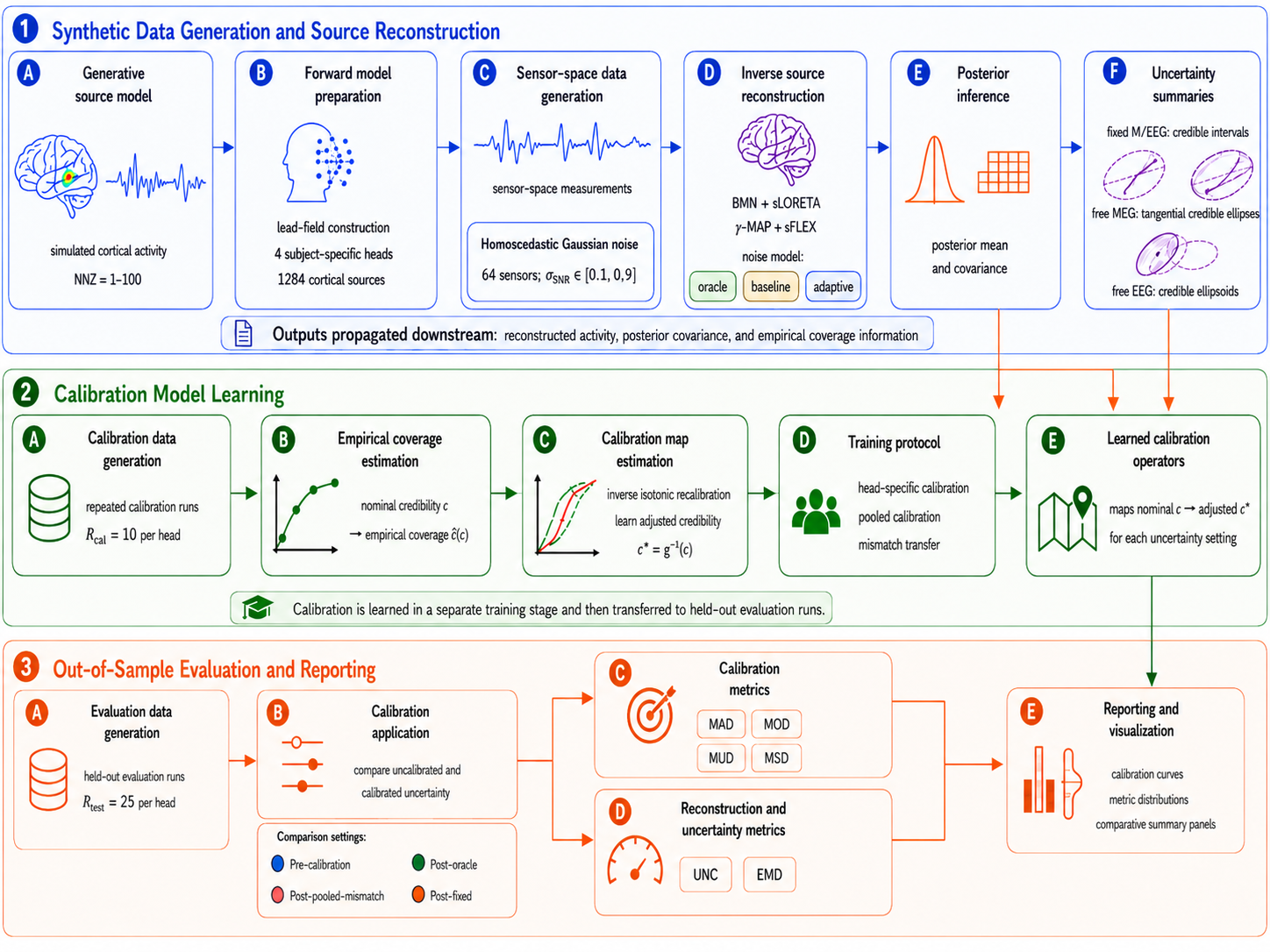}
    \caption{%
    Unified workflow for simulation-based source reconstruction, posterior uncertainty estimation, calibration learning, and out-of-sample evaluation. In Stage~1, synthetic cortical activity is generated, the forward model is prepared, and sensor-space measurements are simulated under homoscedastic Gaussian noise. Source reconstruction is performed using BMN + sLORETA and $\gamma$-MAP + S-FLEX under oracle, baseline, and adaptive noise-learning models. Posterior means and covariances are used to construct credible intervals for fixed M/EEG, credible ellipses for free MEG, and credible ellipsoids for free EEG. In Stage~2, calibration operators are learned from repeated calibration runs using empirical coverage estimation and inverse isotonic recalibration. In Stage~3, the learned calibration is applied to held-out evaluation runs. Calibration metrics (MAD, MOD, MUD, and MSD) and reconstruction and uncertainty metrics (UNC and EMD) are computed in parallel and summarized through comparative visualizations.
    }
    \label{fig:unified_calibration_workflow}
\end{figure}

This section presents the experimental protocol used to benchmark the considered inverse solvers, as summarized in Figure~\ref{fig:unified_calibration_workflow}. We first describe the generation of synthetic datasets, followed by the noise specifications and evaluation criteria used in the simulation study.

\subsection{Synthetic data generation}
\label{subsec:synthetic_data_generation}

Synthetic datasets are generated from sparse source
activity, forward projection to sensor space, and
additive noise obtained by rescaling Gaussian white-noise
realizations. The simulation framework supports fixed-
and free-orientation configurations for EEG and MEG,
including reduced two-dimensional MEG and full Cartesian
three-dimensional EEG representations. The numerical
simulation results presented here are based on MEG.
Details of the head model, channel selection, and
lead-field subset construction are provided in 
Section~\ref{subsec:experimental_settings_parameter_grids}.

The simulated source time courses are designed to resemble event-related potentials (ERPs) in EEG and event-related fields (ERFs) in MEG. Each epoch comprises pre-stimulus and post-stimulus intervals, with post-stimulus event-related activity
defined relative to the pre-stimulus baseline
\citep{Luck2014}.

For each simulation run, active source locations are
sampled uniformly without replacement. At each selected
location, a scalar ERP/ERF-like waveform $s_i(t)$ is
generated independently. The simulator selects an interval
within the post-stimulus period, generates white noise
over that interval, and applies a fourth-order Butterworth band-pass filter over the prescribed frequency band. The filtered segment is multiplied by a Hanning window, normalized by its peak absolute value, and rescaled by an amplitude drawn from a clipped log-normal distribution.
The segment is inserted into the full epoch, with zero
activity outside its selected interval. All inactive source locations are assigned zero activity.

Source coefficients are expressed in the orientation
coordinates defined in Supplementary
Section~\ref{sec:source_orientation_leadfields}.
Using the prescribed unit cortical normal $\mathbf n_i$
and the retained local MEG basis $\mathbf Q_i$ defined
there, the waveform at each active source is represented as
\[
  \begin{aligned}
    x_i(t)
    &=s_i(t),
    &&\text{fixed orientation }(d=1),\\
    \mathbf x_i(t)
    &=\mathbf Q_i^\top\mathbf n_i\,s_i(t),
    &&\text{reduced free-orientation MEG }(d=2),\\
        \mathbf x_i(t)
    &=\mathbf n_i s_i(t),
    &&\text{free-orientation EEG }(d=3).
  \end{aligned}
\]
Thus, the simulated activity at each location follows
a single waveform with a time-invariant orientation.
For reduced free-orientation MEG, this construction
is equivalent to projecting the three-dimensional
cortical-normal source into the retained local basis:
\[
  \mathbf Q_i^\top
  \bigl(\mathbf n_i s_i(t)\bigr)
  =
  \bigl(\mathbf Q_i^\top\mathbf n_i\bigr)s_i(t).
\]

Stacking the source coefficients over locations and time
gives $\mathbf X\in\mathbb R^{dN\times T}$. The noise-free
sensor data are obtained from the forward model in
Eq.~\eqref{eq:forward_model_general}:
\[
  \mathbf Y_{\mathrm{sig}}=\mathbf L\mathbf X,
  \qquad
  \mathbf L\in\mathbb R^{M\times dN},
\]
where the lead field and source coefficients use the
same orientation representation.

The observed sensor data are
\[
  \mathbf Y=\mathbf Y_{\mathrm{sig}}+\eta\mathbf E,
\]
where $\mathbf E\in\mathbb R^{M\times T}$ has independent
standard normal entries. For
$\alpha_{\mathrm{SNR}}\in(0,1)$, the noise scale is
\[
  \eta
  =
  \frac{1-\alpha_{\mathrm{SNR}}}{\alpha_{\mathrm{SNR}}}
  \frac{\|\mathbf Y_{\mathrm{sig}}\|_F}{\|\mathbf E\|_F}.
\]
This choice fixes the realized signal-to-noise ratio through
\[
  \alpha_{\mathrm{SNR}}
  =
  \frac{\|\mathbf Y_{\mathrm{sig}}\|_F}
  {\|\mathbf Y_{\mathrm{sig}}\|_F+\|\eta\mathbf E\|_F},
  \qquad
  \mathrm{SNR}_{\mathrm{lin}}
  =
  \frac{\|\mathbf Y_{\mathrm{sig}}\|_F^2}
  {\|\eta\mathbf E\|_F^2}
  =
  \left(
    \frac{\alpha_{\mathrm{SNR}}}
    {1-\alpha_{\mathrm{SNR}}}
  \right)^2.
\]
Larger values of $\alpha_{\mathrm{SNR}}$ therefore
correspond to cleaner sensor data. The boundary cases
$\alpha_{\mathrm{SNR}}=1$ and $\alpha_{\mathrm{SNR}}=0$
are handled separately, yielding noise-free and noise-only observations, respectively.

The scale \(\eta\) is determined separately for each
simulation run. In the oracle-noise setting, the same
run-specific scale is supplied to the inverse solver as
its assumed noise level. The baseline and adaptive
noise specifications are described in
Section~\ref{subsec:model_selection}.

For each run, the ground-truth source coefficients are
also temporally averaged using the same aggregation
applied to the posterior summaries. These quantities
provide the reference values for evaluating credible-region coverage and recalibration performance.

\subsection{Noise Specification}
\label{subsec:model_selection}

For a fixed source reconstruction method, posterior uncertainty depends not only on the source prior but also on how the sensor-noise covariance is specified. In the present study, this aspect is treated as part of the experimental design. We therefore consider
three noise-specification settings: \emph{oracle noise}, \emph{baseline noise}, and \emph{adaptive noise learning}. The oracle setting uses ground-truth information from the simulation, whereas baseline estimation and adaptive noise learning are applicable to real ERP/ERF data. In the oracle and baseline settings, the noise covariance is specified before inversion. In the adaptive setting, the noise parameters are estimated jointly with the source hyperparameters using the Type-II procedure described in Section~\ref{subsec:inverse_solvers}.

In the synthetic framework described in
Section~\ref{subsec:synthetic_data_generation}, sensor noise is generated as additive Gaussian white noise followed by \(\alpha_{\mathrm{SNR}}\)-dependent scaling. In each run, the simulated noise is \(\eta\mathbf E\), with
\(\eta\) defined in
Section~\ref{subsec:synthetic_data_generation}.
Because \(\eta\) depends on the realized \(\mathbf E\),
the scaled noise is not exactly Gaussian. We use
\(\eta^2\) as the run-specific nominal variance in the
Gaussian inverse model.

\subsubsection{Oracle noise}
\label{subsec:oracle_noise}

In the oracle setting, the inverse solver is supplied
with the same run-specific scale \(\eta\) used to generate
the sensor data. Its assumed noise covariance is
\[
  \boldsymbol{\Sigma}_{\mathbf e}^{\mathrm{oracle}}
  =
  \eta^2\mathbf I_M.
\]
This setting uses the known simulation scale; baseline
noise instead estimates a variance from the pre-stimulus
data, whereas adaptive noise learning updates its noise
parameter jointly with the source parameters.

\subsubsection{Baseline noise}
\label{subsec:baseline_noise}

In the baseline setting, the sensor-noise variance is estimated from the pre-stimulus portion of the noisy sensor data. Let
\[
\mathbf Y_{\mathrm{base}}
\in
\mathbb R^{M\times T_{\mathrm{base}}}
\]
denote the restriction of the sensor data to the pre-stimulus interval. Since the simulated source activity is zero during this interval, the baseline segment contains only sensor noise.

For each sensor \(m=1,\dots,M\), we compute the sample variance
\(\hat{\sigma}_m^2\) across the pre-stimulus time points. The scalar baseline variance is then defined by
\[
\hat{\lambda}_{\mathrm{base}}
=
\frac{1}{M}
\sum_{m=1}^{M}
\hat{\sigma}_m^2,
\]
and the corresponding noise covariance supplied to the inverse solver is
\[
\boldsymbol{\Sigma}_{\mathbf e}^{\mathrm{base}}
=
\hat{\lambda}_{\mathrm{base}}\mathbf I_M.
\]
Unlike the oracle setting, the baseline specification does not use the true simulation noise variance. It therefore represents a data-driven noise estimate obtained independently of the post-stimulus source reconstruction.

\subsubsection{Adaptive noise learning}

In the adaptive setting, the sensor-noise variance is estimated jointly
with the source hyperparameters from the observed data. Consistently with
the homoscedastic noise model used throughout the repeated benchmark, both
inverse methods assume
\[
\boldsymbol{\Sigma}_e=\lambda \mathbf{I}_M,
\]
with a single common sensor-noise variance $\lambda>0$.

For $\gamma$-MAP+S-FLEX, the grouped source-variance hyperparameters
$\{\gamma_i\}_{i=1}^{N}$ and the common noise variance $\lambda$ are updated
alternately using the Type-II convex-bounding procedure described in
Section~2.1.3. For BMN+sLORETA, the shared source variance
$\gamma$ and the same type of common sensor-noise variance $\lambda$ are
learned jointly. The corresponding Type-II objectives and update rules are
given in Section~2.1.3.

\subsection{Evaluation Metrics}
\label{subsec:evaluation_metrics}

We evaluate the considered source reconstruction methods from three complementary perspectives: source localization accuracy, uncertainty magnitude, and uncertainty calibration. Since the uncertainty analysis in this work is based on temporally aggregated posterior summaries, the uncertainty-related metrics below are formulated in terms of aggregated source quantities.

\subsubsection{Source localization metric: Earth Mover's Distance}
\label{subsec:emd_metric}

To assess spatial localization accuracy while accounting for the geometry of the cortical source space, we employ the Earth Mover's Distance (EMD)~\citep{Rubner2000,Haufe2008}. The EMD measures the minimum transport cost required to transform the reconstructed source distribution into the ground-truth source distribution, with the spatial distances between source locations defining the transport cost.

\textcolor{red}{For all orientation settings, the nonnegative source mass at location
\(i=1,\dots,N\) is defined as the Euclidean norm of the temporally aggregated local source block,}
\begin{equation}
\textcolor{red}{
  x_i^{\mathrm{true}}
  =
  \left\|
    \bar{\mathbf{x}}_i^{\mathrm{true}}
  \right\|_2,
  \qquad
  \hat{x}_i
  =
  \left\|
    \bar{\boldsymbol{\mu}}_i
  \right\|_2.
}
\label{eq:emd_source_mass}
\end{equation}
\textcolor{red}{This definition applies uniformly to \(d\in\{1,2,3\}\); for
\(d=1\), the Euclidean norm reduces to the absolute value. In the reduced
free-orientation MEG setting, \(\mathbf{Q}_i\in\mathbb{R}^{3\times2}\) contains the two
dominant right singular vectors obtained from the SVD of the local lead-field block and
therefore satisfies
\(\mathbf{Q}_i^\top\mathbf{Q}_i=\mathbf{I}_2\). Consequently,
\(\|\mathbf{Q}_i\mathbf{a}_i\|_2=\|\mathbf{a}_i\|_2\), so the source magnitude is identical
in the reduced two-dimensional coordinates and in the corresponding representation in the
retained three-dimensional source subspace.}

To compare spatial distributions rather than total mass, these quantities are normalized as
\[
  \tilde{x}_i^{\mathrm{true}}
  =
  \frac{
    x_i^{\mathrm{true}}
  }{
    \sum\limits_{j=1}^{N}
    x_j^{\mathrm{true}}
  },
  \qquad
  \tilde{\hat{x}}_i
  =
  \frac{
    \hat{x}_i
  }{
    \sum\limits_{j=1}^{N}
    \hat{x}_j
  }.
\]

Let \(\mathbf{p}_i\in\mathbb{R}^{3}\) denote the spatial location of source \(i\), and define
the ground-cost matrix by
\[
  M_{ij}
  =
  \left\|
    \mathbf{p}_i-\mathbf{p}_j
  \right\|_2.
\]
The EMD between the ground-truth and reconstructed source distributions is then defined as
\begin{equation}
  \mathrm{EMD}
  \left(
    \tilde{\mathbf{x}}^{\mathrm{true}},
    \tilde{\hat{\mathbf{x}}}
  \right)
  =
  \min_{\mathbf{F}\geq0}
  \sum_{i=1}^{N}
  \sum_{j=1}^{N}
  F_{ij}M_{ij},
  \label{eq:emd_def}
\end{equation}
subject to the mass-conservation constraints
\begin{equation}
  \sum_{j=1}^{N}
  F_{ij}
  =
  \tilde{x}_i^{\mathrm{true}},
  \qquad
  \sum_{i=1}^{N}
  F_{ij}
  =
  \tilde{\hat{x}}_j.
  \label{eq:emd_constraints}
\end{equation}
Here, \(\mathbf{F}=(F_{ij})\) denotes the transport plan. Lower EMD values indicate better spatial agreement between the reconstructed and true source distributions. The optimization problem in Eqs.~\eqref{eq:emd_def}--\eqref{eq:emd_constraints} is solved using the Python Optimal Transport (POT) package~\citep{Flamary2021}.

\subsubsection{Uncertainty magnitude metric: UNC}
\label{subsec:unc_metric}

To quantify the overall magnitude of posterior uncertainty, we compute \(\mathrm{UNC}\), defined as the mean posterior standard deviation across all source locations. This metric summarizes how diffuse or concentrated the posterior uncertainty is over the source space.

In the fixed-orientation case, the aggregated posterior variance at source location \(i\) is the scalar quantity \(\bar{\Sigma}_{ii}\), so the corresponding posterior standard deviation is \(\bar{\sigma}_i=\sqrt{\bar{\Sigma}_{ii}}\). In the free-orientation case, the aggregated local posterior covariance block is \(\bar{\boldsymbol{\Sigma}}_{ii}\in\mathbb{R}^{d\times d}\)
with \(d=2\) for reduced free-orientation MEG and \(d=3\) for free-orientation EEG. We summarize this local uncertainty by the root mean marginal variance
\begin{equation}
  \bar{\sigma}_i
  =
  \sqrt{\frac{1}{d}\,\operatorname{tr}\!\bigl(\bar{\boldsymbol{\Sigma}}_{ii}\bigr)}.
  \label{eq:free_rms}
\end{equation}
Equivalently, since the posterior covariance is static over time, the aggregated covariance is obtained from the pointwise covariance by scaling with \(1/T\).

Using these local standard deviations, \(\mathrm{UNC}\) is defined by
\begin{equation}
  \mathrm{UNC}
  =
  \frac{1}{N}\sum_{i=1}^{N}\bar{\sigma}_i.
  \label{eq:unc_def}
\end{equation}
Larger values of \(\mathrm{UNC}\) indicate greater overall posterior uncertainty.

\subsubsection{Uncertainty calibration metrics}
\label{subsec:calibration_metrics}

Using the calibration curve introduced in Section~\ref{subsec:calibration_curve}, we quantify calibration performance by four scalar summary metrics. \textcolor{red}{For the nominal credibility levels \(c_k\) defined there, let \(\hat{c}(c_k)\) denote the corresponding empirical coverage.}

\paragraph{Mean absolute deviation}
\begin{equation}
  \mathrm{MAD}
  =
  \frac{1}{K}\sum_{k=1}^{K}
  \left|
    \hat{c}(c_k)-c_k
  \right|.
  \label{eq:mad_metric}
\end{equation}
This measures the average magnitude of deviation between empirical and nominal coverage. Lower values indicate better overall calibration.

\paragraph{Mean signed deviation}
\begin{equation}
  \mathrm{MSD}
  =
  \frac{1}{K}\sum_{k=1}^{K}
  \bigl(
    \hat{c}(c_k)-c_k
  \bigr).
  \label{eq:msd_metric}
\end{equation}
This metric captures the average signed deviation of the calibration curve from the diagonal. Positive values indicate a tendency toward empirical overcoverage, whereas negative values indicate a tendency toward empirical undercoverage.

\paragraph{Maximum underconfidence deviation}
In the reported experiments, maximum underconfidence deviation (MUD) quantifies the largest empirical overcoverage, i.e. the largest positive excess of empirical coverage over nominal coverage. We therefore use the positive-part definition
\begin{equation}
  \mathrm{MUD}=
  \max_{1\le k\le K}
  \bigl[\hat{c}(c_k)-c_k\bigr]_+.
  \label{eq:mud_metric}
\end{equation}
Large MUD values indicate the presence of conservative credible regions whose empirical coverage is much higher than intended.

\paragraph{Maximum overconfidence deviation}
Maximum overconfidence deviation (MOD) quantifies the largest empirical undercoverage, i.e. the largest positive excess of nominal coverage over empirical coverage. We use
\begin{equation}
\mathrm{MOD}=
  \max_{1\le k\le K}
  \bigl[c_k-\hat{c}(c_k)\bigr]_+.
  \label{eq:mod_metric}
\end{equation}
Large MOD values indicate the presence of overconfident credible regions whose empirical coverage falls much below the nominal level.

\subsection{Simulation Study Design}
\label{subsec:simulation_study_design}

The simulation study was designed to evaluate source-reconstruction accuracy, posterior uncertainty, and empirical uncertainty calibration under controlled variation of head geometry, signal-to-noise level, source sparsity, source orientation, and noise specification. It comprises two complementary components. First, a repeated benchmark was generated from subject-specific MEG forward models and provided the data for the calibration curves, run-level metric distributions, and parameter-sweep analyses. Second, a
separate fixed-orientation EEG realization was generated for the illustrative source- and time-resolved multipanel described below.

Each repeated simulation followed the same sequence: source generation, forward propagation, sensor-noise addition, Bayesian source reconstruction, posterior uncertainty estimation, post-hoc uncertainty recalibration, and performance evaluation. The default configuration was evaluated for both fixed-orientation MEG and reduced free-orientation MEG. All analyses are reported for the free-orientation source model, whereas fixed-orientation results are provided in the supplement. Fixed- and reduced free-orientation simulations followed the same overall design but were generated independently and were therefore not paired run by run.

Pre- and post-calibration results were always evaluated on the same held-out test runs.  Recalibration maps were estimated independently for every estimator, orientation model, noise specification, uncertainty diagnostic, and calibration strategy.

\subsubsection{Experimental settings and parameter ranges}
\label{subsec:experimental_settings_parameter_grids}

The repeated benchmark used realistic subject-specific volume-conductor models derived from the publicly available CamCAN dataset~\citep{CamCAN}. Forward models were constructed for four subjects from their T1-weighted anatomical images using MNE-Python~\citep{MNE}. A minimum source-to-inner-skull distance of
\(3\)~mm was imposed, and boundary-element method (BEM) forward solutions were computed for both fixed and free dipolar source orientations. Before source-space restriction, each forward model contained 4686 cortical locations and 204 MEG gradiometer channels.

To obtain a computationally controlled benchmark while preserving
subject specific cortical geometry, 642 locations were selected in each
hemisphere, yielding \(N=1284\) cortical locations per subject. This source
subset was sampled once for each head model using the fixed subset seed 0 and was
then held constant across all calibration runs, evaluation runs, estimators,
noise specifications, and uncertainty analyses for that subject. The same
restricted source ordering and cortical coordinates were used for source generation, sensor simulation, inverse estimation, posterior evaluation, and
Earth Mover's Distance (EMD) computation.

Synthetic activity was generated with the ERP/ERF-type source simulator
described above. The simulated interval was
\([t_{\min},t_{\max})=[-0.5,0.5)\)~s, stimulus onset was at \(t=0\), the
sampling frequency was \(f_s=250\)~Hz, and the waveform-generating filter band
was \([f_{\min},f_{\max}]=[1,5]\)~Hz. Source amplitudes followed a clipped
log-normal distribution with median \(20\)~nAm, logarithmic standard deviation
\(0.2\), and support restricted to \([2.5,50]\)~nAm. Additive sensor noise was
generated as Gaussian white noise and scaled according to the prescribed
\(\alpha_{\mathrm{SNR}}\) value.

The general simulation framework supports local source dimensions
\(d\in\{1,2,3\}\), corresponding to fixed orientation, reduced
free-orientation MEG, and free-orientation EEG, respectively.
The benchmark reported here used the fixed-orientation and reduced free-orientation MEG models. 

\paragraph{Run-level randomization and reproducibility}

Each calibration or evaluation run was indexed by a distinct reproducible run
seed. Conditional on the fixed head model and its fixed source subset, the run
seed determined (i) the active source locations, sampled uniformly without
replacement; (ii) the stochastic ERP/ERF waveform at each active source,
including its post-stimulus temporal placement, duration, filtered Gaussian driving
sequence, and amplitude draw; and (iii) the Gaussian sensor-noise realization.
The source and sensor simulators used separate random-number-generator
instances initialized reproducibly from the run index. Calibration and
evaluation sets used disjoint collections of run seeds. Within one run and one
orientation setting, the same source realization, forward model, and noisy
sensor data were supplied to both inverse-estimator families and to the three
noise-specification analyses. Therefore, estimator and noise-model comparisons
within a run were based on identical simulated data, whereas fixed- and
reduced free-orientation experiments used independent realizations.

Source variance hyperparameters were initialized to one, and, in the adaptive noise-learning
variants, the noise variance was initialized to one in the normalized
optimization scale. Hence, the run-to-run variability reported in the figures
originates from the simulated source support, source waveforms, amplitudes,
noise realizations, and head geometry rather than from random solver
initialization.

\paragraph{Inverse-solver and spatial-basis settings}
Two estimator families were compared: BMN with sLORETA depth compensation and
\(\gamma\)-MAP with S-FLEX integration. Each estimator was evaluated with
adaptive noise learning and with a pre-specified variance under the oracle and
baseline noise settings. In the repeated benchmark, the Gaussian S-FLEX kernel
used \(\sigma_B = 20~\mathrm{mm}\). The maximum iteration counts used in the repeated
benchmark were 300 for the BMN+sLORETA and 1000 for the
\(\gamma\)-MAP+S-FLEX solvers, with convergence tolerance \(10^{-6}\).

The principal experimental factors were the head model, the number of active
sources, the signal-to-noise parameter, and the run seed. The parameter grids
were
\[
  \alpha_{\mathrm{SNR}}
  \in
  \{0.1,0.3,0.5,0.7,0.9\},
  \qquad
  \mathrm{NNZ}
  \in
  \{1,3,5,10,100\},
\]
with default values \(\alpha_{\mathrm{SNR}}=0.5\) and
\(\mathrm{NNZ}=5\). For each of the \(H=4\) head models,
\(R_{\mathrm{cal}}=10\) independent simulations were reserved for learning a
recalibration map and \(R_{\mathrm{test}}=25\) independent simulations were
reserved for final evaluation. Thus, every curve or box distribution reported
at one fixed evaluation condition contains
\(H \times R_{\mathrm{test}}=4\times25=100\) run-level observations.

\paragraph{Condition-specific evaluation and calibration training}
Calibration curves at the default configuration used
\(\alpha_{\mathrm{SNR}}=0.5\) and \(\mathrm{NNZ}=5\). In the SNR sweep,
\(\alpha_{\mathrm{SNR}}\) was varied while \(\mathrm{NNZ}=5\) was held fixed;
in the source-number sweep, \(\mathrm{NNZ}\) was varied while
\(\alpha_{\mathrm{SNR}}=0.5\) was held fixed. Evaluation summaries were formed
separately at every displayed parameter value. 

\paragraph{Calibration regimes}
The calibration regimes differed exclusively in the calibration simulations used
to estimate ("train") the inverse-isotonic credibility map. For every combination of
inverse estimator, noise specification, orientation model, uncertainty
geometry, and calibration strategy, a separate map was learned from the
designated calibration runs. 
Calibration and evaluation runs were disjoint, and every mode was evaluated
on the same \(R_{\mathrm{test}}=25\) held-out runs per head, yielding
\(4\times25=100\) run-level observations at each fixed test condition.

\textit{Pre-calibration.}
In this mode, no isotonic model was fitted. Credible intervals or ellipses were constructed
directly at the requested nominal credibility level \(c\) from the raw
posterior mean and covariance. It therefore required no calibration
runs and served as the uncorrected reference.

\textit{Post-oracle.}
For each head model and each test condition, one head-specific isotonic map was
fitted using \(R_{\mathrm{cal}}=10\) independent calibration runs from the same
head, the same \(\alpha_{\mathrm{SNR}}\), and the same \(\mathrm{NNZ}\) value
as the evaluation runs. At the default setting this means same-head training
at \(\alpha_{\mathrm{SNR}}=0.5\) and \(\mathrm{NNZ}=5\); in the SNR and NNZ
sweeps, a new matched map was learned at every displayed parameter value. The
map was then applied to 25 held-out runs from that head. This mode represents
the best-case post-hoc recalibration condition because both anatomy and
data-generating condition are matched.


\textit{Post-pooled-mismatch.}
For each held-out test head, one isotonic map was learned from the other three
head models using \(3\times5\times5\times10=750\) calibration runs pooled across the complete \(\alpha_{\mathrm{SNR}}\)-by-\(\mathrm{NNZ}\) grid. The map was then applied to the 25 held-out runs of the excluded head. This strategy jointly tests transfer across experimental conditions and across subject-specific head
geometry. It is denoted by "post-mismatch" in the retained figure legends.

\textit{Post-fixed.}
For each head model, one isotonic map was learned from the 10 same-head
calibration runs at the default configuration
\(\alpha_{\mathrm{SNR}}=0.5\), \(\mathrm{NNZ}=5\), and was then kept fixed
when evaluating every value in the SNR or NNZ sweep. Consequently,
post-fixed coincides with post-oracle at the default value and quantifies
condition mismatch away from that value: SNR transfer in the 
\(\alpha_{\mathrm{SNR}}\) sweep and sparsity transfer in the
\(\mathrm{NNZ}\) sweep.

\paragraph{Illustrative fixed-orientation EEG realization}
A separate simulation was constructed for the representative
source-level multipanel reported with the simulation results.
Unlike the CamCAN-based MEG benchmark, this realization used
the MNE \texttt{fsaverage} anatomy, a cortical source space based on third-order icosahedral subdivision (\texttt{ico3})
with 642 locations per hemisphere, a three-layer EEG
boundary-element model, and a 64-channel BioSemi
montage~\citep{MNE}. Average referencing reduced the
effective inverse sensor dimension from 64 to 63.

The source model was fixed-normal and contained two bilateral auditory generators, one in each hemisphere, with \(\mathrm{NNZ}=2\) and
\(\alpha_{\mathrm{SNR}}=0.5\). A single reproducible realization was generated
using source and noise seeds equal to 42. Both BMN+sLORETA and \(\gamma\)-MAP+S-FLEX were fitted with adaptive common-noise learning. For this
illustrative realization, \(\beta_{\mathrm{rel}}=0.1\), the S-FLEX kernel width
was \(\sigma_B=2.138\)~mm, and the post-convergence relative \(\gamma\)-threshold was \(0.2\). This kernel width is specific to the illustrative \texttt{fsaverage} ico3 experiment in analogy  to the \(20\)-mm width used in the repeated CamCAN benchmark.

\subsection{Real-data study design}
\label{subsec:realdata_design}

The real-data study used the Brainstorm CTF auditory MEG dataset distributed through MNE-Python
\citep{MNEBrainstormAuditoryTutorial,Tadel2011,Gramfort2013,MNE}.
The dataset comprises two auditory acquisition runs from one participant,
recorded with a CTF 275 system at a sampling rate of 2400~Hz. The recordings contain 274 MEG axial gradiometer channels and
26 MEG reference channels. Each run included frequent standard tones and less frequent deviant tones presented at variable interstimulus intervals. We analysed the standard condition to examine the posterior source structure around the M100 (N100m), the auditory evoked-field
component occurring approximately 100~ms after stimulus onset and
corresponding to the auditory N100 response in EEG. This component
provides a well-characterized auditory cortical response for source
analysis without requiring a standard-deviant contrast.

Data preparation followed the MNE-Python implementation of the Brainstorm auditory workflow~\citep{MNEBrainstormAuditoryTutorial}. The two CTF recordings were concatenated, and the temporal offset between the digital trigger and the acoustic stimulus was corrected from the recorded audio channel. Previously identified bad segments and noisy channels were excluded, and signal-space projection (SSP) was applied to attenuate ocular and saccade-related artifacts.
Epochs were extracted from \(-100\) to \(500\)~ms and baseline-corrected with respect to the pre-stimulus interval. Following the reference workflow, the first 40 accepted standard epochs from each acquisition run were combined, giving an evoked response based on 80 trials. The data were resampled to 600~Hz and low-pass filtered at 40~Hz. The subsequent inverse analysis was restricted to the interval from \(-100\) to \(300\)~ms.

The MNE-Python workflow~\citep{MNEBrainstormAuditoryTutorial} was retained through the construction of the sensor-level evoked response, the empty-room noise covariance, the subject-specific anatomical model, and the cortical forward operator. At the source-estimation stage, however, the two Bayesian inverse methods investigated in this study, BMN+sLORETA and \(\gamma\)-MAP+S-FLEX, replaced the reference inverse estimator. Both methods were applied to the same measured response, cortical geometry, and forward operator. The covariance estimated from the separate empty-room recording was used as a common input for sensor whitening in both methods. 

The subject-specific free-orientation forward solution supplied with the dataset was restricted from its original oct6 source space to the nested oct5 grid, resulting in \(N=2052\) cortical locations. After application of the two signal-space projection (SSP) projectors, the effective rank of the whitened sensor model was \(M_r=268\). At each cortical location, the whitened three-component lead-field block was reduced to its two dominant MEG-sensitive components by the source-wise singular-value decomposition method. The inverse model consequently contained \(2N=4104\) reduced source coefficients. BMN+sLORETA and \(\gamma\)-MAP+S-FLEX used the same reduced lead field, source ordering, whitened sensor response, and fixed covariance model.

For BMN, depth-dependent sensitivity differences were compensated using the sLORETA normalization with \(\beta_{\mathrm{rel}}=0.1\). 
For \(\gamma\)-MAP+S-FLEX, the Gaussian width was
chosen relative to the cortical sampling density as
\(\sigma_B=0.30\,\widetilde d_{\mathrm{NN}}=1.832\)~mm, where
\(\widetilde d_{\mathrm{NN}}\) denotes the median within-hemisphere
nearest-neighbor distance. The \(\gamma\)-MAP+S-FLEX hyperparameters were initially estimated over the complete cortical grid. After full-grid convergence, coefficient groups with
\(\gamma_i/\max_j\gamma_j<0.02\) were removed to exclude
negligible basis components from the final source
reconstruction. The posterior mean and covariance used
for the cortical maps and directional credible intervals
were computed from the retained groups.


For cortical visualization, the activity associated with source \(i\) was summarized over the M100 response interval
\(\mathcal T_a=[80,120]\)~ms using the root-mean-square amplitude of the local two-dimensional posterior mean,
\begin{equation*}
  a_i
  =
  \left(
    \frac{1}{|\mathcal T_a|}
    \sum_{t\in\mathcal T_a}
    \left\|
      \boldsymbol\mu_i(t)
    \right\|_2^2
  \right)^{1/2}.
\end{equation*}
Mesh local maxima were identified from the complete cortical activity map, and
the strongest maximum was selected independently in each hemisphere. The same
cortical index was used to extract the local posterior mean, the corresponding
\(2\times2\) covariance block, and the directional posterior interval.

The directional posterior at each selected source was defined using the SVD-based coordinate introduced in the Methods. Specifically, the uncentered local posterior-mean trajectory over the prespecified reference interval
\(\mathcal T_d=[70,120]\)~ms was decomposed by singular-value decomposition,
and the first left singular vector was adopted as the dominant local
orientation. The direction was estimated once and held fixed over the complete
epoch, so that the resulting scalar posterior time course represents temporal
variation along a common source-specific coordinate.

Because the leading singular vector is identifiable only up to sign, its
polarity was chosen so that the projected posterior mean at \(100\)~ms was
non-negative. This deterministic convention affects only the displayed sign;
it does not alter the projected posterior variance, interval width, or zero-containment decision.

\section{Results}

\subsection{Results on Simulated Data}
\label{subsec:results-simulated-data-final}

The simulation study jointly evaluates spatial reconstruction, the magnitude of posterior uncertainty, and the empirical calibration of posterior credible regions.

\subsubsection{Default-setting calibration under adaptive noise learning}
\label{subsubsec:main-free-calibration}

\begin{figure}[!t]
  \centering
  \includegraphics[width=0.98\linewidth]{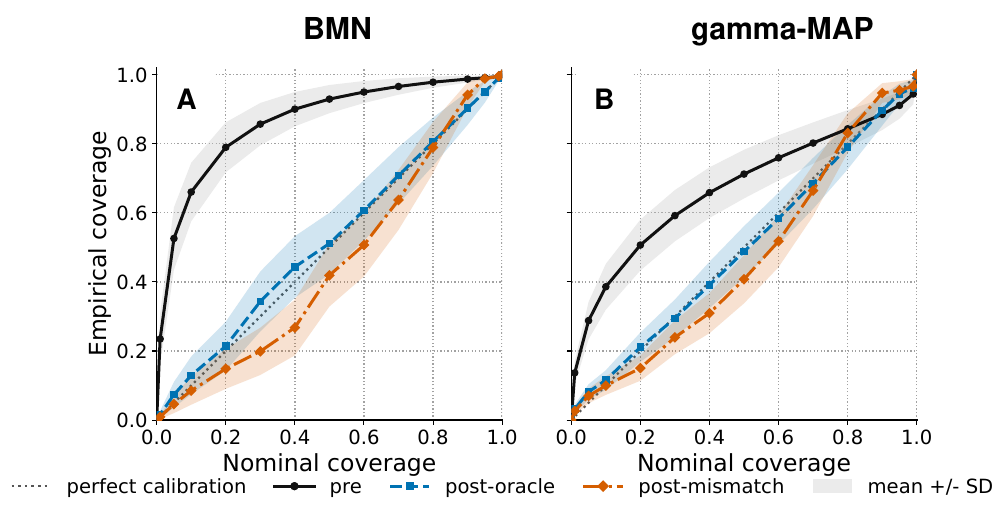}
  
\caption{Calibration curves derived from reduced free-MEG joint two-dimensional credible ellipses under adaptive noise learning, at $\alpha_{\mathrm{SNR}}=0.5$ and $\mathrm{NNZ}=5$. \textbf{A}: BMN+sLORETA. \textbf{B}: $\gamma$-MAP+S-FLEX. Lines show mean empirical coverage over 100 evaluation runs; color-matched bands show plus or minus one run-level standard deviation. The dotted diagonal indicates ideal calibration (empirical coverage equal to nominal credibility). Pre denotes the raw results; post-oracle uses calibration simulations matched in head model, SNR, and NNZ; post-mismatch uses pooled-mismatch calibration simulations.}
  \label{fig:main-free-ellipse-calibration}
\end{figure}

Figure~\ref{fig:main-free-ellipse-calibration} shows the calibration curves derived from joint credible-ellipse containment in the default setting $\alpha_{\mathrm{SNR}}=0.5$ and $\mathrm{NNZ}=5$. In panel \textbf{A}, the BMN+sLORETA curve lies above the diagonal over most nominal levels. Its local regions therefore contain the simulated current vector more often than their nominal credibility would suggest. Post-oracle recalibration substantially reduces this conservative deviation. Post-mismatch recalibration also removes much of the raw overcoverage but overshoots the correction at intermediate credibility levels, resulting in undercoverage.

For $\gamma$-MAP+S-FLEX, panel \textbf{B} shows a smaller raw deviation over much of the range, with overcoverage at some nominal levels and undercoverage at others. Both post-oracle and post-mismatch recalibration bring the curve closer to the diagonal, although residual curvature remains. Because the raw curve crosses the identity line, its miscalibration cannot be described by a single global inflation or contraction of posterior uncertainty. Since MAD does not distinguish overcoverage from undercoverage, we also report MSD, MOD, and MUD.

\subsubsection{Sensitivity to signal quality and source number}
\label{subsubsec:free-meg-sweeps}

\begin{figure}[!htbp]
  \centering
  \includegraphics[width=0.98\linewidth,height=0.73\textheight,keepaspectratio]{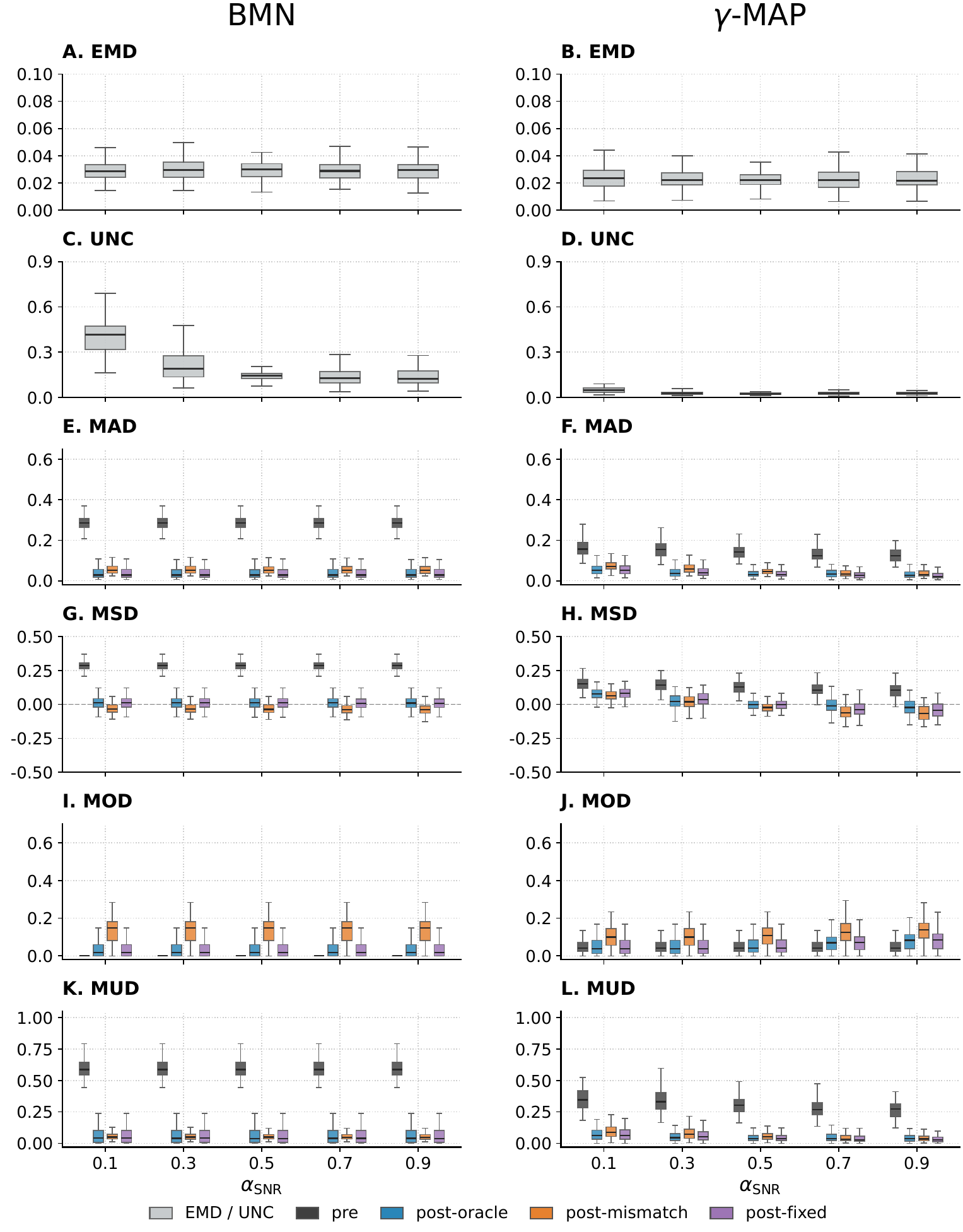}
  \caption{Reduced free-MEG SNR sweep with adaptive noise learning and $\mathrm{NNZ}=5$. The sweep uses the same benchmark and inverse-solver settings as the default configuration. The left column shows BMN+sLORETA and the right column $\gamma$-MAP+S-FLEX results. Panel pairs \textbf{A--B}, \textbf{C--D}, \textbf{E--F}, \textbf{G--H}, \textbf{I--J}, and \textbf{K--L} show EMD, UNC, MAD, MSD, MOD, and MUD, respectively. Each box contains 100 run-level observations at one $\alpha_{\mathrm{SNR}}$ value. Calibration regimes are pre, post-oracle, post-mismatch, and post-fixed. Post-fixed and post-oracle coincide at the default condition $\alpha_{\mathrm{SNR}}=0.5$. }
\label{fig:main-free-snr-adaptive}
\end{figure}

\begin{figure}[!htbp]
  \centering
  \includegraphics[width=0.98\linewidth,height=0.73\textheight,keepaspectratio]{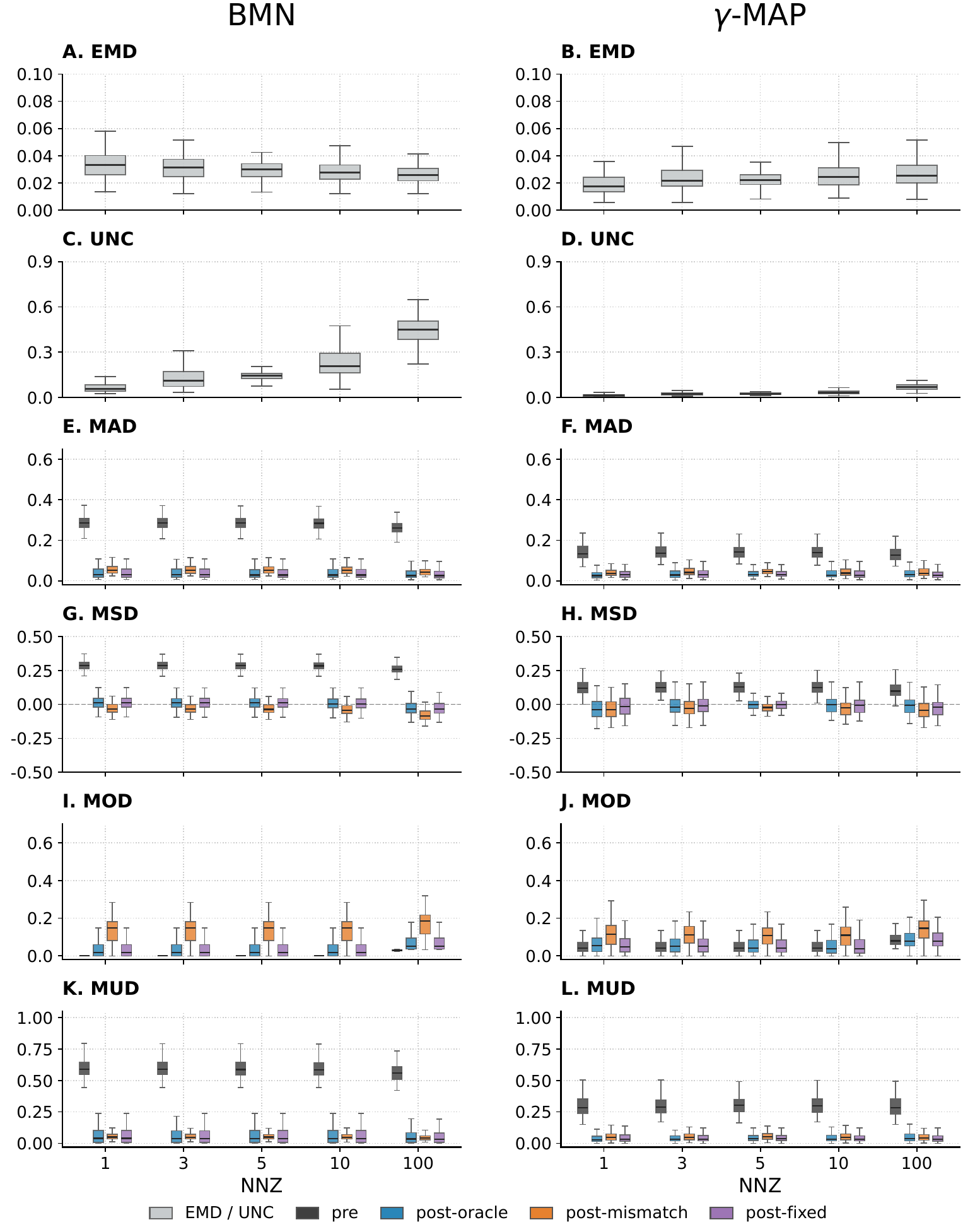}
  \caption{Reduced free-MEG source-number sweep with adaptive noise learning and $\alpha_{\mathrm{SNR}}=0.5$. The sweep uses the same benchmark and inverse-solver settings as the default configuration. The column arrangement and panel pairs \textbf{A--L} match
Figure~\ref{fig:main-free-snr-adaptive}. NNZ takes values $1,3,5,10,100$ and denotes the number of active sources. Each box contains 100 evaluation runs at one NNZ value. Calibration regimes are pre, post-oracle, post-mismatch, and post-fixed. Post-fixed and post-oracle coincide at the default condition $\mathrm{NNZ}=5$. }
  \label{fig:main-free-nnz-adaptive}
\end{figure}

Figures~\ref{fig:main-free-snr-adaptive} and
\ref{fig:main-free-nnz-adaptive} examine reduced free-MEG sensitivity to SNR and source number, respectively. Both sweeps use the default-setting benchmark, with 100 held-out evaluation runs per condition.

At the default condition, the median BMN+sLORETA MAD decreases from $0.285$ (pre) to $0.029$ (post-oracle) and $0.052$ (post-mismatch); the post-oracle median for $\gamma$-MAP+S-FLEX is $0.031$.

In Figure~\ref{fig:main-free-snr-adaptive}, posterior uncertainty decreases as the sensor data become cleaner, with the strongest contraction observed for BMN+sLORETA. EMD changes only modestly across the SNR range for both estimators. BMN+sLORETA retains predominantly conservative raw coverage throughout the sweep, while post-oracle recalibration substantially reduces the calibration error at all evaluated SNR levels.

For $\gamma$-MAP+S-FLEX, the raw calibration error also decreases toward the cleaner-data regimes, although mixed signed deviations remain. Post-oracle recalibration consistently reduces MAD and the dominant coverage deviations. The post-fixed and post-oracle corrections remain similar over much of the SNR range, whereas post-mismatch leaves larger residual deviations and can introduce undercoverage at some SNR levels.

Figure~\ref{fig:main-free-nnz-adaptive} shows that increasing the number of active generators affects the two estimators differently. BMN+sLORETA EMD decreases moderately with increasing NNZ, whereas $\gamma$-MAP+S-FLEX shows a modest increase. UNC increases for both methods as the number of active generators grows, with a substantially larger absolute uncertainty scale for BMN+sLORETA. 

BMN+sLORETA remains predominantly conservative before recalibration throughout the NNZ grid, while post-oracle recalibration substantially reduces this bias. $\gamma$-MAP+S-FLEX retains smaller raw calibration errors with both positive and negative signed deviations, and recalibration further reduces their magnitude. The post-fixed and post-oracle corrections remain similar over much of the NNZ grid, whereas post-mismatch shows larger residual deviations.

\subsubsection{Noise-specification comparisons in reduced free-MEG}
\label{subsubsec:main-noise-comparison}

Supplementary Figure~\ref{fig:supp-free-oracle-calibration} shows the oracle-noise calibration curves at the default setting. The curves closely resemble those obtained with adaptive noise learning for both estimators. In both settings, BMN+sLORETA shows predominantly conservative raw coverage, whereas $\gamma$-MAP+S-FLEX exhibits smaller, mixed-sign deviations; post-oracle recalibration improves coverage for both estimators.

\begin{figure}[!htbp]
  \centering
  \includegraphics[width=0.98\linewidth,height=0.73\textheight,keepaspectratio]{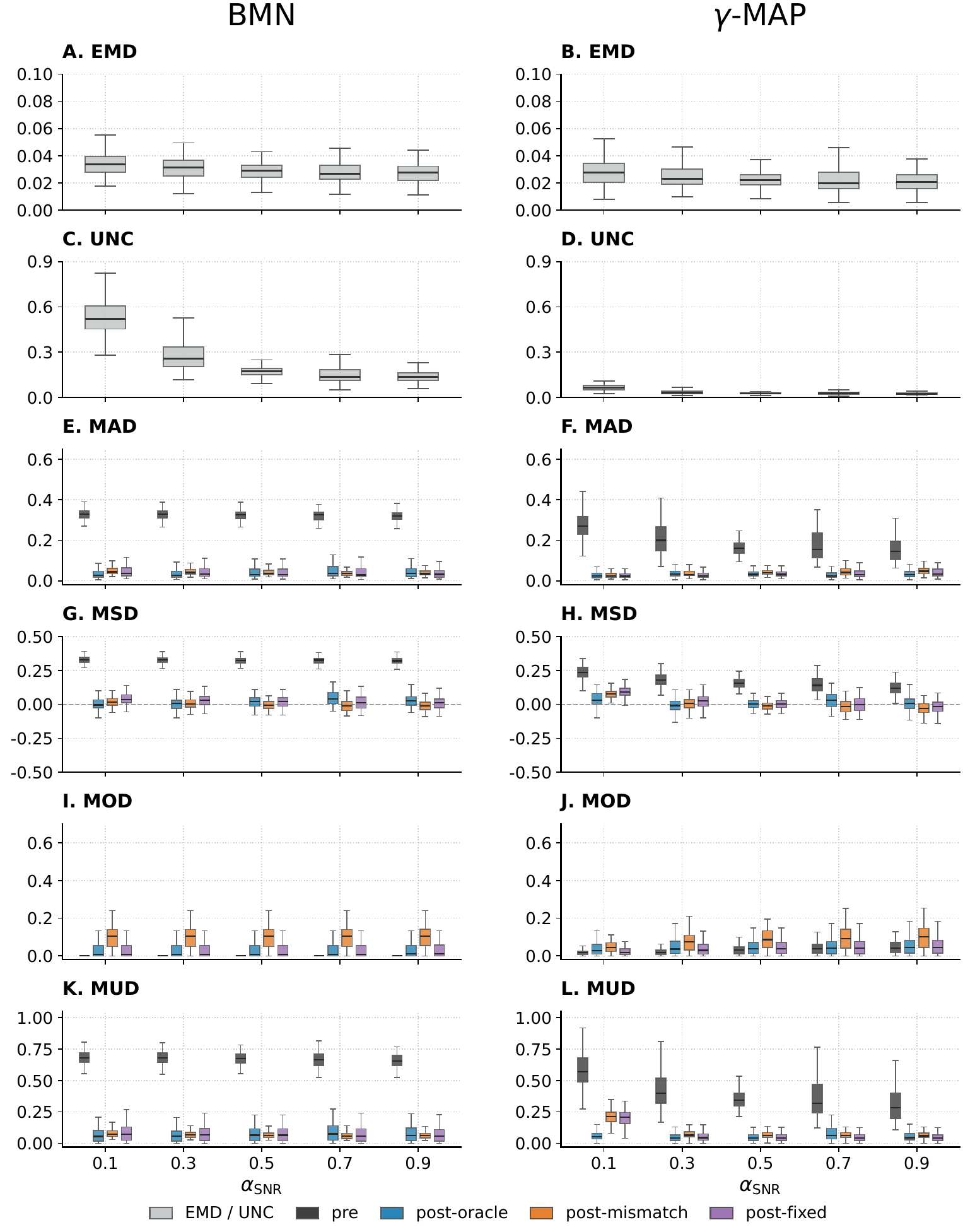}
  
\caption{Reduced free-MEG SNR sweep under oracle noise, varying $\alpha_{\mathrm{SNR}}\in\{0.1,0.3,0.5,0.7,0.9\}$ with $\mathrm{NNZ}=5$. Panel pairs \textbf{A--B}, \textbf{C--D}, \textbf{E--F}, \textbf{G--H}, \textbf{I--J}, and \textbf{K--L} show EMD, UNC, MAD, MSD, MOD, and MUD, respectively, for BMN+sLORETA (left) and $\gamma$-MAP+S-FLEX (right). Calibration modes are pre, post-oracle, post-mismatch, and post-fixed.}
  \label{fig:supp-free-snr-oracle}
\end{figure}

\begin{figure}[!htbp]
  \centering
  \includegraphics[width=0.98\linewidth,height=0.73\textheight,keepaspectratio]{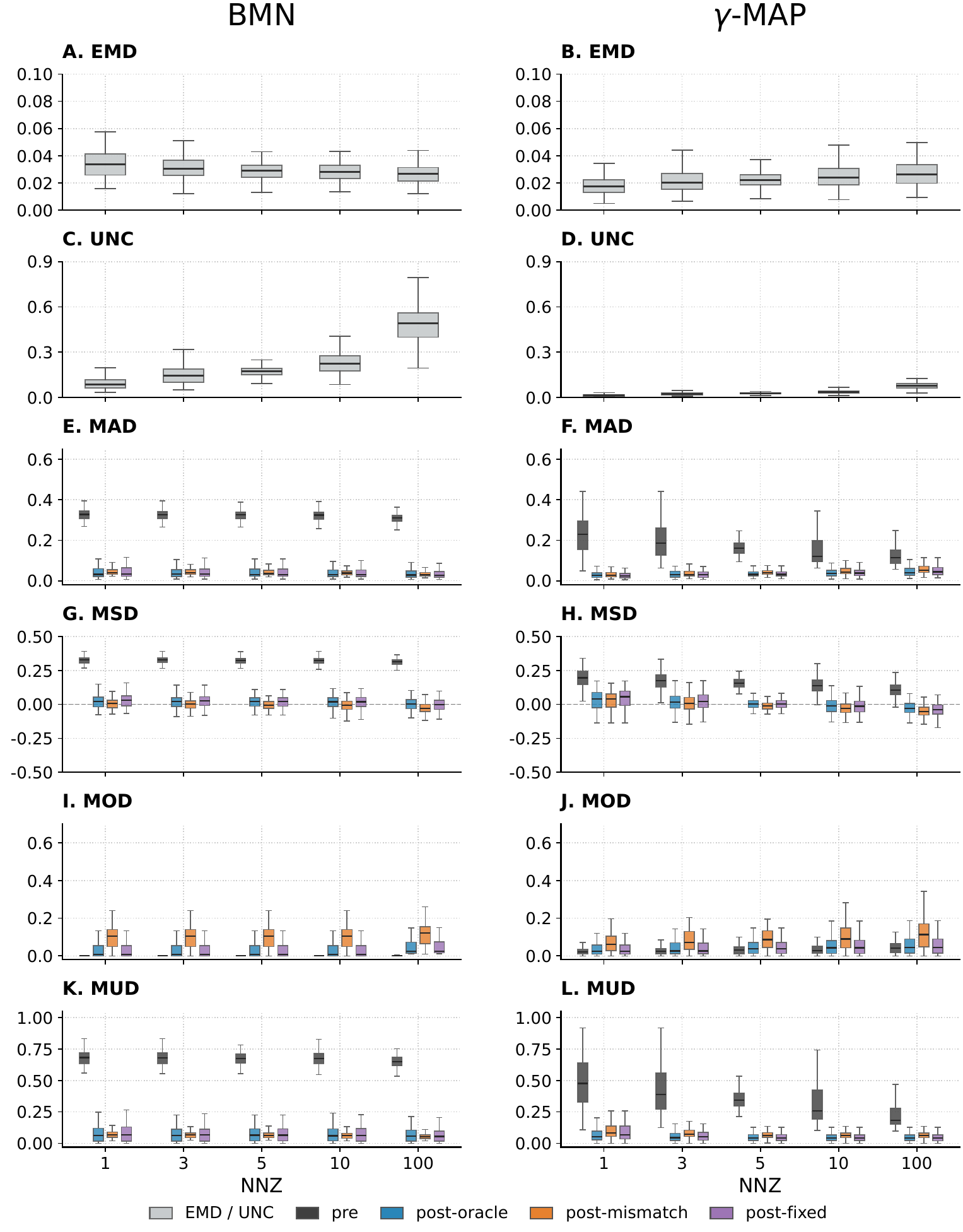}
  
\caption{Reduced free-MEG source-number sweep under oracle noise, varying $\mathrm{NNZ}\in\{1,3,5,10,100\}$ with $\alpha_{\mathrm{SNR}}=0.5$. Panel pairs \textbf{A--B}, \textbf{C--D}, \textbf{E--F}, \textbf{G--H}, \textbf{I--J}, and \textbf{K--L} show EMD, UNC, MAD, MSD, MOD, and MUD, respectively, for BMN+sLORETA (left) and $\gamma$-MAP+S-FLEX (right). Calibration modes are pre, post-oracle, post-mismatch, and post-fixed.}
\label{fig:supp-free-nnz-oracle}
\end{figure}

The oracle-noise sweeps (Supplementary Figures~\ref{fig:supp-free-snr-oracle} and~\ref{fig:supp-free-nnz-oracle}) show the same broad EMD and UNC trends as the adaptive-noise sweeps. Increasing SNR reduces UNC for both estimators and is accompanied by a modest decrease in EMD, particularly for $\gamma$-MAP+S-FLEX. Increasing NNZ raises UNC for both methods, while BMN+sLORETA EMD decreases and $\gamma$-MAP+S-FLEX EMD increases moderately. The raw calibration errors differ quantitatively from the adaptive-noise case, but post-oracle recalibration substantially reduces the dominant error throughout both sweeps. MSD, MOD, and MUD show that residual signed errors are condition dependent.

\begin{figure}[!htbp]
  \centering
  \includegraphics[width=0.98\linewidth,height=0.73\textheight,keepaspectratio]{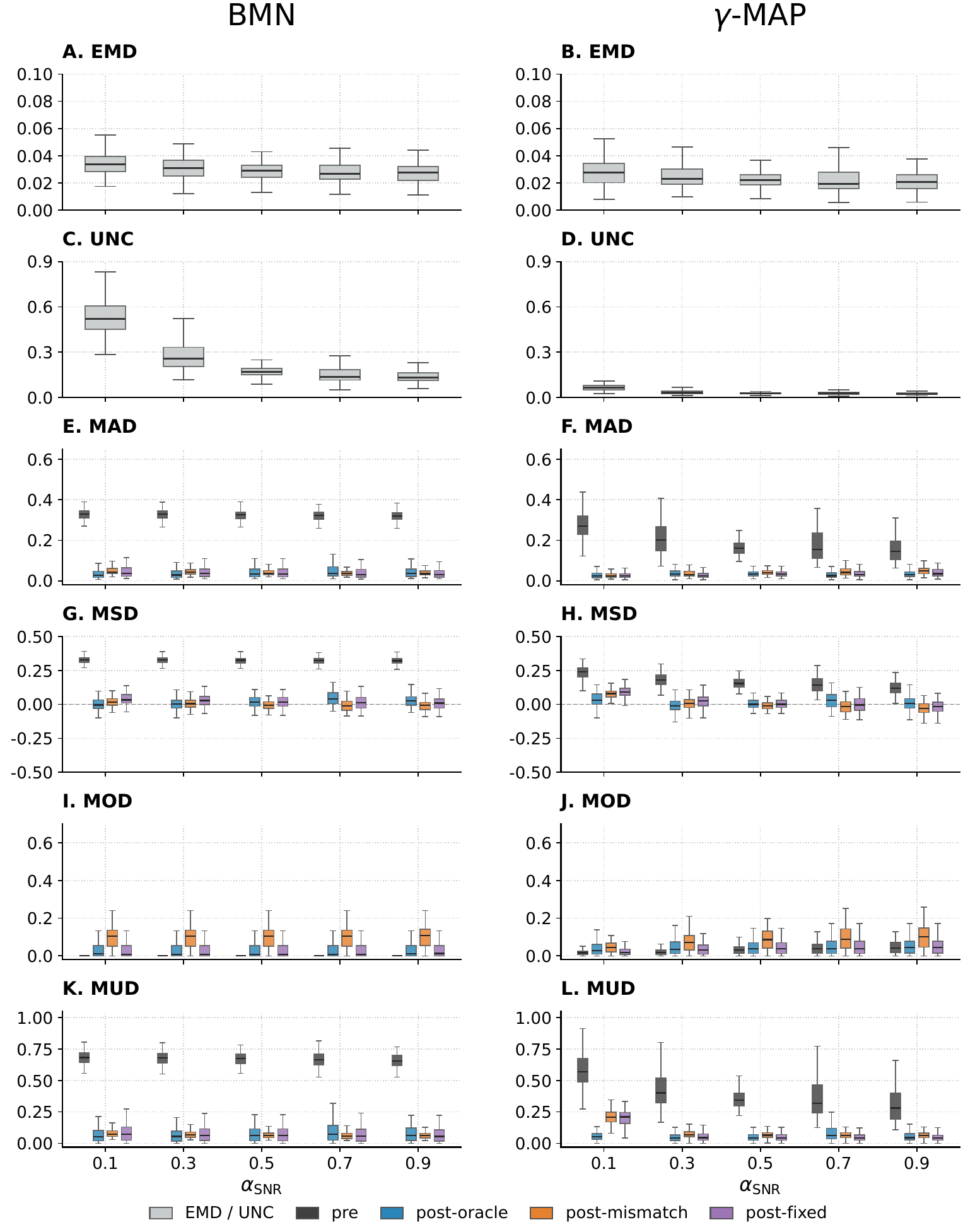}
  
\caption{Reduced free-MEG SNR sweep under baseline noise, varying $\alpha_{\mathrm{SNR}}\in\{0.1,0.3,0.5,0.7,0.9\}$ with $\mathrm{NNZ}=5$. Panel pairs \textbf{A--B}, \textbf{C--D}, \textbf{E--F}, \textbf{G--H}, \textbf{I--J}, and \textbf{K--L} show EMD, UNC, MAD, MSD, MOD, and MUD, respectively, for BMN+sLORETA (left) and $\gamma$-MAP+S-FLEX (right). Calibration modes are pre, post-oracle, post-mismatch, and post-fixed.}
\label{fig:supp-free-snr-baseline}
\end{figure}

\begin{figure}[!htbp]
  \centering
  \includegraphics[width=0.98\linewidth,height=0.73\textheight,keepaspectratio]{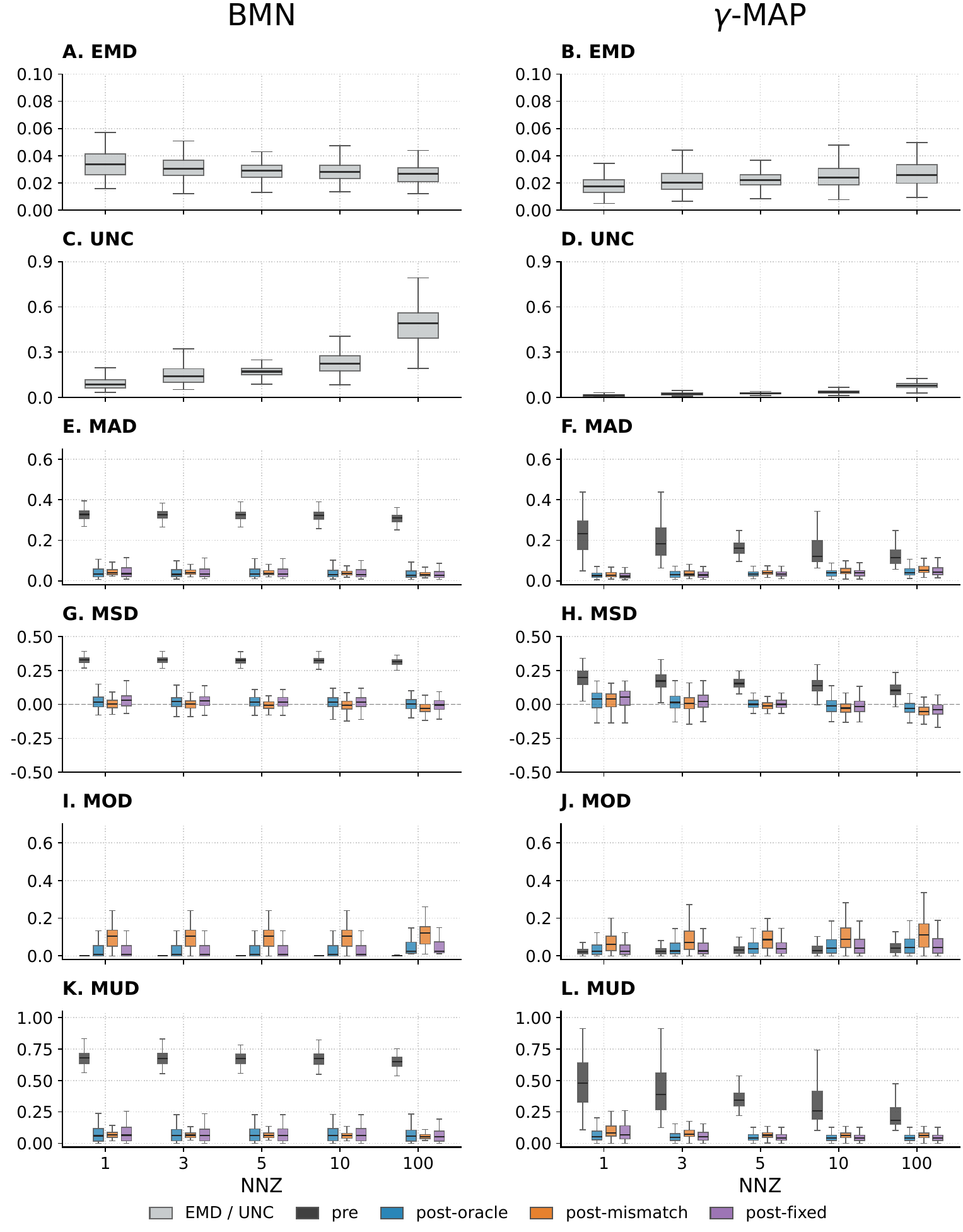}
  
\caption{Reduced free-MEG source-number sweep under baseline noise, varying $\mathrm{NNZ}\in\{1,3,5,10,100\}$ with $\alpha_{\mathrm{SNR}}=0.5$. Panel pairs \textbf{A--B}, \textbf{C--D}, \textbf{E--F}, \textbf{G--H}, \textbf{I--J}, and \textbf{K--L} show EMD, UNC, MAD, MSD, MOD, and MUD, respectively, for BMN+sLORETA (left) and $\gamma$-MAP+S-FLEX (right). Calibration modes are pre, post-oracle, post-mismatch, and post-fixed.}
\label{fig:supp-free-nnz-baseline}
\end{figure}

Supplementary Figures~\ref{fig:supp-free-snr-baseline} and~\ref{fig:supp-free-nnz-baseline} show the corresponding baseline-noise sweeps. Their reconstruction and uncertainty trends closely follow the oracle-noise results, including the estimator-specific EMD response to NNZ. Post-oracle recalibration remains effective across the tested conditions, with residual differences between calibration regimes most clearly reflected in the signed and maximum-deviation metrics.

Across the default setting and parameter sweeps, oracle, baseline, and adaptive noise specifications yielded similar reconstruction and uncertainty trends within each estimator, although residual calibration errors differed. The adaptive and baseline noise variance estimates were close to the data-generating variance, which may partly explain these similarities.

\subsubsection{Fixed-orientation companion results}
\label{subsubsec:fixed-companion-results}

The fixed-orientation results retain the main calibration
patterns observed in reduced free-MEG: BMN+sLORETA is
predominantly conservative before recalibration, whereas
$\gamma$-MAP+S-FLEX shows smaller, mixed-sign deviations.
Post-oracle recalibration improves coverage in both settings.
At the default condition, the BMN+sLORETA post-mismatch
and post-oracle curves remain close over much of the nominal
range for fixed orientation, whereas post-mismatch
recalibration produces residual undercoverage at some
intermediate credibility levels in reduced free-MEG.

The EMD trends also differ: EMD decreases with increasing
SNR under fixed orientation but changes only modestly
in reduced free-MEG. With increasing NNZ, the
$\gamma$-MAP+S-FLEX EMD remains comparatively stable
for fixed orientation but increases moderately in
reduced free-MEG. These comparisons are descriptive
rather than controlled estimates of an orientation effect,
since the coverage targets and simulation realizations
differ. Further fixed-orientation results are reported
in Supplementary Section~\ref{sec:supp-fixed-results}.

\paragraph{Fixed-orientation EEG reconstruction example}

\begin{figure}[!htbp]
  \centering
  \includegraphics[width=0.98\linewidth]{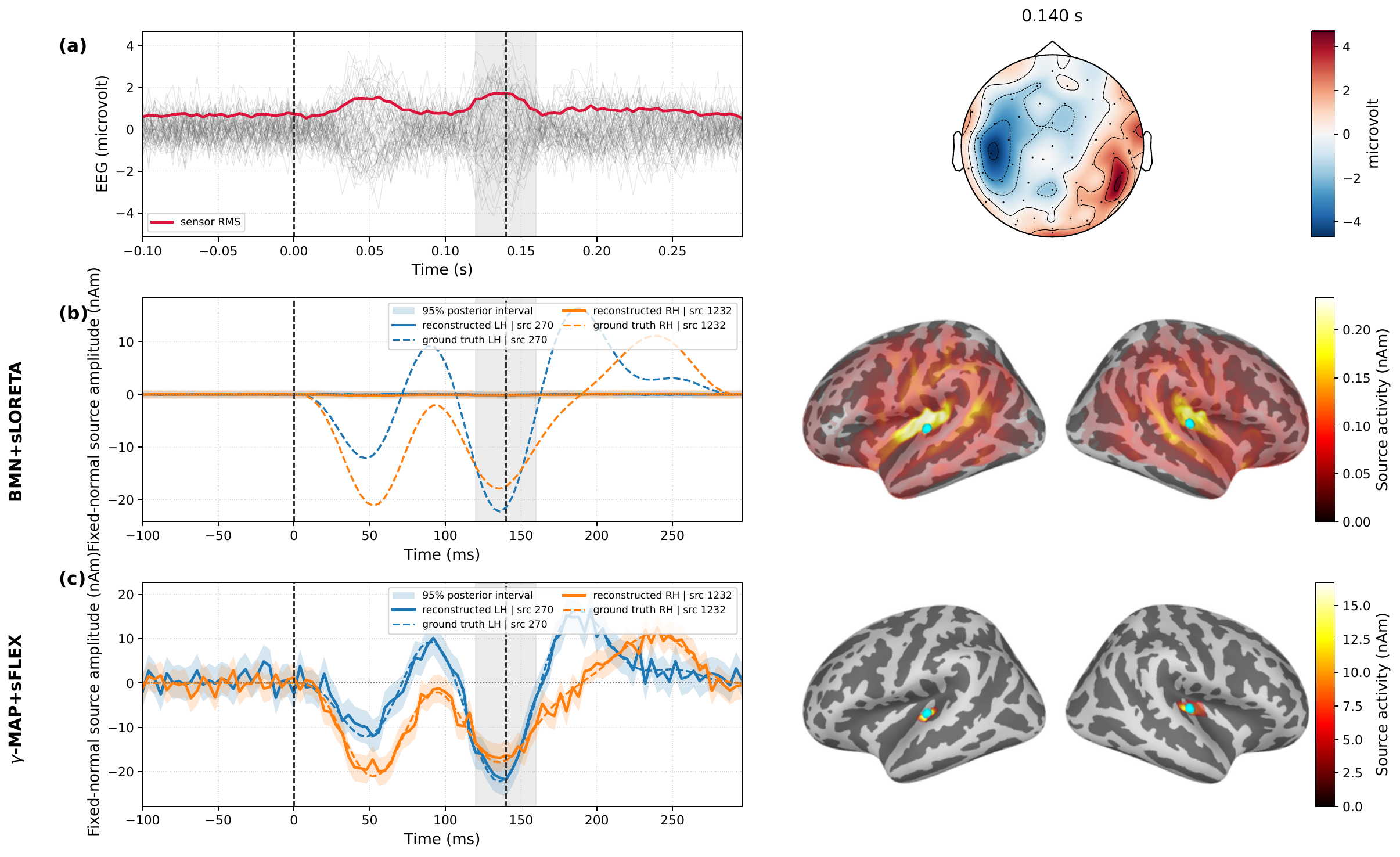}
  \caption{Fixed-orientation EEG reconstruction with adaptive noise learning on the \texttt{fsaverage} ico3 cortex ($N=1284$), using a BioSemi64 montage, a three-layer boundary-element model, two bilateral auditory generators, and $\alpha_{\mathrm{SNR}}=0.5$. \textbf{(a)} Simulated EEG channels in gray, sensor RMS in red, and signed scalp-potential topography at $140$~ms. The shaded interval denotes $120$--$160$~ms. \textbf{(b)} BMN+sLORETA posterior means at the two generating locations, with the corresponding ground-truth waveforms shown as dashed curves and pointwise 95\% posterior credible intervals. \textbf{(c)} Corresponding $\gamma$-MAP+S-FLEX reconstruction. Cortical maps show fixed-normal source-activity RMS over $120$--$160$~ms in nAm; cyan markers indicate the generating locations. Separate color scales are used for the two cortical maps.}
  \label{fig:simulation-fixed-eeg-single-run}
  \label{fig:single_run_fixed_eeg}
\end{figure}

Figure~\ref{fig:simulation-fixed-eeg-single-run} complements the repeated calibration experiments with a source- and time-resolved EEG example. This example illustrates local reconstruction and uncertainty, while we refer to the previous sections for a comprehensive quantitative assessment of reconstruction quality, uncertainty, and empirical calibration. The highlighted $120$--$160$~ms interval contains the response around the predominant peak at $\sim140$~ms, and the sensor and source topographies are evaluated at that latency.

Panel \textbf{(b)} shows that BMN+sLORETA distributes reconstructed activity broadly over the cortex while strongly attenuating the time courses at the two generating locations. Its local credible intervals are centered on those attenuated means. Panel \textbf{(c)} shows a more focal $\gamma$-MAP+S-FLEX reconstruction and closer recovery of the signed waveform deflections, particularly around the highlighted response. Note that the cortical maps use separate physical-amplitude scales.

\subsection{Results on Auditory MEG Data}
\label{subsec:results-real-data-auditory-meg}

The measured auditory MEG data provide a real-data evaluation in a setting with strong prior anatomical and physiological expectations for the cortical origin of the auditory response~\citep{Reite1994,LiegeoisChauvel1991,Nourski2017}. The comparison can therefore assess whether the reconstructed activity is
consistent with the expected bilateral auditory-cortical pattern, together with the evoked sensor response and the reported local uncertainty. Unlike in the simulations, however, the exact subject-specific latent source amplitudes are not known, so empirical coverage cannot be quantified directly.

\begin{figure}[!htbp]
  \centering
  \includegraphics[width=0.98\linewidth]{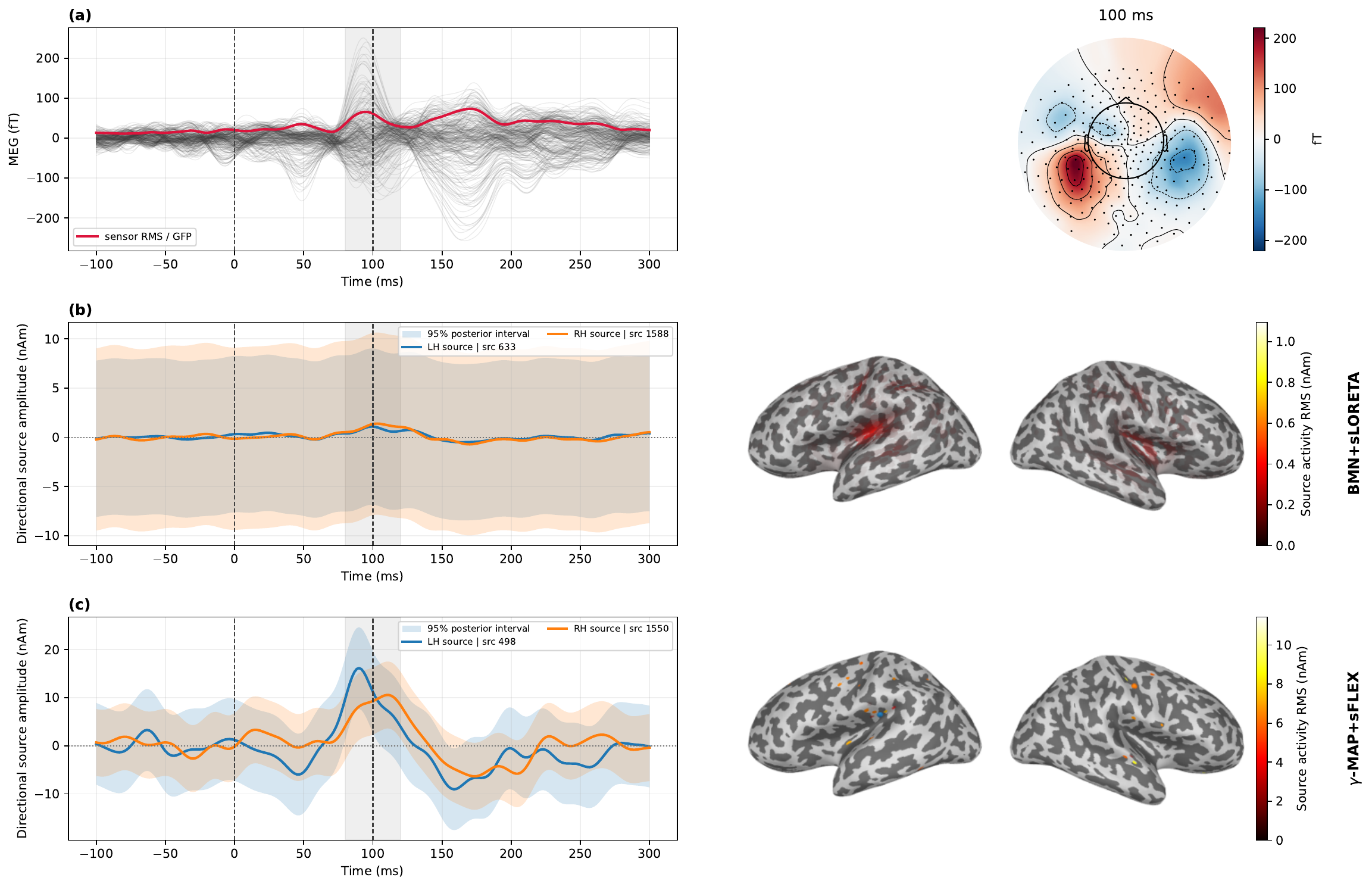}
  \caption{Measured auditory MEG and reduced free-orientation posterior summaries. \textbf{(a)} Evoked sensor fields, sensor RMS, and signed magnetic-field topography at $100$~ms. \textbf{(b)} BMN+sLORETA signed directional posterior means and pointwise 95\% credible intervals for the projected directional response at the selected left- and right-hemisphere sources, together with the cortical source-activity RMS map. \textbf{(c)} Corresponding $\gamma$-MAP+S-FLEX summaries. The curves are one-dimensional projections of the local vector posterior onto the fixed, source-specific leading SVD direction. Cortical color bars show source RMS in nAm and use separate ranges. The shaded response interval marks the response window.}
  \label{fig:auditory-meg-directional-posteriors}
  \label{fig:realdata_bst_standard_bmn_gamma}
\end{figure}

Figure~\ref{fig:auditory-meg-directional-posteriors}, panel \textbf{(a)}, shows a pronounced response in the M100 range and an opposite-polarity field pattern at $100$~ms. In panel \textbf{(b)}, BMN+sLORETA produces comparatively weak signed directional responses at the selected hemispheric sources, with broad intervals that include zero throughout the displayed response. Panel \textbf{(c)} shows stronger response-window deflections and a more focal $\gamma$-MAP+S-FLEX reconstruction. The intervals exclude zero around parts of the main positive response. 
BMN+sLORETA exhibits substantially smaller posterior mean amplitudes than $\gamma$-MAP+S-FLEX, while their uncertainty intervals have comparable widths, resulting in a more pronounced response relative to uncertainty for $\gamma$-MAP+S-FLEX.
 
The two estimators peak at different cortical vertices, for which time courses are displayed. Their curves therefore summarize estimator-specific hemispheric responses. For both, the reconstructed patterns, however, reproduce well-established activations of auditory-cortical sources.

\section{Discussion}
\label{sec:discussion}

The present study systematically evaluates whether posterior uncertainty from empirical-Bayesian M/EEG inverse imaging is empirically meaningful for the target source quantity, rather than treating posterior variance only as a by-product of regularization or prior specification. We hypothesized that joint adaptive learning of source and noise hyperparameters could reduce the dependence of uncertainty estimates on externally specified variance parameters and lead to near-calibration.

We found that empirical-Bayesian hyperparameter learning alone does not make raw posterior credible regions uniformly calibrated. Joint source and noise hyperparameter learning nevertheless yielded
reconstruction and uncertainty trends similar to those obtained with oracle and baseline noise
specifications, without requiring access to the data-generating noise variance. This illustrates that a covariance specification yielding a plausible posterior mean can still produce overconfident or
underconfident credible regions. In the default setting, $\gamma$-MAP+S-FLEX exhibited comparatively
small raw calibration errors. Post-hoc recalibration using isotonic regression substantially improved
empirical calibration for both reconstruction methods across the considered noise specifications.

Residual calibration errors depend on estimator structure, the geometry of the target uncertainty region, and the correspondence between calibration and evaluation conditions. The present work provides methods and tools for empirically assessing posterior credible-region calibration and, when necessary, improving it using independent simulations.

\subsection{Interpretation of the main findings}
\label{subsec:discussion_main_findings}

Spatial reconstruction, posterior uncertainty, and empirical calibration answer distinct questions. A smaller UNC indicates smaller average local posterior uncertainty, but only the relation between posterior error and posterior scale determines coverage. Similarly, a lower EMD concerns the spatial distribution of the posterior mean and need not imply better calibration of the local current vector. In the reported comparisons, EMD and UNC are calculated from the original fitted moments and are held fixed across recalibration regimes. The calibration metrics instead change when recalibration modifies the credibility thresholds used to construct the reported regions.

SESAMEEG characterizes uncertainty in source number and locations
through Monte Carlo approximation of their posterior, while source
strengths are treated analytically conditional on the dipole
configuration~\citep{Luria2024}. Here, we focus on Type-II Bayesian
learning and assess uncertainty in local source currents within
distributed empirical-Bayesian models whose posterior means and
covariances are available in closed form conditional on fitted
hyperparameters and, where applicable, selected coefficient support. Our results show that analytically available conditional posterior summaries do not by themselves ensure empirical calibration, motivating systematic coverage assessment and post-hoc recalibration.

The dense and sparse estimators exhibit markedly different default-setting uncertainty profiles. BMN+sLORETA ties the source variance across locations, whereas $\gamma$-MAP+S-FLEX estimates source-specific variance components~\citep{SekiharaNagarajan2015,Wipf2009,Cai2021}. At the default condition, BMN+sLORETA has larger UNC and predominantly conservative raw coverage, whereas $\gamma$-MAP+S-FLEX has lower EMD, substantially smaller UNC, and lower raw MAD (Figure~\ref{fig:main-free-snr-adaptive}). Post-oracle recalibration brings the two estimators to comparable median MAD values despite their different posterior scales. This reinforces that posterior concentration and empirical calibration are distinct properties. The fixed-orientation sweeps show the same broad contrast in uncertainty magnitude (Supplementary Figures~\ref{fig:supp-adaptive-snr}, \ref{fig:supp-oracle-snr}, and~\ref{fig:supp-baseline-snr}). Nevertheless, dense posterior support is not a mathematical guarantee of overcoverage, and sparsity is not a guarantee of calibration. The observed behavior depends jointly on the fitted mean, local covariance, evaluated truth, and the distribution of active and inactive locations.

In a conventional sparse formulation, coefficient pruning can lead to zero marginal posterior variance at individual source locations. Because retained S-FLEX basis functions overlap spatially, they can contribute nonzero uncertainty to locations whose corresponding coefficients are inactive and can therefore alleviate such voxel-wise marginal degeneracy. S-FLEX does not, however, restore source-space directions excluded by sparse coefficient selection. The joint source covariance remains confined to the span of the retained basis functions and can therefore remain globally rank deficient even when individual local covariance blocks are non-degenerate. In reduced free-orientation MEG, the basis must additionally respect the varying local orientation frames on the cortical surface. Thus, the relevant methodological questions concern how basis construction, normalization, spatial support, and coefficient selection determine local uncertainty geometry and the global rank structure. A systematic analysis of these properties goes beyond the present study and warrants a dedicated methodological investigation.

\begin{figure}[!htbp]
  \centering
  \includegraphics[width=0.98\linewidth]{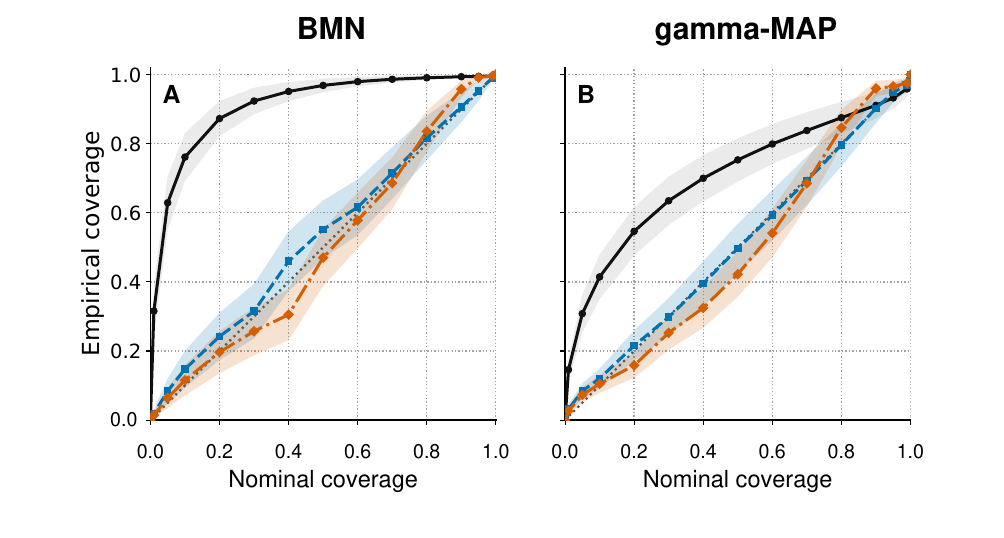}
  \caption{Reduced free-MEG calibration curves derived from joint two-dimensional credible-ellipse containment under oracle noise at $\alpha_{\mathrm{SNR}}=0.5$ and $\mathrm{NNZ}=5$. \textbf{A}: BMN+sLORETA. \textbf{B}: $\gamma$-MAP+S-FLEX. Lines denote mean empirical coverage over 100 evaluation runs; color-matched bands denote plus or minus one run-level standard deviation. The dotted diagonal indicates perfect calibration.}
  \label{fig:supp-free-oracle-calibration}
\end{figure}

\begin{figure}[!htbp]
  \centering
  \includegraphics[width=0.98\linewidth]{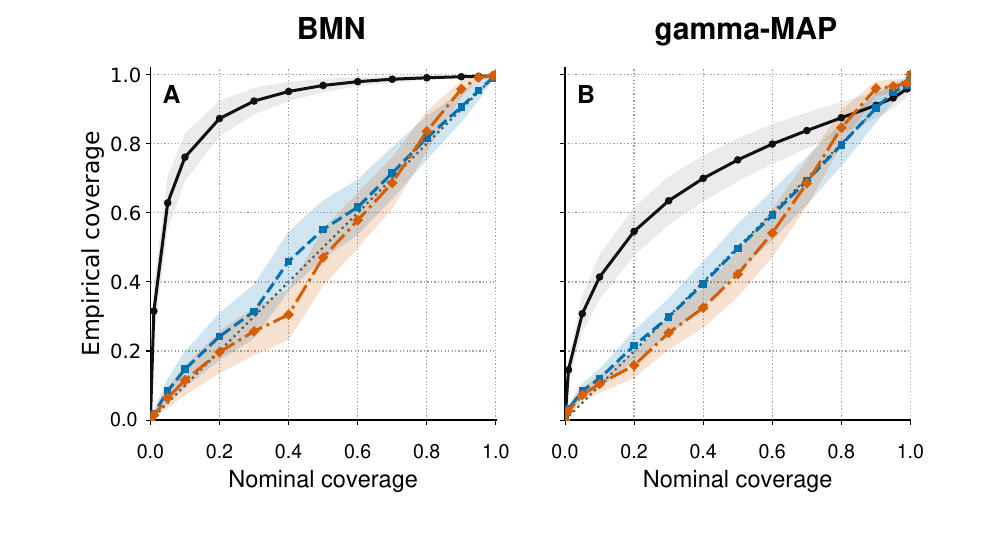}
  \caption{Reduced free-MEG calibration curves derived from joint two-dimensional credible-ellipse containment under baseline noise at $\alpha_{\mathrm{SNR}}=0.5$ and $\mathrm{NNZ}=5$. \textbf{A}: BMN+sLORETA. \textbf{B}: $\gamma$-MAP+S-FLEX. Lines denote mean empirical coverage over 100 evaluation runs; color-matched bands denote plus or minus one run-level standard deviation. The dotted diagonal indicates perfect calibration.}
  \label{fig:supp-free-baseline-calibration}
\end{figure}

The oracle- and baseline-noise calibration curves closely resemble those obtained under adaptive noise learning (Figure~\ref{fig:main-free-ellipse-calibration}; Supplementary Figures~\ref{fig:supp-free-oracle-calibration} and~\ref{fig:supp-free-baseline-calibration}). Their similar coverage patterns are consistent with the controlled stationary Gaussian-noise simulations, in which the pre-stimulus interval is representative of the simulated noise process. Source-prior structure, plug-in hyperparameter estimation, and reconstruction bias may nevertheless contribute to residual calibration errors.

The reduced free-MEG sweeps show broadly consistent SNR- and source-number-dependent reconstruction and uncertainty trends across adaptive, oracle, and baseline noise specifications, although the magnitude and sign of residual calibration errors differ. Joint learning of source and noise variances may also permit compensation between their effects. Consequently, the calibration properties of adaptively learned posteriors should be evaluated empirically rather than inferred from evidence-maximization alone.

The SNR and source-number sweeps show systematic condition dependence in both orientation settings. Increasing SNR consistently contracts posterior uncertainty, with the strongest absolute reduction for BMN+sLORETA. In the fixed-orientation setting, EMD also decreases with increasing SNR, whereas the reduced free-MEG EMD trajectories vary more modestly. Increasing NNZ raises posterior uncertainty in both orientation settings. BMN+sLORETA EMD decreases with source number in both cases, whereas the $\gamma$-MAP+S-FLEX response is comparatively flat in the fixed-orientation setting and shows a modest increase in the reduced free-MEG setting. Post-oracle recalibration reduces the dominant calibration errors across these conditions, while the signed and maximum-deviation metrics reveal residual condition dependence.

Recalibration transfer is condition dependent. The pooled-mismatch setting differs not only in head geometry but also in the breadth and size of its calibration set. Post-oracle recalibration is more effective across the tested conditions, although it can still leave larger residual errors or introduce undercoverage in parts of the parameter range. However, as SNR and sparsity profiles are often unknown, the practical value of the post-oracle calibration findings is limited -- they primarily serve as a hypothetical upper bound for calibration performance.

The reduced free-MEG credible ellipse describes joint uncertainty in the retained two-dimensional current vector, whereas fixed-orientation scalar intervals concern a different coverage event. Their empirical coverage should therefore be interpreted relative to the corresponding target. When the prescribed cortical-normal constraint is correct, it reduces model flexibility; when inaccurate, it can introduce model-specification bias~\citep{MNEOrientations2026}. Across the evaluated SNR and source-number ranges, the two orientation settings show broadly comparable reconstruction and calibration trends. However, their different coverage targets and independently generated simulation realizations preclude interpreting the observed differences as a controlled causal effect of orientation dimensionality.

The fixed-orientation EEG example (Figure~\ref{fig:simulation-fixed-eeg-single-run}) shows strongly attenuated BMN+sLORETA posterior mean waveforms at the generating locations, whereas $\gamma$-MAP+S-FLEX recovers the highlighted responses more closely. Although the posterior uncertainty intervals have comparable widths, $\gamma$-MAP+S-FLEX exhibits larger mean deflections relative to uncertainty. In the displayed pointwise intervals, the BMN+sLORETA responses remain compatible with zero, whereas $\gamma$-MAP+S-FLEX shows several positive and negative deflections whose intervals exclude zero.

A similar contrast is observed in the measured auditory MEG data (Figure~\ref{fig:auditory-meg-directional-posteriors}). $\gamma$-MAP+S-FLEX produces a more focal spatial pattern and larger M100-aligned posterior mean deflections, whereas BMN+sLORETA yields a broader reconstruction and substantially smaller mean responses, with uncertainty intervals of comparable width. The predominant activations in bilateral auditory cortices are consistent with the known neuroanatomy and physiology of auditory responses~\citep{Reite1994,LiegeoisChauvel1991,Nourski2017}.

\subsection{Additional technical considerations}
\label{subsec:discussion_technical}

\paragraph{Shrinkage and calibration}

Both BMN+sLORETA and $\gamma$-MAP+S-FLEX employ zero-mean source priors, which can shrink posterior means toward zero. For a fixed posterior uncertainty scale, shrinkage may increase coverage at inactive locations while reducing coverage at truly active sources. Because inactive sources dominate the simulated source grids, aggregate calibration can mask substantial active-source errors. The predominantly conservative BMN+sLORETA coverage and mixed-sign deviations of $\gamma$-MAP+S-FLEX cannot be attributed to shrinkage alone. Establishing its contribution requires separate calibration analyses for active and inactive sources.

A simple Gaussian shrinkage model provides insight into the different coverage behavior of active
and inactive sources. For a zero-mean Gaussian prior, the posterior mean is attenuated by a
factor $a\in(0,1)$, whereas the posterior standard deviation scales with $\sqrt{a}$. At a truly
inactive source, this difference yields conservative coverage in the scalar model. At an active source,
the standardized estimation error additionally contains a bias term proportional to
$(1-a)x^{\mathrm{true}}$, which can lead to undercoverage for sufficiently large source
amplitudes. In M/EEG reconstruction, however,
source leakage, the structure of the learned covariance matrices, and support selection complicate this relationship. Thus, shrinkage
provides a possible explanation for the predominantly conservative BMN+sLORETA coverage,
but its contribution requires further analysis conditional on source activity and amplitude.


\paragraph{Bayesian uncertainty and resampling}

Resampling repeated, approximately exchangeable ERP/ERF trials can estimate the sampling variability of a source estimator. It is a legitimate alternative or complement to Bayesian posteriors. For sparse estimators, perturbing the data can change the resulting source support; a source-wise resampling distribution can consequently contain both a point mass at zero and nonzero estimates, or several regimes associated with competing source locations.
 
Conversely, an empirical-Bayesian Gaussian posterior conditional on fitted hyperparameters does not automatically integrate uncertainty in support selection. The attraction of the Bayesian formulation here is the availability of conditional local uncertainty from a single fitted dataset without requiring a trial-resampling ensemble. Comparisons with selection-aware resampling procedures are a possible subject of future work.

\paragraph{Meaning of the reported recalibration}

The reported procedure remaps credibility thresholds within a nested family of regions. Accordingly, the EMD and UNC values shown here remain those of the original fitted posterior. In particular, a calibration map for a given credibility level and raw posterior does not by itself define a unique new multivariate posterior distribution. If a separate recalibrated distribution is constructed, its second moments can differ from those of the original Gaussian posterior even when its credible regions retain the original local geometry. At a fixed credibility level, the threshold adjustment can be represented as a rescaling of the corresponding Gaussian credible region. However, level-dependent adjustments do not generally correspond to a single recalibrated Gaussian covariance matrix. The empirical interpretation of a remapped level is relative to the data-generating distribution represented by the calibration runs; independent validation remains necessary, as in other post-hoc uncertainty-calibration procedures~\citep{Kuleshov2018}.

\paragraph{Other target quantities}

The same Gaussian propagation principle extends from temporal averages to fixed linear targets such as regional averages, condition contrasts, and spatial or spectral projections~\citep{Anderson2003}. Both the mean and the covariance must be transformed. The simple $1/T$ covariance scaling for a temporal average assumes conditionally independent time points with equal marginal covariance; temporally correlated posteriors require cross-time covariance terms. Nonlinear targets, including current magnitude and orientation, generally require transformed distributions, posterior sampling, or approximations. Calibration behavior must then be defined and evaluated separately for the new target, as it is not inherited from scalar or vector-valued sources.

\subsection{Limitations and future directions}
\label{subsec:discussion_limitations}

Several limitations apply. First, the reported posteriors are conditional on point estimates of variance hyperparameters and, where applicable, a selected coefficient support. For sparse SBL-type solutions, S-FLEX can redistribute retained uncertainty across cortical locations but does not restore discarded source-space directions. The resulting estimates may therefore miss uncertainty components associated with these directions. Fully Bayesian hierarchical inference that accounts for hyperparameter and support uncertainty offers a possible extension~\citep{Jun2008,Lucka2012}, although computational cost and approximation error must be considered.

Although the present experiments compare adaptive, oracle, and baseline noise specifications, alternative hyperparameter-selection criteria and hierarchical variance models were not systematically evaluated. Future work should compare noise estimates from baseline segments and empty-room recordings, held-out predictive criteria, hierarchical variance models, and calibration-aware hyperparameter-selection procedures, with calibration and evaluation data kept strictly separate.

The fixed- and reduced free-orientation benchmarks were generated independently rather than paired run by run and evaluate different coverage targets. Consequently, comparisons between the two settings remain descriptive.

The simulated data cover a controlled family of source and noise conditions, including Gaussian sensor noise, four head models, prescribed SNR and NNZ grids, and a restricted family of event-related source dynamics. These choices permit interpretable comparisons but do not represent structured environmental or physiological artifacts, broader forward-model errors, correlated or spatially extended sources, nonstationary noise, or the full range of temporal source dynamics encountered in measured data. Future evaluations should therefore broaden the data-generating distribution and include dedicated noise-misspecification experiments.

Whole-grid performance aggregation can also mask practically important local errors. In sparse simulations, inactive locations dominate the source count, so an apparently satisfactory aggregate curve need not imply satisfactory coverage at active, deep, or weak sources. Future work should examine calibration conditional on activity, amplitude, depth, and anatomical region and should distinguish uncertainty in the current at a prescribed cortical location from uncertainty in source location or spatial support. Such analyses would connect the reported credible intervals and ellipses more directly to source-detection and localization decisions.

The recalibration maps themselves inherit the scope of their training distribution. Post-oracle recalibration is effective in the tested conditions but requires accurately matched simulation conditions, which may be unavailable in practice. Post-fixed and post-mismatch transfer can leave larger residual errors away from their representative conditions.

Larger and more heterogeneous simulation libraries or regularized condition-dependent calibration models may improve transfer. However, a global one-dimensional calibration map cannot generally guarantee calibration across source locations, amplitudes, orientations, or individual runs.

Although the mathematical framework accommodates fixed orientation, reduced free-orientation MEG, and full three-dimensional free-orientation sources, the principal vector-valued calibration benchmark in the present study is reduced free-orientation MEG. Empirical calibration of full three-dimensional EEG credible ellipsoids therefore remains to be established. More generally, the current credible regions quantify uncertainty in a source coefficient or local current vector at a prescribed cortical location; they are not credible regions for the unknown source location, spatial support, orientation, or complete spatio-temporal activation pattern. Future work should address spatial credible sets, uncertainty in current magnitude and orientation, and simultaneous temporal uncertainty.

Finally, the measured-data analysis contains one auditory MEG dataset. This setting nevertheless provides a particularly informative real-data example because the neuroanatomy and physiology of auditory responses are well characterized: MEG studies have consistently localized the M100 response to auditory cortex in or around Heschl's gyrus, while direct intracranial recordings provide complementary evidence for the functional organization of Heschl's gyrus and adjacent auditory cortex~\citep{Reite1994,LiegeoisChauvel1991,Nourski2017}. However, the subject-specific source currents are unobserved, so empirical coverage cannot be evaluated directly.

Future real-world investigations should include multiple subjects, tasks, and acquisition systems, together with forward-matched simulations and explicit comparison with established invasive and multimodal evidence. An important downstream direction is to propagate calibrated source uncertainty into uncertainty-aware decoding, source classification, and clinical decision rules.


Although the present study focuses on Type-II empirical-Bayesian inverse solvers with conditionally Gaussian posteriors, the calibration framework is not inherently restricted to this setting. The same coverage assessment and post-hoc recalibration can also
be applied to Type-I Bayesian approaches, fully Bayesian hierarchical models, and non-Gaussian inverse models whenever posterior credible
regions or reliable posterior samples can be constructed for the chosen target. The broader methodological point is the separation of three questions that should not be conflated in source imaging: where the posterior places the activity, how much uncertainty it assigns, and whether the stated credibility of that uncertainty is empirically trustworthy.

\section{Conclusion}

We presented a unified framework for uncertainty quantification and empirical calibration in Bayesian M/EEG source imaging, covering fixed- and free-orientation source models and dense
and sparse empirical-Bayesian inverse methods. Across different noise specifications, $\gamma$-MAP+S-FLEX achieved lower source localization error, lower posterior uncertainty, and smaller
raw calibration errors than BMN+sLORETA in the principal default-setting comparison. Post-hoc isotonic recalibration substantially improved agreement between empirical and nominal
coverage for both methods, although its effectiveness depended on the correspondence between calibration and evaluation conditions. 
All code and data used in this study are made openly available. Analyses were implemented within an accompanying extendable open-source Python package supporting the future development and benchmarking of calibrated M/EEG inverse modeling approaches.

\section*{Acknowledgments}

The first author would like to express his particular gratitude to Ali Hashemi for his valuable theoretical insights and code-related advice concerning the formulation and implementation of Champagne algorithm.

\section*{Code Availability Statement}

The core software implementing the inverse solvers, uncertainty-quantification methods, simulation workflows, and calibration analyses used in this study is publicly available as part of the CaliBrain package at
\url{https://github.com/braindatalab/CaliBrain}.

\bibliography{references}
\clearpage

\beginsupplement

\begin{center}
  {\Large\bfseries Uncertainty calibration in \\ Bayesian M/EEG source imaging \par}
  \vspace{0.35em}
  {\large\bfseries Supplementary Material\par}

  \vspace{0.75em}
  {\normalsize
    Ismail Huseynov$^{1,2}$, Mohammad Orabe$^{1,2}$, \\Srikantan Nagarajan$^{3}$, Stefan Haufe$^{1,2,4}$\par
  }

  \vspace{0.75em}
  {\footnotesize
    $^{1}$ Physikalisch\mbox{-}Technische Bundesanstalt Braunschweig und Berlin, Berlin, Germany\\
    $^{2}$ Technische Universit\"at Berlin, Berlin, Germany\\
    $^{3}$ University of California San Francisco, Biomagnetic Imaging Laboratory, Department of Radiology and Biomedical Imaging, San Francisco, CA, United States\\
    $^{4}$ Charit\'e\,-\,Universit\"atsmedizin Berlin, Berlin, Germany\par
  }
\end{center}
\bigskip

\section{Gaussian Marginals and Linear Projections}
\label{app:gauss_marginals}

The closure of the multivariate normal distribution under affine transformations and marginalization is standard~\citep{Anderson2003}. The results are stated here to make explicit how the source-wise covariance blocks and temporally aggregated covariances used in the manuscript follow from the full Gaussian posterior.

\begin{theorem}[Affine transformation of a Gaussian vector]
Let $\mathbf{x}\in\mathbb{R}^{n}$ satisfy
\[
  \mathbf{x}\sim\mathcal{N}(\boldsymbol{\mu},\boldsymbol{\Sigma}),
\]
where $\boldsymbol{\Sigma}$ is symmetric positive semidefinite. For a deterministic matrix $A\in\mathbb{R}^{m\times n}$ and vector $\mathbf{b}\in\mathbb{R}^{m}$, define
\[
  \mathbf{y}=A\mathbf{x}+\mathbf{b}.
\]
Then
\[
  \mathbf{y}\sim\mathcal{N}\!\left(A\boldsymbol{\mu}+\mathbf{b},\,A\boldsymbol{\Sigma}A^{\top}\right).
\]
\end{theorem}

\begin{theorem}[Gaussian marginal selected from a subvector]
Let $\mathbf{x}\sim\mathcal{N}(\boldsymbol{\mu},\boldsymbol{\Sigma})$, and let $\mathbf{x}_{s}\in\mathbb{R}^{m}$ contain any selected set of components of $\mathbf{x}$. Then $\mathbf{x}_{s}$ is Gaussian. Its mean is the corresponding subvector of $\boldsymbol{\mu}$, and its covariance is the corresponding principal submatrix of $\boldsymbol{\Sigma}$.
\end{theorem}

\begin{proof}
Let $S\in\mathbb{R}^{m\times n}$ be the selection matrix whose rows select the required components of $\mathbf{x}$. Then
\[
  \mathbf{x}_{s}=S\mathbf{x}.
\]
The affine-transformation theorem with $A=S$ and $\mathbf{b}=\mathbf{0}$ gives
\[
  \mathbf{x}_{s}\sim
  \mathcal{N}\!\left(S\boldsymbol{\mu},\,S\boldsymbol{\Sigma}S^{\top}\right).
\]
By construction, $S\boldsymbol{\mu}$ is the selected mean subvector and $S\boldsymbol{\Sigma}S^{\top}$ is the associated principal covariance submatrix.
\end{proof}

\subsection{Application to source-wise posterior blocks}
\label{app:source_block_marginal}

Let the single-time source vector be partitioned into $N$ local blocks of dimension $d$,
\[
  \mathbf{x}(t)=
  \begin{bmatrix}
    \mathbf{x}_{1}(t)^{\top} & \cdots & \mathbf{x}_{N}(t)^{\top}
  \end{bmatrix}^{\top}
  \in\mathbb{R}^{dN},
\]
and suppose
\[
  \mathbf{x}(t)\mid\mathbf{Y}
  \sim
  \mathcal{N}\!\left(\boldsymbol{\mu}(t),\boldsymbol{\Sigma}_{x}\right).
\]
For source $i$, define the block-selection matrix
\[
  S_i=\mathbf{e}_{i}^{\top}\otimes I_d
  \in\mathbb{R}^{d\times dN},
\]
where $\mathbf{e}_{i}$ is the $i$th canonical basis vector in $\mathbb{R}^{N}$. Since $\mathbf{x}_{i}(t)=S_i\mathbf{x}(t)$, Gaussian marginalization gives
\begin{equation}
  \mathbf{x}_{i}(t)\mid\mathbf{Y}
  \sim
  \mathcal{N}\!\left(
    S_i\boldsymbol{\mu}(t),
    S_i\boldsymbol{\Sigma}_{x}S_i^{\top}
  \right)
  =
  \mathcal{N}\!\left(\boldsymbol{\mu}_{i}(t),\boldsymbol{\Sigma}_{ii}\right).
  \label{eq:supp_local_gaussian_marginal}
\end{equation}
Thus, the local covariance $\boldsymbol{\Sigma}_{ii}$ is exactly the $i$th diagonal $d\times d$ block of the full posterior covariance; that is, off-diagonal source blocks do not affect the local marginal distribution, although they do encode cross-source dependence that would be required for a joint multi-source credible set.

\subsection{Temporal averaging}
\label{app:temporal_average_gaussian}

Stack the $T$ source vectors as
\[
  \mathbf{x}_{1:T}=
  \begin{bmatrix}
    \mathbf{x}(1)^{\top} & \cdots & \mathbf{x}(T)^{\top}
  \end{bmatrix}^{\top}.
\]
Under the temporally independent posterior model used in the manuscript,
\[
  \mathbf{x}_{1:T}\mid\mathbf{Y}
  \sim
  \mathcal{N}\!\left(
    \boldsymbol{\mu}_{1:T},
    I_T\otimes\boldsymbol{\Sigma}_{x}
  \right).
\]
Define
\[
  A_T=\frac{1}{T}\left(\mathbf{1}_{T}^{\top}\otimes I_{dN}\right),
  \qquad
  \bar{\mathbf{x}}=A_T\mathbf{x}_{1:T}.
\]
The affine-transformation theorem yields
\[
  \bar{\mathbf{x}}\mid\mathbf{Y}
  \sim
  \mathcal{N}\!\left(\bar{\boldsymbol{\mu}},\bar{\boldsymbol{\Sigma}}\right),
\]
with
\begin{align}
  \bar{\boldsymbol{\mu}}
  &=
  \frac{1}{T}\sum_{t=1}^{T}\boldsymbol{\mu}(t),\\
  \bar{\boldsymbol{\Sigma}}
  &=
  A_T\left(I_T\otimes\boldsymbol{\Sigma}_{x}\right)A_T^{\top}
  =
  \frac{1}{T}\boldsymbol{\Sigma}_{x}.
  \label{eq:supp_temporal_average_covariance}
\end{align}
Applying $S_i$ once more gives the aggregated local marginal
\begin{equation}
  \bar{\mathbf{x}}_{i}\mid\mathbf{Y}
  \sim
  \mathcal{N}\!\left(
    \bar{\boldsymbol{\mu}}_{i},
    \frac{1}{T}\boldsymbol{\Sigma}_{ii}
  \right).
  \label{eq:supp_aggregated_local_marginal}
\end{equation}
The factor $1/T$ follows from posterior independence across time. If temporal posterior dependence is introduced, the covariance of the average must instead include all cross-time covariance terms.

\subsection{SVD-defined directional posterior marginals and credible intervals for free-orientation sources}
\label{subsec:supp_svd_directional_marginal}

At cortical location $i$, consider a free-orientation posterior represented in
a local coordinate system of dimension $d$,
\begin{equation*}
  \mathbf x_i(t)\mid\mathbf Y
  \sim
  \mathcal N_d
  \left(
    \boldsymbol\mu_i(t),
    \boldsymbol\Sigma_{ii}
  \right),
  \qquad
  \boldsymbol\mu_i(t)\in\mathbb R^d,
  \quad
  \boldsymbol\Sigma_{ii}\in\mathbb R^{d\times d}.
\end{equation*}
The reduced free-orientation MEG representation corresponds to $d=2$,
whereas a full free-orientation EEG representation corresponds to $d=3$.
For $d=1$, the local posterior is already scalar and the construction below
reduces to the conventional fixed-orientation credible interval without
requiring an SVD-defined direction.

The full $d$-dimensional credible ellipse or ellipsoid remains the
dimension-matched posterior uncertainty summary. A one-dimensional
directional marginal can additionally be defined by selecting a single
source-specific axis from the temporal structure of the local posterior mean.

Let $\mathcal T_d$ be a prespecified reference interval and define
\begin{equation*}
  \mathbf M_i
  =
  \begin{bmatrix}
    \boldsymbol\mu_i(t_1) &
    \cdots &
    \boldsymbol\mu_i(t_K)
  \end{bmatrix}
  \in\mathbb R^{d\times K},
  \qquad
  t_k\in\mathcal T_d.
\end{equation*}
The matrix $\mathbf M_i$ is not temporally centered. Thus, its dominant
direction is determined relative to the origin of the local source space,
which represents the physical null state of zero source current, rather than
relative to the temporal average of the response.

Let
\begin{equation*}
  \mathbf M_i
  =
  \mathbf U_i
  \mathbf D_i
  \mathbf V_i^\top
\end{equation*}
be the thin singular value decomposition, where
\begin{equation*}
  \mathbf D_i
  =
  \operatorname{diag}
  \left(
    \sigma_{i1},\ldots,\sigma_{ir}
  \right),
  \qquad
  \sigma_{i1}\geq\cdots\geq\sigma_{ir}\geq0,
  \qquad
  r=\min(d,K).
\end{equation*}
Provided that $\sigma_{i1}>0$, the leading left singular vector
\begin{equation*}
  \mathbf u_i
  =
  \mathbf U_i\mathbf e_1,
  \qquad
  \|\mathbf u_i\|_2=1,
\end{equation*}
defines the dominant orientation of the local posterior-mean trajectory over
$\mathcal T_d$. The direction is estimated once and subsequently held fixed
over the complete epoch, thereby defining a common scalar coordinate for the
temporal posterior.

When the leading singular value is separated from the remaining singular
spectrum, $\mathbf u_i$ is uniquely identified up to sign. If the leading
singular values are nearly equal, the local trajectory does not determine a
well-separated dominant axis, and the corresponding directional coordinate
may be sensitive to small perturbations. In such cases, the full
$d$-dimensional credible ellipse or ellipsoid remains the more appropriate
uncertainty representation.

The sign of $\mathbf u_i$ is intrinsically arbitrary and is therefore resolved
by a deterministic convention specified in the corresponding study design.
Replacing $\mathbf u_i$ by $-\mathbf u_i$ reverses the displayed posterior
mean but leaves the projected variance, interval width, and zero-containment
decision unchanged.

If
\[
  \mathbf Q_i\in\mathbb R^{3\times d}
\]
maps the local source coordinates to physical three-dimensional source space,
the corresponding physical directional axis is
\begin{equation*}
  \widetilde{\mathbf u}_i
  =
  \mathbf Q_i\mathbf u_i.
\end{equation*}
For reduced free-orientation MEG, $\mathbf Q_i$ contains the two retained local
MEG-sensitive basis vectors. For a free-orientation EEG representation already
expressed in a three-dimensional orthonormal source frame,
$\mathbf Q_i$ may be taken as the corresponding $3\times3$ local basis,
including the identity matrix when Cartesian source coordinates are used.

Define the directional source component by
\begin{equation*}
  a_i(t)
  =
  \mathbf u_i^\top\mathbf x_i(t).
\end{equation*}
Because a linear transformation of a Gaussian random vector is Gaussian,
\begin{equation*}
  a_i(t)\mid\mathbf Y
  \sim
  \mathcal N
  \left(
    m_i(t),
    v_i
  \right),
\end{equation*}
where
\begin{equation*}
  m_i(t)
  =
  \mathbf u_i^\top\boldsymbol\mu_i(t),
  \qquad
  v_i
  =
  \mathbf u_i^\top
  \boldsymbol\Sigma_{ii}
  \mathbf u_i.
\end{equation*}

Let $c\in(0,1)$ denote the posterior credibility level and let
$z_p=\Phi^{-1}(p)$ be the $p$-quantile of the standard normal distribution.
The central pointwise $100c\%$ directional credible interval is
\begin{equation*}
  \mathcal I_{i,\mathrm{dir}}^{(c)}(t)
  =
  \left[
    m_i(t)
    -
    z_{(1+c)/2}\sqrt{v_i},
    \;
    m_i(t)
    +
    z_{(1+c)/2}\sqrt{v_i}
  \right].
\end{equation*}
Equivalently,
\begin{equation*}
  \mathcal I_{i,\mathrm{dir}}^{(c)}(t)
  =
  \left[
    m_i(t)
    -
    \sqrt{\chi_1^2(c)\,v_i},
    \;
    m_i(t)
    +
    \sqrt{\chi_1^2(c)\,v_i}
  \right],
\end{equation*}
since
\begin{equation*}
  \chi_1^2(c)
  =
  z_{(1+c)/2}^{\,2}.
\end{equation*}
Consequently,
\begin{equation*}
  \Pr
  \left(
    a_i(t)
    \in
    \mathcal I_{i,\mathrm{dir}}^{(c)}(t)
    \,\middle|\,
    \mathbf Y
  \right)
  =
  c.
\end{equation*}

The resulting interval is a one-dimensional posterior marginal along the
dominant local orientation. It is conditional on the fitted posterior, the
selected cortical location, the reference interval $\mathcal T_d$, and the
direction derived from the posterior mean. It therefore does not incorporate
source-selection or direction-estimation uncertainty. Moreover, it is a
pointwise temporal interval rather than a simultaneous credible band, and it
does not replace the dimension-matched spatial credible ellipse or ellipsoid
associated with the full free-orientation posterior.

\subsection{Lead-field representations for fixed and
reduced free orientations}
\label{sec:source_orientation_leadfields}

Let $\mathbf L^{(3)}\in\mathbb R^{M\times3N}$ denote
the Cartesian lead-field matrix, with local blocks
$\mathbf L_i^{(3)}\in\mathbb R^{M\times3}$.

\paragraph{Fixed orientation ($d=1$)}
The current at source location $i$ is constrained to the
prescribed unit surface normal $\mathbf n_i\in\mathbb R^3$,
expressed in the same coordinate frame as the source
components of $\mathbf L_i^{(3)}$. Writing
$\mathbf x_i^{(3)}(t)=\mathbf n_i x_i(t)$, where $x_i(t)$
is the signed current amplitude, gives
\begin{equation}
  \mathbf l_i^{\mathrm{fix}}
  =
  \mathbf L_i^{(3)}\mathbf n_i
  \in\mathbb R^M.
  \label{eq:fixed_orientation_leadfield}
\end{equation}
The fixed-orientation lead-field matrix is
\[
  \mathbf L^{(1)}
  =
  [\mathbf l_1^{\mathrm{fix}},\ldots,
   \mathbf l_N^{\mathrm{fix}}]
  \in\mathbb R^{M\times N}.
\]

\paragraph{Reduced free-orientation MEG ($d=2$)}
The retained local MEG subspace is defined by the singular
value decomposition
\[
  \mathbf L_i^{(3)}
  =
  \mathbf U_i\mathbf S_i\mathbf V_i^\top,
  \qquad
  s_{i,1}\ge s_{i,2}\ge s_{i,3}\ge0,
\]
where $s_{i,j}$ are the singular values.
The two leading right singular vectors form the
orthonormal basis
\begin{equation}
  \mathbf Q_i
  =
  [\mathbf v_{i,1},\mathbf v_{i,2}]
  \in\mathbb R^{3\times2},
  \qquad
  \mathbf Q_i^\top\mathbf Q_i=\mathbf I_2.
  \label{eq:meg_tangent_basis}
\end{equation}
Representing the local current as
$\mathbf x_i^{(3)}(t)=\mathbf Q_i\mathbf x_i^{(2)}(t)$
yields the reduced lead-field block
\[
  \mathbf L_i^{(2)}
  =
  \mathbf L_i^{(3)}\mathbf Q_i
  \in\mathbb R^{M\times2}.
\]
The full reduced lead field is
\[
  \mathbf L^{(2)}
  =
  \mathbf L^{(3)}\mathbf Q
  \in\mathbb R^{M\times2N},
  \qquad
  \mathbf Q
  =
  \operatorname{blkdiag}
  (\mathbf Q_1,\ldots,\mathbf Q_N).
\]
The SVD-defined subspace need not coincide with the local cortical tangent plane.

In the main text, $\mathbf L=\mathbf L^{(d)}$ for the selected source dimension $d\in\{1,2,3\}$.

\section{S-FLEX basis construction and covariance properties}

\subsection{Gaussian basis and grid-adaptive width}
\label{sec:sflex_grid_width}

The spatial overlap induced by a Gaussian basis depends on its width relative to the source-grid spacing. For a fixed physical width, a finer grid generally produces greater overlap between neighboring basis functions. We therefore express the width relative to a characteristic within-hemisphere nearest-neighbor distance.

For source coordinates
$\mathbf r_1,\ldots,\mathbf r_N\in\mathbb R^3$, let
$\sigma_B>0$ denote the Gaussian width. The kernel is
\begin{equation}
  K_{ij}
  =
  \exp\!\left(
    -\frac{\|\mathbf r_i-\mathbf r_j\|_2^2}{2\sigma_B^2}
  \right),
  \qquad i,j=1,\ldots,N.
  \label{eq:S-FLEX_gaussian_kernel}
\end{equation}
The scalar spatial basis is obtained by symmetric normalization:
\begin{equation}
  \mathbf B_0
  =
  \mathbf D^{-1/2}\mathbf K\mathbf D^{-1/2},
  \qquad
  \mathbf D
  =
  \operatorname{diag}\!\left(
    \sum_{j=1}^{N}K_{1j},\ldots,\sum_{j=1}^{N}K_{Nj}
  \right).
  \label{eq:S-FLEX_normalized_basis}
\end{equation}
This normalization preserves symmetry and rescales the kernel according to its local weight sums.

Let $\mathcal H(i)$ denote the set of source locations in the same cortical hemisphere as source $i$. Define
\begin{equation}
  d_{i,\mathrm{NN}}
  =
  \min_{\substack{j\in\mathcal H(i)\\j\neq i}}
  \|\mathbf r_i-\mathbf r_j\|_2,
  \qquad
  \widetilde d_{\mathrm{NN}}
  =
  \operatorname{median}_{i=1,\ldots,N} d_{i,\mathrm{NN}}.
  \label{eq:S-FLEX_grid_spacing}
\end{equation}
The within-hemisphere restriction excludes cross-hemisphere neighbors when estimating grid spacing, while the median reduces sensitivity
to local mesh irregularities. 


The basis width is parameterized as
\begin{equation}
  \sigma_B = \kappa_B \widetilde d_{\mathrm{NN}},
  \qquad \kappa_B>0,
  \label{eq:S-FLEX_grid_adaptive_width}
\end{equation}
where $\kappa_B$ controls spatial overlap relative to the grid spacing. The unnormalized Gaussian weight at the characteristic distance is
\begin{equation}
  w_{\mathrm{NN}}
  =
  \exp\!\left(
    -\frac{\widetilde d_{\mathrm{NN}}^{\,2}}{2\sigma_B^2}
  \right)
  =
  \exp\!\left(-\frac{1}{2\kappa_B^2}\right).
  \label{eq:S-FLEX_typical_nn_weight}
\end{equation}

Small values of $\kappa_B$ yield a localized basis, whereas larger values increase spatial overlap. For distinct source locations,
$\mathbf K \to \mathbf I_N$ as $\kappa_B \to 0^+$. For every $\kappa_B>0$, all entries of the scalar basis $\mathbf B_0$ remain strictly positive in exact arithmetic. Thus, the width controls its effective spatial extent without changing its full-support property.

\subsection{Orientation-specific S-FLEX operators}
\label{sec:sflex_orientation}

In the fixed-orientation setting, $\mathbf B=\mathbf B_0$ acts on scalar coefficients associated with the prescribed
source orientations. For free-orientation EEG expressed in a common Cartesian frame, the basis acts componentwise:
\begin{equation*}
  \mathbf B=\mathbf B_0\otimes\mathbf I_3.
\end{equation*}

In reduced free-orientation MEG, the retained coordinates depend on the source location. Thus,
$\mathbf B_0\otimes\mathbf I_2$ would treat distinct local frames as a common coordinate frame and is not appropriate. Let
$\mathbf Q_i\in\mathbb R^{3\times2}$ contain the orthonormal basis vectors of the retained MEG-sensitive subspace at source $i$. The orientation-aware S-FLEX operator has blocks
\begin{equation}
  \mathbf B_{ij}^{(2)}
  =
  (\mathbf B_0)_{ij}\mathbf Q_i^\top\mathbf Q_j,
  \label{eq:S-FLEX_reduced_orientation_coupling}
\end{equation}
where $\mathbf B_0$ is the normalized scalar spatial basis.
Equivalently,
\begin{equation}
  \mathbf B^{(2)}
  =
  \mathbf Q^\top
  (\mathbf B_0\otimes\mathbf I_3)
  \mathbf Q,
  \qquad
  \mathbf Q
  =
  \operatorname{blkdiag}(\mathbf Q_1,\ldots,\mathbf Q_N).
  \label{eq:S-FLEX_reduced_orientation_global}
\end{equation}
This construction maps local coefficients into a common Cartesian frame, applies the spatial basis componentwise, and projects the result onto the retained local subspaces.

Although the scalar Gaussian weights are strictly positive,
an off-diagonal orientation-aware block can be rank deficient when $\mathbf Q_i^\top\mathbf Q_j$ is singular. This alone does
not imply a singular local posterior covariance: its rank
depends on the combined retained basis blocks at that source, as analyzed in Section~\ref{sec:sflex_covariance_rank}.


\subsection{Rank of the source-space posterior covariance}
\label{sec:sflex_covariance_rank}

Let $\mathcal A$ denote the set of retained coefficient groups,
with $|\mathcal A|=K_{\mathrm{act}}$. After removing the zero-variance coefficients, the source-space posterior covariance can be written as
\begin{equation*}
  \boldsymbol{\Sigma}_{\mathbf x}
  =
  \mathbf B_{\mathcal A}
  \boldsymbol{\Sigma}_{\mathbf c,\mathcal A}
  \mathbf B_{\mathcal A}^{\top},
\end{equation*}
where $\mathbf B_{\mathcal A}\in\mathbb R^{dN\times dK_{\mathrm{act}}}$
contains the retained basis columns and
$\boldsymbol{\Sigma}_{\mathbf c,\mathcal A}$ is their posterior covariance. Hence,
\begin{equation}
  \operatorname{rank}(\boldsymbol{\Sigma}_{\mathbf x})
  \leq
  \operatorname{rank}(\boldsymbol{\Sigma}_{\mathbf c,\mathcal A})
  \leq dK_{\mathrm{act}}.
  \label{eq:sflex_covariance_rank}
\end{equation}
When $K_{\mathrm{act}}<N$, the full source-space covariance
is therefore singular, even if all individual source components
have positive marginal variances. The source currents remain restricted to the subspace spanned by the retained basis columns.

This global rank deficiency does not necessarily make each
local covariance block singular. At source $i$,
\begin{equation*}
  \boldsymbol{\Sigma}_{\mathbf x,ii}
  =
  \mathbf B_{i,\mathcal A}
  \boldsymbol{\Sigma}_{\mathbf c,\mathcal A}
  \mathbf B_{i,\mathcal A}^{\top},
\end{equation*}
where $\mathbf B_{i,\mathcal A}$ contains the corresponding
$d$ rows of $\mathbf B_{\mathcal A}$. If
$\boldsymbol{\Sigma}_{\mathbf c,\mathcal A}$ is positive definite, the local block is positive definite precisely when $\mathbf B_{i,\mathcal A}$ has row rank $d$.


For fixed orientation and for free-orientation EEG in a common
Cartesian frame, any nonempty active set $\mathcal A$ gives
$\mathbf B_{i,\mathcal A}$ row rank $d$, since it contains a
positive scalar multiple of $\mathbf I_d$. In reduced
free-orientation MEG, row rank is lost at source $i$ precisely
when there exists a nonzero $\mathbf u\in\mathbb R^2$ such that
$\mathbf Q_j^\top\mathbf Q_i\mathbf u=\mathbf 0$ for every $j\in\mathcal A$. In other words, a direction in the retained local subspace at $i$ is orthogonal to all active local subspaces. This cannot occur when $i\in\mathcal A$, since
$\mathbf B_{ii}^{(2)}=(\mathbf B_0)_{ii}\mathbf I_2$
and $(\mathbf B_0)_{ii}>0$.

\section{Supplementary Simulation Figures}
\label{sec:supp-simulation-results-final}

The default setting is $\alpha_{\mathrm{SNR}}=0.5$ and $\mathrm{NNZ}=5$. Each calibration curve or boxplot distribution summarizes 100 held-out evaluation runs (25 per head model across four head models). Sweep conditions are evaluated separately. Recalibration maps are learned from independent
calibration simulations for each orientation and noise specification. Post-fixed coincides with post-oracle at the default setting and is therefore shown only in the SNR and NNZ sweeps. Axis ranges may differ between metric panels.

\subsection{Fixed-orientation simulation results}
\label{sec:supp-fixed-results}

\begin{figure}[!htbp]
  \centering
  \includegraphics[width=0.98\linewidth]{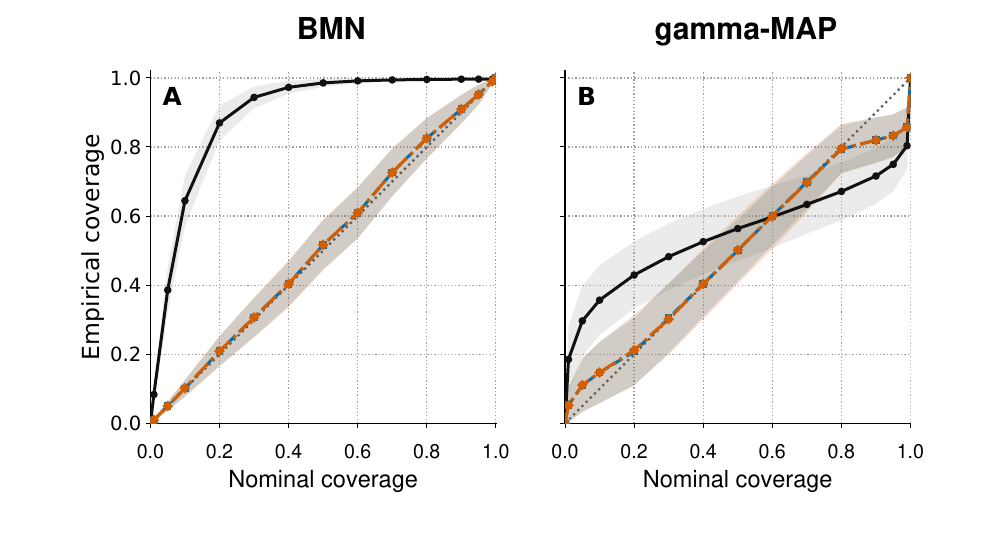}
  \caption{Fixed-orientation calibration curves derived from scalar credible-interval containment under adaptive noise at $\alpha_{\mathrm{SNR}}=0.5$ and $\mathrm{NNZ}=5$. \textbf{A}: BMN+sLORETA. \textbf{B}: $\gamma$-MAP+S-FLEX. Lines denote mean empirical coverage over 100 evaluation runs; color-matched bands denote plus or minus one run-level standard deviation. The dotted diagonal indicates perfect calibration.}
  \label{fig:supp-adaptive-fixed-calibration}
\end{figure}

Supplementary Figure~\ref{fig:supp-adaptive-fixed-calibration}
shows the fixed-orientation calibration curves under adaptive noise learning. BMN+sLORETA exhibits predominantly conservative coverage before recalibration, with post-oracle and post-mismatch
corrections remaining close over much of the nominal credibility range. The $\gamma$-MAP+S-FLEX curve shows both overcoverage and undercoverage, with residual deviations near the upper credibility levels.

\begin{figure}[!htbp]
  \centering
  \includegraphics[width=0.98\linewidth,height=0.73\textheight,keepaspectratio]{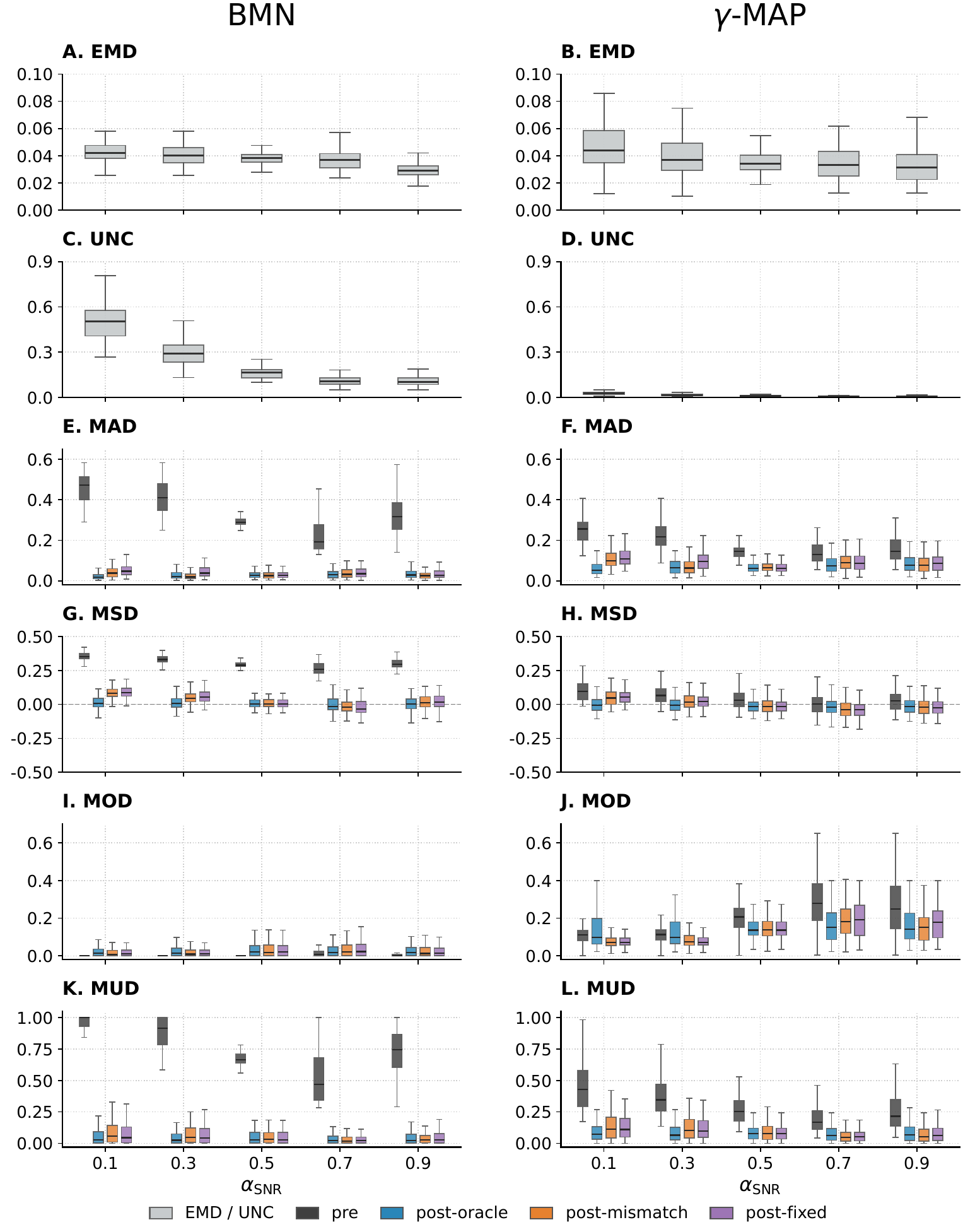}
  
\caption{Fixed-orientation SNR sweep under adaptive noise, varying $\alpha_{\mathrm{SNR}}\in\{0.1,0.3,0.5,0.7,0.9\}$ with $\mathrm{NNZ}=5$. Panel pairs \textbf{A--B}, \textbf{C--D}, \textbf{E--F}, \textbf{G--H}, \textbf{I--J}, and \textbf{K--L} show EMD, UNC, MAD, MSD, MOD, and MUD, respectively, for BMN+sLORETA (left) and $\gamma$-MAP+S-FLEX (right). Calibration modes are pre, post-oracle, post-mismatch, and post-fixed.}
\label{fig:supp-adaptive-snr}
\end{figure}

\begin{figure}[!htbp]
  \centering
  \includegraphics[width=0.98\linewidth,height=0.73\textheight,keepaspectratio]{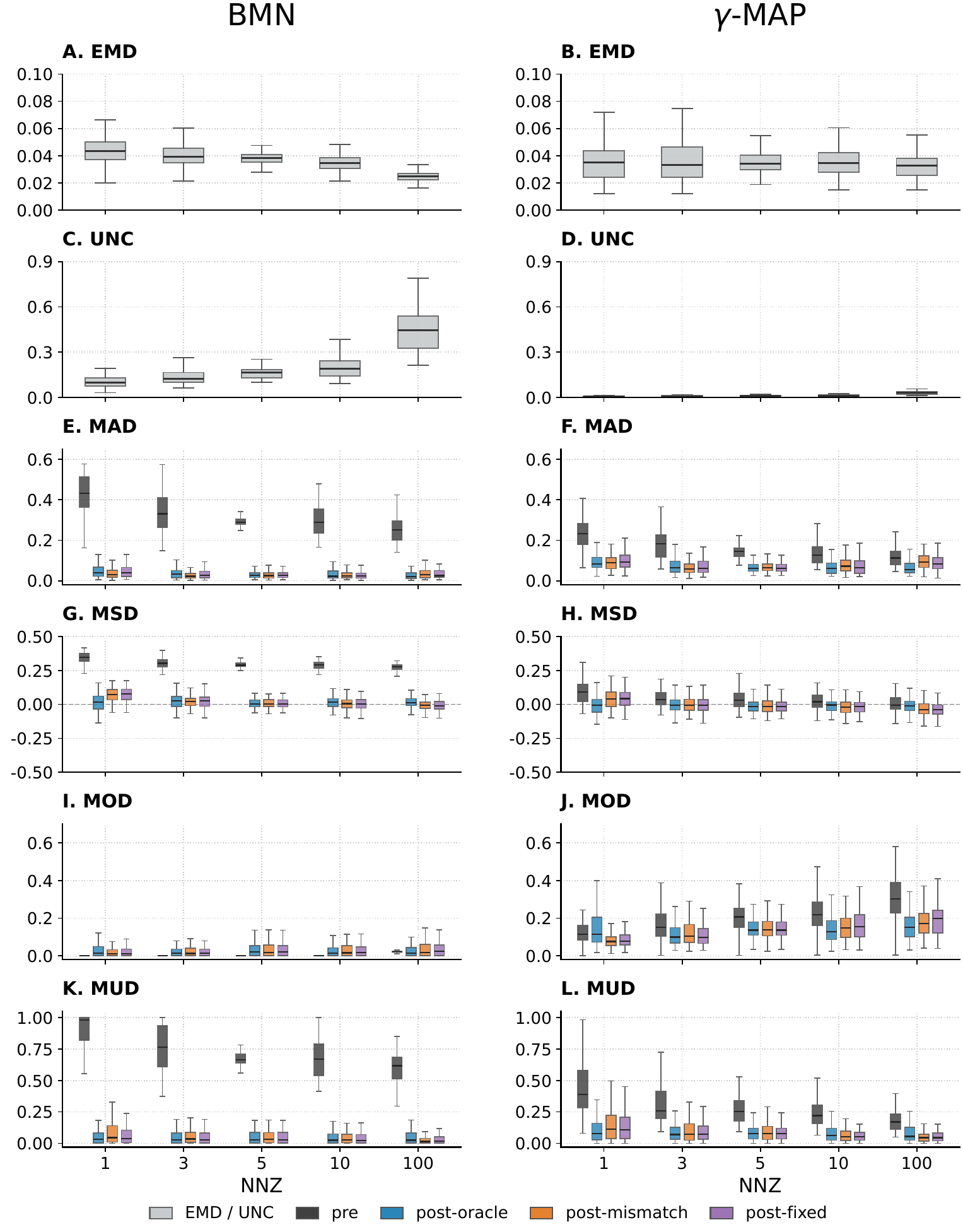}
  
\caption{Fixed-orientation source-number sweep under adaptive noise, varying $\mathrm{NNZ}\in\{1,3,5,10,100\}$ with $\alpha_{\mathrm{SNR}}=0.5$. Panel pairs \textbf{A--B}, \textbf{C--D}, \textbf{E--F}, \textbf{G--H}, \textbf{I--J}, and \textbf{K--L} show EMD, UNC, MAD, MSD, MOD, and MUD, respectively, for BMN+sLORETA (left) and $\gamma$-MAP+S-FLEX (right). Calibration modes are pre, post-oracle, post-mismatch, and post-fixed.}
\label{fig:supp-adaptive-nnz}
\end{figure}

In the adaptive-noise SNR sweep
(Supplementary Figure~\ref{fig:supp-adaptive-snr}),
both EMD and UNC decrease as SNR increases, with a stronger contraction of UNC for BMN+sLORETA. Post-oracle recalibration reduces the calibration deviations across the grid, whereas the transfer strategies leave larger residual errors. In the NNZ sweep (Supplementary Figure~\ref{fig:supp-adaptive-nnz}),
BMN+sLORETA EMD decreases with source number, while
$\gamma$-MAP+S-FLEX EMD remains comparatively stable.
UNC increases with NNZ for both estimators. Raw calibration errors are more variable in the sparsest configurations.

\begin{figure}[!htbp]
  \centering
  \includegraphics[width=0.98\linewidth]{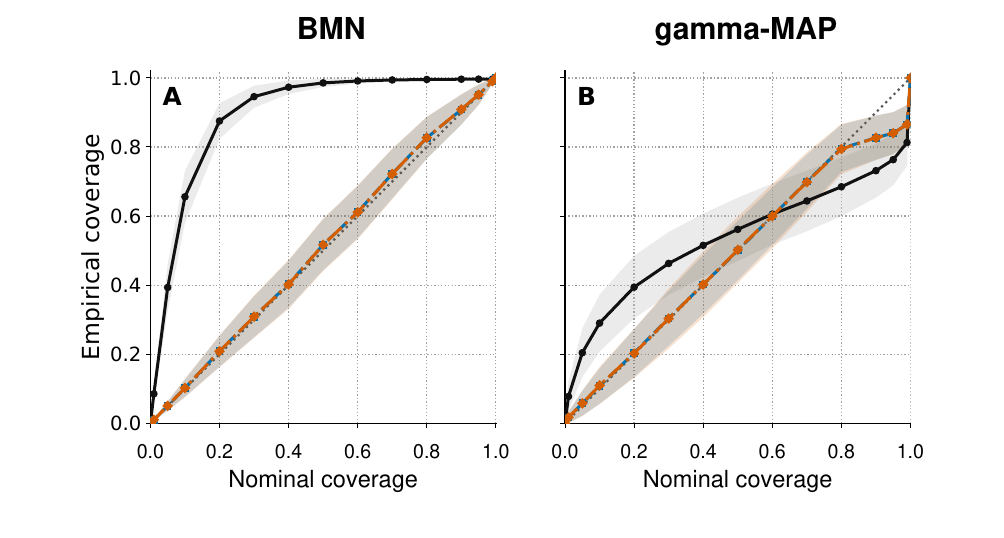}
  \caption{Fixed-orientation calibration curves derived from scalar credible-interval containment under oracle noise at $\alpha_{\mathrm{SNR}}=0.5$ and $\mathrm{NNZ}=5$. \textbf{A}: BMN+sLORETA. \textbf{B}: $\gamma$-MAP+S-FLEX. Lines denote mean empirical coverage over 100 evaluation runs; color-matched bands denote plus or minus one run-level standard deviation. The dotted diagonal indicates perfect calibration.}
  \label{fig:supp-oracle-fixed-calibration}
  \label{fig:supp-oracle-calibration}
\end{figure}

\begin{figure}[!htbp]
  \centering
  \includegraphics[width=0.98\linewidth,height=0.73\textheight,keepaspectratio]{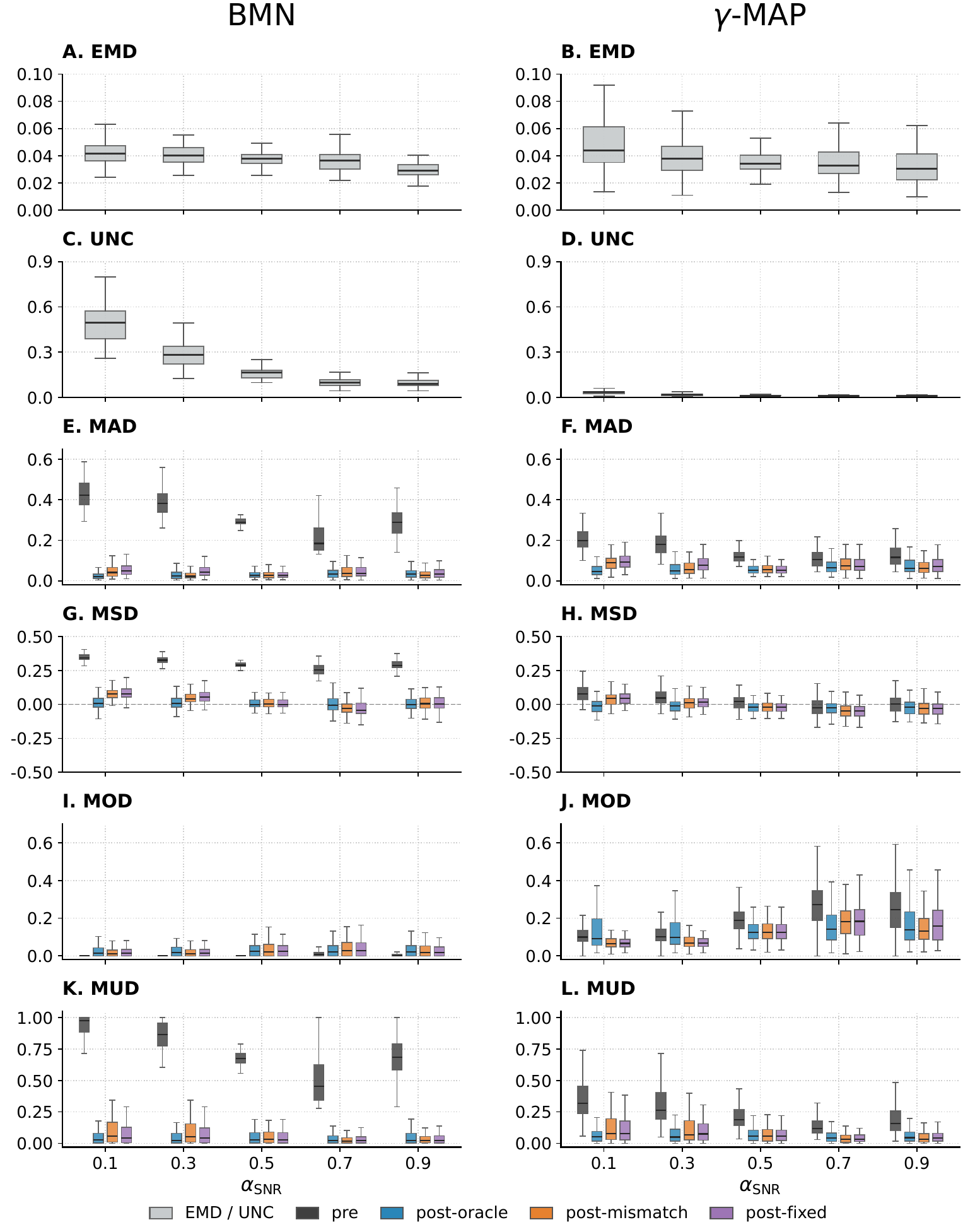}
  
\caption{Fixed-orientation SNR sweep under oracle noise, varying $\alpha_{\mathrm{SNR}}\in\{0.1,0.3,0.5,0.7,0.9\}$ with $\mathrm{NNZ}=5$. Panel pairs \textbf{A--B}, \textbf{C--D}, \textbf{E--F}, \textbf{G--H}, \textbf{I--J}, and \textbf{K--L} show EMD, UNC, MAD, MSD, MOD, and MUD, respectively, for BMN+sLORETA (left) and $\gamma$-MAP+S-FLEX (right). Calibration modes are pre, post-oracle, post-mismatch, and post-fixed.}
\label{fig:supp-oracle-snr}
\end{figure}

\begin{figure}[!htbp]
  \centering
  \includegraphics[width=0.98\linewidth,height=0.73\textheight,keepaspectratio]{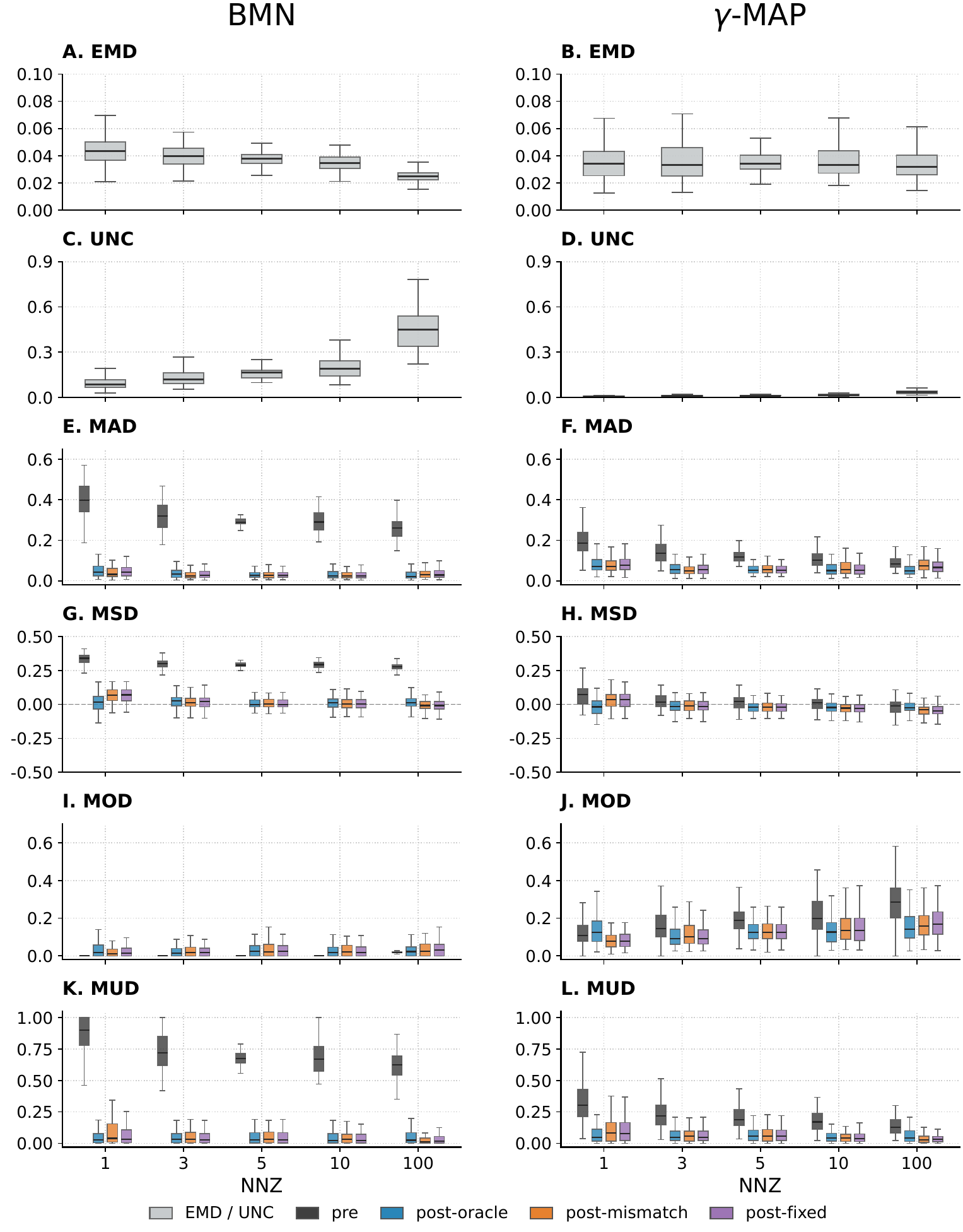}
  
\caption{Fixed-orientation source-number sweep under oracle noise, varying $\mathrm{NNZ}\in\{1,3,5,10,100\}$ with $\alpha_{\mathrm{SNR}}=0.5$. Panel pairs \textbf{A--B}, \textbf{C--D}, \textbf{E--F}, \textbf{G--H}, \textbf{I--J}, and \textbf{K--L} show EMD, UNC, MAD, MSD, MOD, and MUD, respectively, for BMN+sLORETA (left) and $\gamma$-MAP+S-FLEX (right). Calibration modes are pre, post-oracle, post-mismatch, and post-fixed.}
  \label{fig:supp-oracle-nnz}
\end{figure}

\begin{figure}[!htbp]
  \centering
  \includegraphics[width=0.98\linewidth]{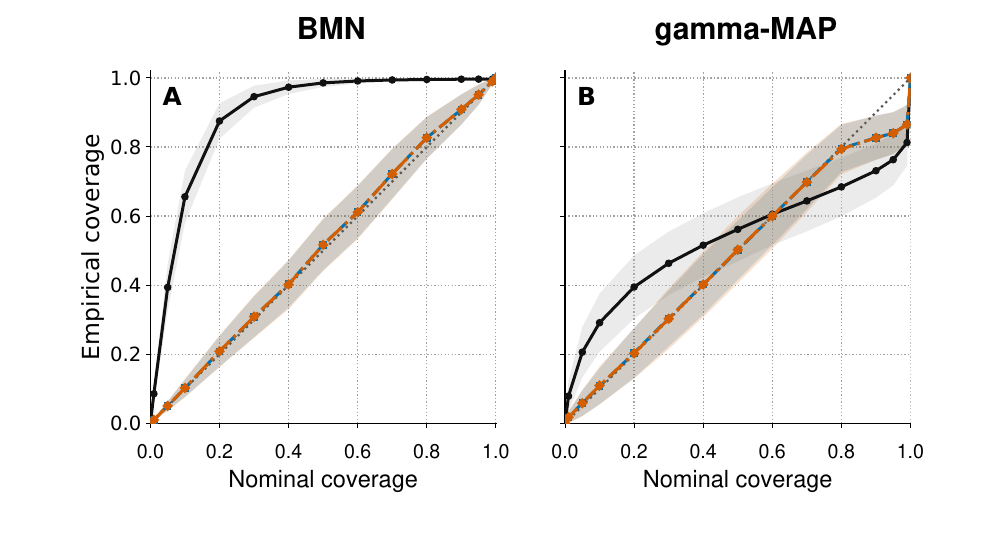}
  \caption{Fixed-orientation calibration curves derived from scalar credible-interval containment under baseline noise at $\alpha_{\mathrm{SNR}}=0.5$ and $\mathrm{NNZ}=5$. \textbf{A}: BMN+sLORETA. \textbf{B}: $\gamma$-MAP+S-FLEX. Lines denote mean empirical coverage over 100 evaluation runs; color-matched bands denote plus or minus one run-level standard deviation. The dotted diagonal indicates perfect calibration.}
  \label{fig:supp-baseline-fixed-calibration}
  \label{fig:supp-baseline-calibration}
\end{figure}

\begin{figure}[!htbp]
  \centering
  \includegraphics[width=0.98\linewidth,height=0.73\textheight,keepaspectratio]{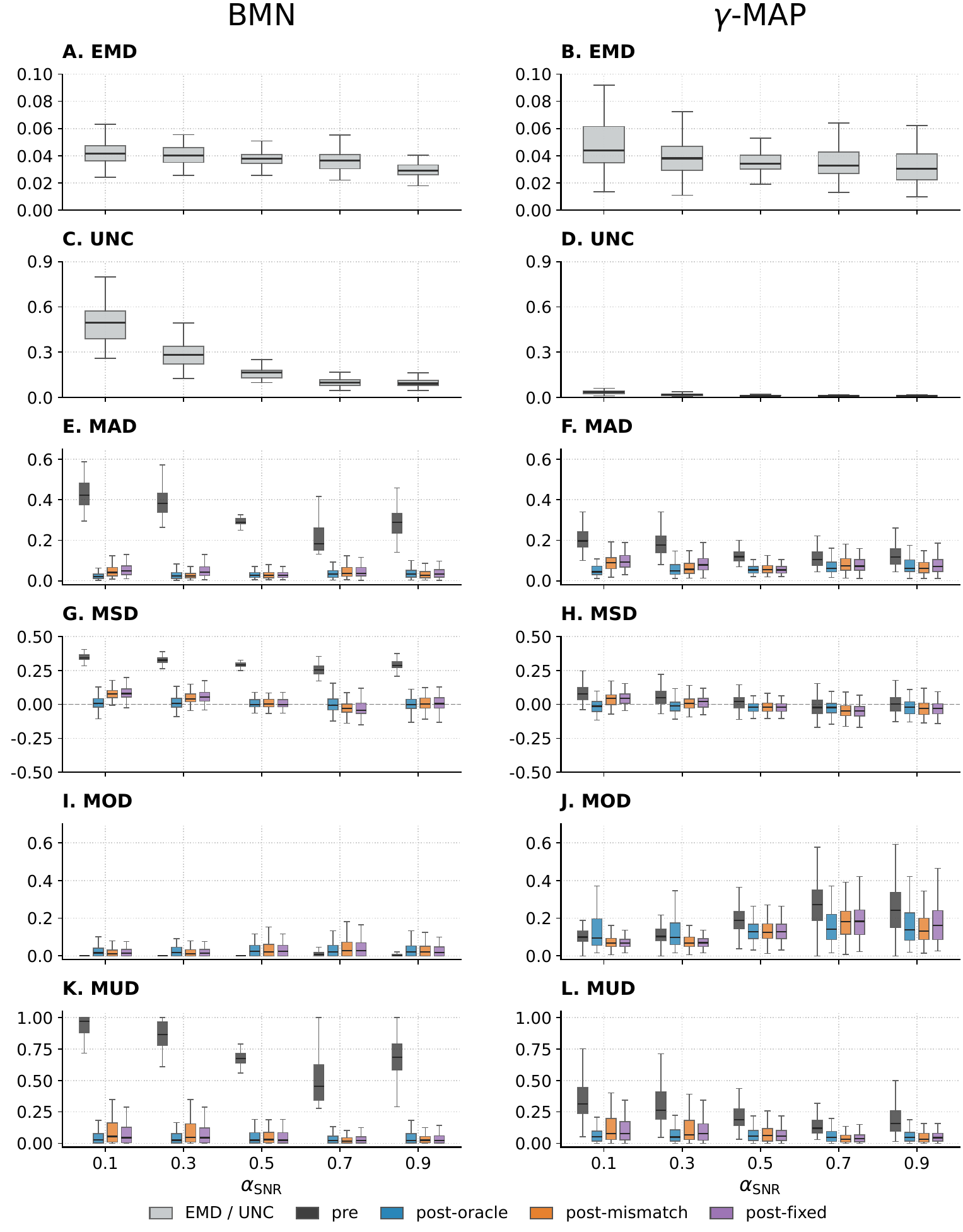}
  
\caption{Fixed-orientation SNR sweep under baseline noise, varying $\alpha_{\mathrm{SNR}}\in\{0.1,0.3,0.5,0.7,0.9\}$ with $\mathrm{NNZ}=5$. Panel pairs \textbf{A--B}, \textbf{C--D}, \textbf{E--F}, \textbf{G--H}, \textbf{I--J}, and \textbf{K--L} show EMD, UNC, MAD, MSD, MOD, and MUD, respectively, for BMN+sLORETA (left) and $\gamma$-MAP+S-FLEX (right). Calibration modes are pre, post-oracle, post-mismatch, and post-fixed.}
  \label{fig:supp-baseline-snr}
\end{figure}

\begin{figure}[!htbp]
  \centering
  \includegraphics[width=0.98\linewidth,height=0.73\textheight,keepaspectratio]{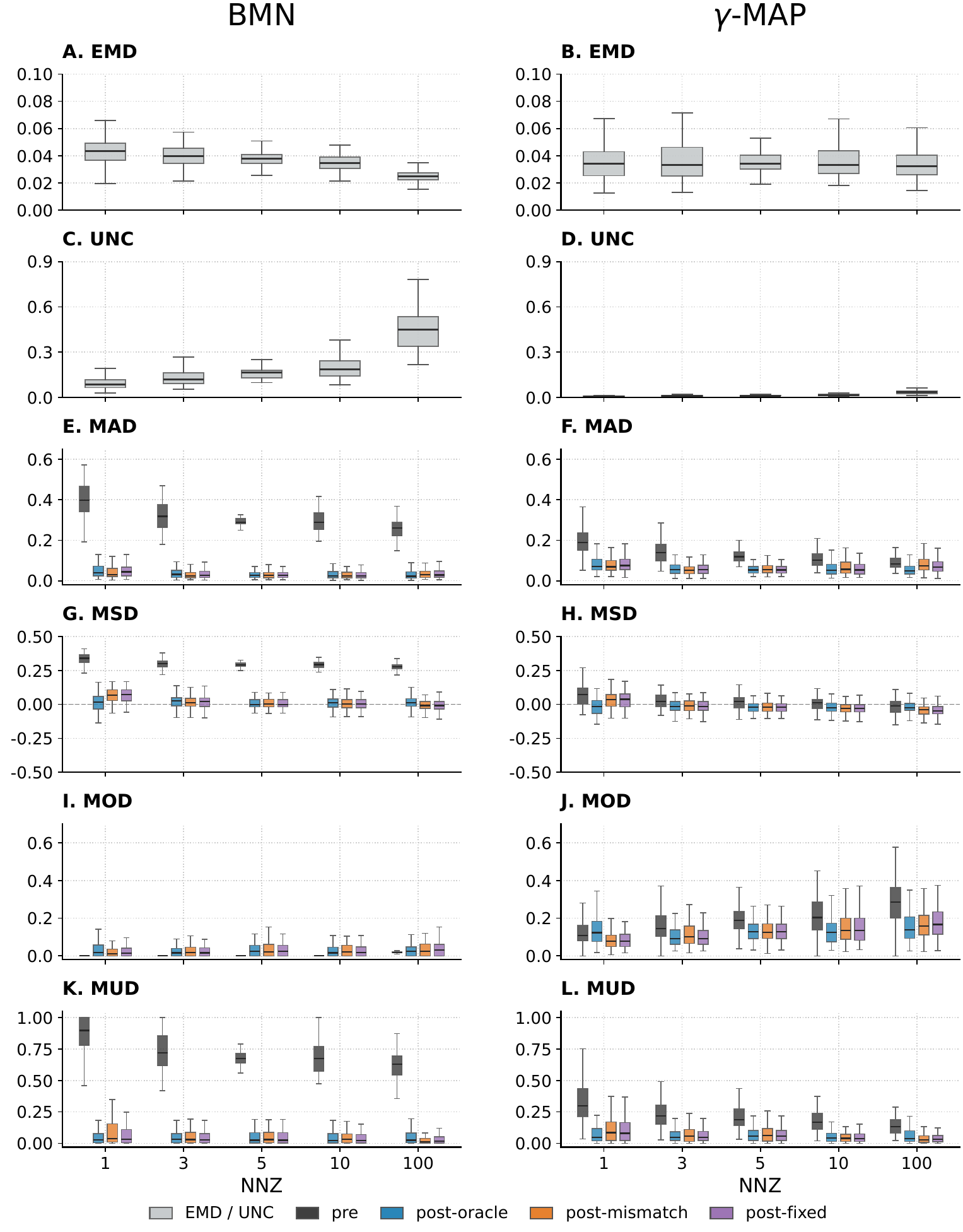}
  
\caption{Fixed-orientation source-number sweep under baseline noise, varying $\mathrm{NNZ}\in\{1,3,5,10,100\}$ with $\alpha_{\mathrm{SNR}}=0.5$. Panel pairs \textbf{A--B}, \textbf{C--D}, \textbf{E--F}, \textbf{G--H}, \textbf{I--J}, and \textbf{K--L} show EMD, UNC, MAD, MSD, MOD, and MUD, respectively, for BMN+sLORETA (left) and $\gamma$-MAP+S-FLEX (right). Calibration modes are pre, post-oracle, post-mismatch, and post-fixed.}
\label{fig:supp-baseline-nnz}
\end{figure}

The oracle-noise results
(Supplementary Figures~\ref{fig:supp-oracle-fixed-calibration},
\ref{fig:supp-oracle-snr}, and~\ref{fig:supp-oracle-nnz})
show similar estimator-specific trends. Post-oracle
recalibration reduces the dominant calibration errors,
although MSD and MOD indicate residual signed deviations
under transfer. The baseline-noise calibration curves
(Supplementary Figure~\ref{fig:supp-baseline-fixed-calibration}) and parameter sweeps
(Supplementary Figures~\ref{fig:supp-baseline-snr}
and~\ref{fig:supp-baseline-nnz}) closely follow their oracle-noise counterparts. Differences between fixed- and free-orientation results cannot be attributed solely to orientation, since the coverage targets and simulation realizations differ.

\end{document}